\documentclass[12pt]{article}
\ifdefined\pdfsuppresswarningpagegroup
\fi
\usepackage[utf8]{inputenc}
\usepackage{graphicx}
\graphicspath{{figures/}}
\usepackage{fontspec}
\usepackage{microtype}
\usepackage{amsmath}
\usepackage{amssymb}
\usepackage{unicode-math}
\usepackage[margin=1in]{geometry}
\usepackage[numbers,sort&compress]{natbib}
\usepackage{xurl}
\usepackage{xcolor}
\usepackage[colorlinks=true, allcolors=blue]{hyperref}
\usepackage{float}
\usepackage[font=small,labelfont=bf]{caption}
\DeclareCaptionLabelSeparator{bar}{ $\vert$\ }
\newif\ifsubmission
\submissionfalse
\ifsubmission\else
\fi
\usepackage{booktabs}
\usepackage{array}
\newcolumntype{L}[1]{>{\raggedright\arraybackslash}p{#1}}
\usepackage{multirow}
\usepackage{longtable}
\usepackage{indentfirst}

\setcitestyle{super,open={},close={}}
\title{\bfseries Variable names provide priors: semantic grounding enables data-efficient numerical prediction by large language models}

\author{
\begin{minipage}{0.92\textwidth}
\centering
Xin-Rui Ma\textsuperscript{1,2}, Shi-Zhi Chen\textsuperscript{1,*},
Hai-Yan Jiang\textsuperscript{1}, De-Cheng Feng\textsuperscript{2,*},
Michael Beer\textsuperscript{3,4,5}\\[0.6em]
\begin{minipage}{\linewidth}
\setlength{\leftskip}{0pt}
\setlength{\rightskip}{0pt}
\setlength{\parfillskip}{0pt plus 1fil}
\fontsize{9.5}{12}\selectfont
\setlength{\parindent}{0pt}
\setlength{\parskip}{3pt}
\hangindent=1.2em\hangafter=1
\makebox[1.2em][l]{\textsuperscript{1}}School of Highway, Chang'an University, Xi'an 710064, China\endgraf
\hangindent=1.2em\hangafter=1
\makebox[1.2em][l]{\textsuperscript{2}}School of Civil Engineering, Southeast University, Nanjing 211189, China\endgraf
\hangindent=1.2em\hangafter=1
\makebox[1.2em][l]{\textsuperscript{3}}Institute for Risk and Reliability, Leibniz University Hannover, Hannover, 30167, Germany\endgraf
\hangindent=1.2em\hangafter=1
\makebox[1.2em][l]{\textsuperscript{4}}Department of Civil and Environmental Engineering, University of Liverpool, Liverpool, L69~3GH, United Kingdom\endgraf
\hangindent=1.2em\hangafter=1
\makebox[1.2em][l]{\textsuperscript{5}}International Joint Research Center for Resilient Infrastructure \& International Joint Research Center for Engineering Reliability and Stochastic Mechanics, Tongji University, Shanghai, 200092, China\endgraf
\vspace{3pt}
\hangindent=1.2em\hangafter=1
\makebox[1.2em][l]{\textsuperscript{*}}Corresponding authors: szchen@chd.edu.cn (S.-Z.C.); dcfeng@seu.edu.cn (D.-C.F.)\endgraf
\end{minipage}
\end{minipage}
}
\date{}

\begin{document}
\maketitle

\begin{abstract}
Large language models (LLMs) have achieved strong performance across diverse tasks, from question answering to code generation. 
Recent studies have found that LLMs can perform numerical prediction on tabular datasets via in-context learning.
However, it remains unclear under what conditions and how LLMs use variable names and in-context examples for numerical prediction.
Here we show that semantic grounding, the match between variable names and the values they describe, allows five LLMs without task-specific training to match or outperform seven supervised machine learning (ML) models across 20 datasets from nine scientific domains, particularly when examples are scarce.
Reasoning traces indicate that variable names allow LLMs to draw on domain-specific knowledge acquired during pretraining.
Using controlled tasks based on physical formulas, we further show that LLMs integrate domain-specific knowledge with in-context examples in a way consistent with Bayesian updating.
Specifically, when the relationship between the inputs and the target cannot be inferred from the examples alone, LLMs rely on domain-specific knowledge as the prior. 
Conversely, as contradicting examples accumulate, LLM predictions shift progressively towards these examples, which act as the likelihood.
Our work reveals how LLMs use domain-specific knowledge in numerical prediction and positions them as data-efficient predictors for data-scarce scientific domains.

\end{abstract}

\section{Introduction}

Pretrained on large and diverse text corpora, general-purpose large language models (LLMs) have acquired broad knowledge and demonstrated capabilities across tasks including language understanding, code generation and mathematical problem solving\cite{brown2020language,openai2023gpt4,guo2025deepseek}.
Recent studies have shown that LLMs can perform numerical prediction on tabular datasets directly from in-context examples containing variable names and numerical values\cite{vacareanu2024words,nafar2025learning}.
Unlike supervised machine learning (ML) models, which typically learn statistical relationships solely from numerical values\cite{grinsztajn2022why,hollmann2025accurate}, LLMs can also use the semantic information conveyed by variable names.
This semantic information may allow LLMs to draw on knowledge acquired during pretraining\cite{requeima2024llm,singhal2023large,luo2024large}.

Previous studies have investigated the roles of numerical values and variable names in LLM prediction.
On the one hand, LLMs can learn linear, nonlinear and high-dimensional functions from the numerical information of in-context examples\cite{vacareanu2024words,codaforno2023meta,agarwal2024manyshot,akata2026incontext}.
On the other hand, LLMs can estimate the effect of each feature on the target from variable names without any numerical values\cite{capstick2025autoelicit}, suggesting that LLMs can exploit the semantic information these names carry.
A few studies further find that adding variable names to in-context examples improves predictions\cite{nafar2025learning,requeima2024llm,wei2023larger}, suggesting that LLMs can use both variable names and the examples.
One study attributes this ability to both the retrieval of pretrained knowledge and learning from the examples\cite{nafar2025learning}, whereas another finds that adding in-context examples can suppress the recall of scientific knowledge\cite{jang2026incontext}.
However, in these studies, variable names and examples reflect the same relation, so their individual contributions cannot be disentangled.
In addition, these studies rely mainly on either synthetic data or real-world datasets from limited domains, which undermines the generalisability of their findings.
Moreover, these datasets may be contaminated\cite{bordt2024elephants}, making it difficult to separate the roles of pretrained knowledge and memorisation.
More fundamentally, it remains unclear what enables LLMs to draw on pretrained knowledge, and these studies have not quantified how LLMs integrate this knowledge with in-context examples.

Here we evaluate five LLMs from five independent developers (Claude Opus 4.5, GPT-5.2, Gemini 3 Flash, Qwen3.5-397B and DeepSeek-V3.2) on 20 real-world tabular datasets spanning nine scientific domains (Fig.~\ref{fig:overview}).
These domains range from structural and geotechnical engineering to chemistry, biomedicine and urban mobility. 
For reference, we also evaluate seven representative supervised ML models on the same in-context examples.
To mitigate data contamination, we include six datasets released after the training cutoffs of all five LLMs (Supplementary Table~19).
Across these domains, LLMs perform comparably to or better than supervised ML models. 
This performance depends on variable names and is most pronounced under semantic grounding, the match between variable names and the values they describe.
Reasoning traces indicate that variable names allow the LLMs to draw on domain-specific knowledge acquired during pretraining.
Using 25 controlled tasks built from established physical formulas, we further show that four of the five LLMs integrate this knowledge with in-context examples in a way consistent with Bayesian updating, in which the knowledge acts as the prior and the examples as the likelihood.
When the examples cannot distinguish the roles of two inputs, the predictions follow the roles implied by the variable names, and exchanging the two names reverses the predicted relation.
As more examples contradict the names, the predictions shift progressively towards these examples.
We quantify this shift and compare it with the predictions of a Bayesian observer.
These findings not only reveal under what conditions and how LLMs perform numerical prediction, but also position LLMs as data-efficient predictors for scientific domains where data are scarce.

\newcommand{\capOverview}{\textbf{Overview of the study design.}
The top panel shows how each tabular dataset pairs variable names, which carry semantic information, with numerical values, which carry numerical information. The 20 datasets used in this study span nine scientific domains, and six of them were released after the training cutoffs of all five large language models (LLMs). On the left, seven supervised machine learning (ML) models are given the labelled examples. On the right, the five LLMs receive the same examples in a prompt and return a single numerical prediction without any parameter updates. Four questions are addressed. (i) What improves the numerical predictions of LLMs? Prompts without names (\textit{Anonymised}) are compared with prompts that include them (\textit{Correct variable names}). (ii) When do variable names help? To test whether names must match the values they describe (semantic grounding), the values are rescaled (\textit{Rescaled values}), the names are reassigned across features (\textit{Shuffled names}) or replaced with names from an unrelated domain (\textit{Cross-domain names}). (iii) Why do variable names help? Reasoning excerpts from DeepSeek-V3.2 on the Shear Force dataset are shown, with domain-specific terms in red and generic terms in blue. (iv) How is domain-specific knowledge integrated with the in-context examples? In the controlled relation tasks, the two inputs share the same value in every example, so only their names reveal the roles of the two inputs. Some of the examples are then replaced by ones that contradict the names. The prior weight measures how strongly each LLM follows the names rather than the examples (logos from top to bottom are Gemini 3 Flash, Claude Opus 4.5, GPT-5.2, Qwen3.5-397B and DeepSeek-V3.2). Supplementary Methods gives the detailed prompts.}

\ifsubmission\else
\begin{figure}[p]
\centering
\includegraphics[width=0.99\textwidth]{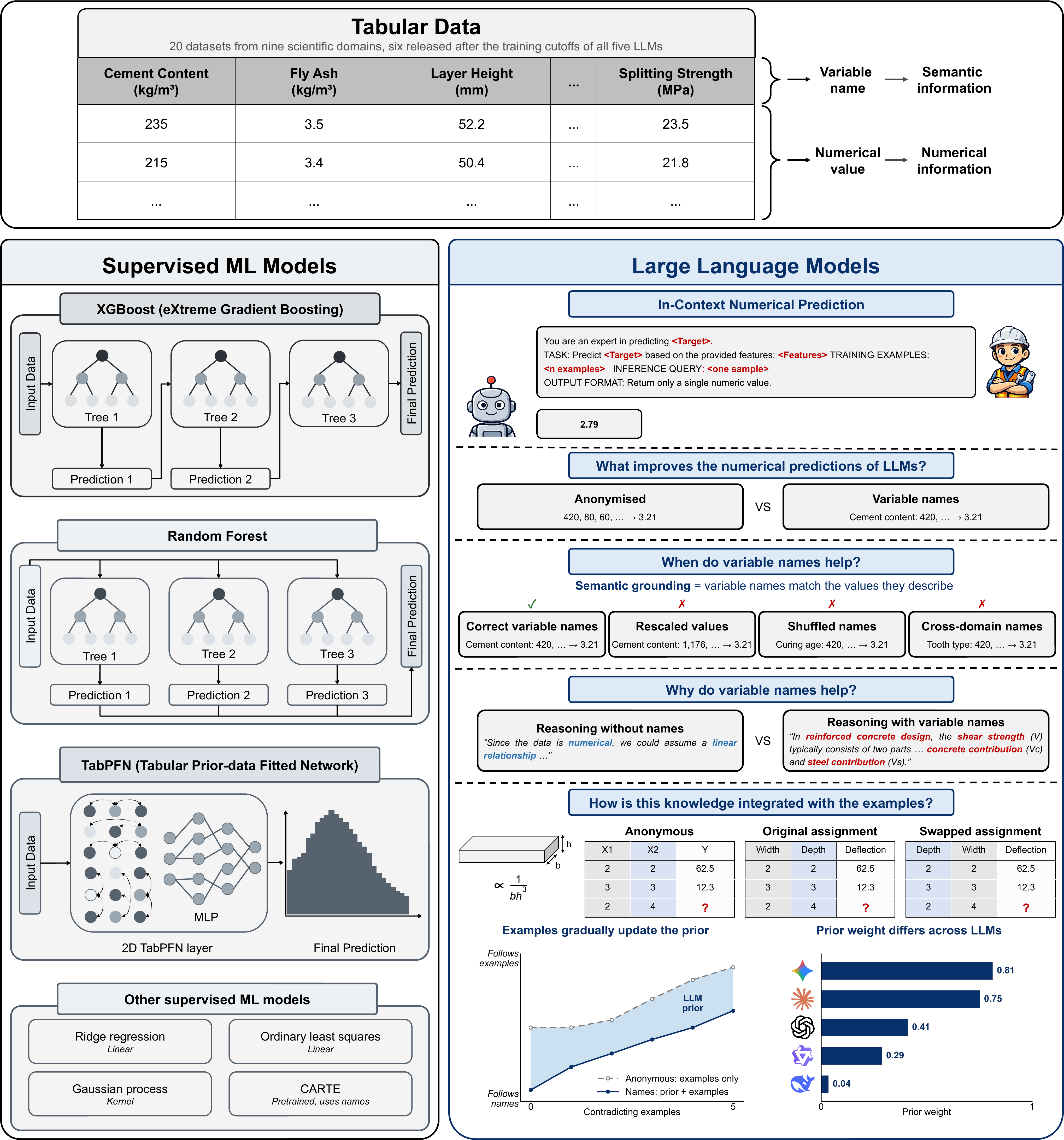}   %
\caption{\capOverview}
\label{fig:overview}
\end{figure}
\fi

\section{Results}

\subsection{\texorpdfstring{LLMs match or outperform supervised ML models across scientific domains}{LLMs match or outperform supervised ML models across scientific domains}}

To test LLMs' generalisability in numerical prediction, we collected 20 real-world tabular datasets from nine domains, ranging from biomedicine to geotechnical engineering (Supplementary Table~18).
These datasets contain 113 to 821 samples and 4 to 30 features, with variable values spanning 11 orders of magnitude ($10^{-4}$ to $10^{7}$).
On these datasets, we evaluated five LLMs from different developers: Claude Opus 4.5, GPT-5.2, Gemini 3 Flash, Qwen3.5-397B and DeepSeek-V3.2.
These LLMs received $n$ in-context examples ($n = 0$ to $100$) and predicted the targets of test samples without parameter updates.
To mitigate data contamination, we included six datasets released after the training cutoffs of all five LLMs (post-cutoff datasets, Supplementary Table~19).
For reference, seven representative supervised ML models were evaluated on the same examples (Fig.~\ref{fig:overview}, left).
The main text reports the three best-performing ones (TabPFN, tuned XGBoost and tuned random forest), and Supplementary Tables~2 and 3 give the results for the other four.

At $n=100$, the five LLMs performed comparably to these three best-performing supervised ML models on datasets from all nine domains (Fig.~\ref{fig:fig1}a).
On the six post-cutoff datasets (shaded in Fig.~\ref{fig:fig1}a, Methods), the LLMs remained comparable to the supervised ML models (Supplementary Note~1, Supplementary Fig.~1 and Supplementary Table~1), suggesting that memorisation of the datasets alone does not explain their performance\cite{bordt2024elephants}.

\newcommand{\capFigOne}{\textbf{Numerical prediction performance of large language models (LLMs) and supervised machine learning (ML) models across scientific domains.}
Five LLMs are compared with the three best-performing of seven supervised ML models (TabPFN, tuned XGBoost and tuned random forest) on 20 datasets over five random seeds, and $n$ is the number of examples per model. Summaries are medians over seeds and then across datasets, and group summaries pool all models in a group within each dataset.
\textbf{a}, Coefficient of determination ($R^2$) for each dataset at $n=100$, grouped by domain. Points show individual models and ranges span the models in each group. Shading marks post-cutoff datasets.
\textbf{b}, Weighted absolute percentage error (WAPE) and $R^2$ from $n=0$ to $n=100$. Bars and lines show group summaries and points show individual models. Whiskers show nominal 95\% confidence intervals for the median, and arrows mark values beyond the plotting range. Supervised ML models are not shown at $n=0$ because they require training examples.
\textbf{c}, Per-dataset comparison. The advantage of the LLMs at small $n$ holds on nearly every dataset and narrows as examples are added, with the two groups close at $n=100$. WAPE is on the left and $R^2$ on the right. Points plot the LLM group summary against the supervised ML group summary for each dataset from $n=5$ to $n=100$, and lines connect the same dataset across numbers of examples. Numbers give the count of datasets in the shaded region, where the LLMs perform better, and stars mark post-cutoff datasets.}

\ifsubmission\else
\begin{figure}[p]   %
\centering
\includegraphics[width=157.63mm]{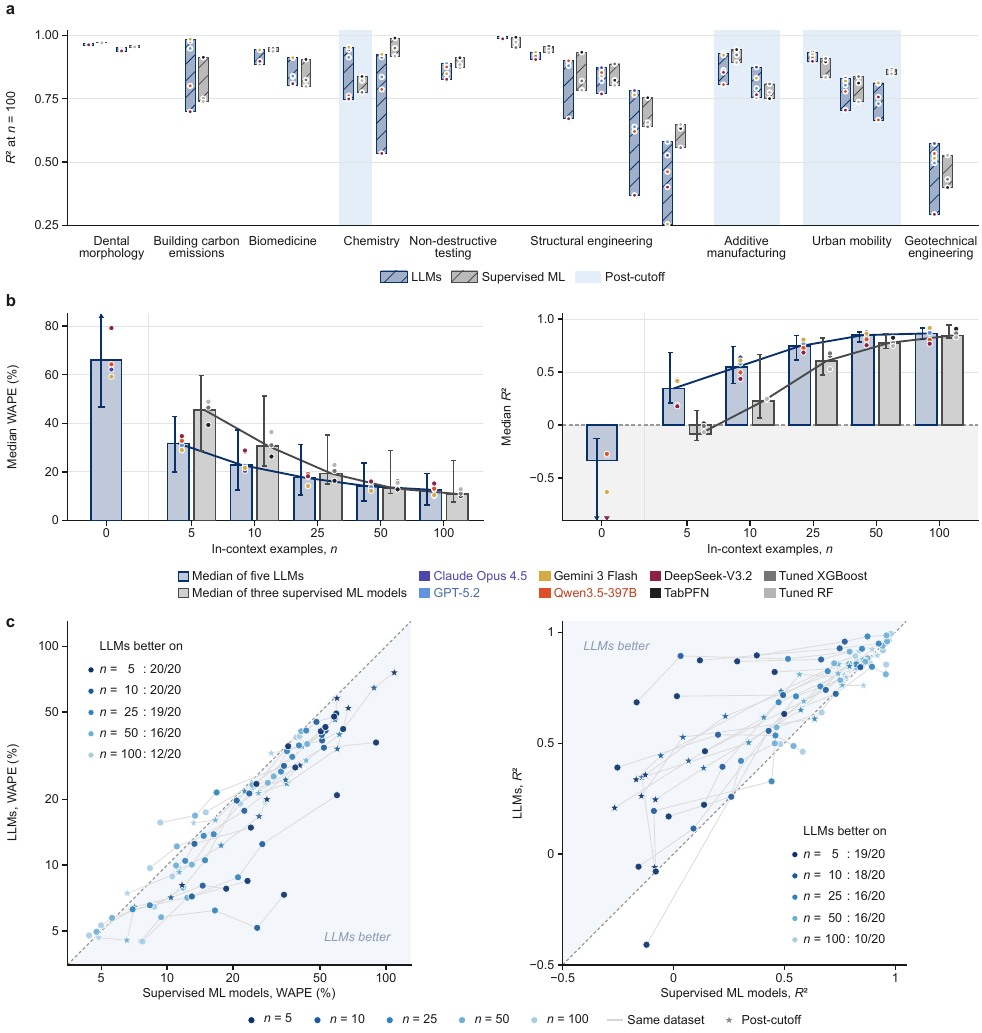}   %
\caption{\capFigOne}
\label{fig:fig1}
\end{figure}
\fi

However, in some scientific domains, even 100 samples can be difficult to obtain.
We therefore evaluated the LLMs with fewer in-context examples, from $n = 0$ to $50$ (Fig.~\ref{fig:fig1}b).
We began with the extreme case of no examples ($n = 0$), in which the LLMs, given only variable names, had to rely solely on pretrained knowledge.
In this setting, their median weighted absolute percentage error (WAPE) was 66.0\%, and their median coefficient of determination ($R^2$) was $-0.335$, indicating that pretrained knowledge alone was insufficient for accurate numerical prediction.
With only five examples, however, every LLM outperformed all three best-performing supervised ML models on both metrics, with $R^2$ of 0.177 to 0.429 versus $-0.066$ to 0.015 and WAPE of 28.9\% to 34.7\% versus 39.3\% to 48.8\%. 
As more examples were added, the gap between the LLMs and the supervised ML models narrowed (Supplementary Figs.~2 and 3). 
From $n=5$ to $n=50$, the median $R^2$ of the LLMs rose from 0.346 to 0.849 and that of the supervised ML models from $-0.081$ to 0.777, reducing the gap from 0.43 to 0.07. 
By $n=100$, the gap had essentially closed, with the LLMs achieving a slightly higher median $R^2$ (0.865 versus 0.846) and the supervised ML models a slightly lower median WAPE (10.9\% versus 12.6\%).

To test whether this trend held on individual datasets, we compared the LLMs with the supervised ML models on each dataset (Fig.~\ref{fig:fig1}c).
At $n=5$, the LLMs held a clear advantage over supervised ML models. The median performance of the five LLMs exceeded that of the three supervised ML models on 19 of the 20 datasets in $R^2$ and on all 20 datasets in WAPE, including all six post-cutoff datasets (stars in Fig.~\ref{fig:fig1}c).
As more examples were added, this advantage diminished, until the LLMs and supervised ML models performed comparably at $n=100$, with the LLMs leading on 10 of the 20 datasets in $R^2$ and 12 in WAPE (Fig.~\ref{fig:fig1}c).
Comparisons paired by random seed gave similar results, with the LLMs ahead on 17 of the 20 datasets in $R^2$ and 19 in WAPE at $n=5$ (Supplementary Note~1, Supplementary Fig.~4 and Supplementary Tables~4 and 22).

Together, these results show that LLMs, given variable names and in-context examples, can perform numerical prediction across diverse scientific domains without task-specific training, even when examples are few.
These results suggest that the observed performance depends on combining pretrained knowledge with in-context examples.

\subsection{LLMs rely on variable names for numerical prediction}

We next asked whether the numerical values or the semantic information conveyed by variable names allows LLMs to draw on pretrained knowledge.
To disentangle their contributions, we compared the original prompts (\textit{Correct variable names}) with an \textit{Anonymised} condition (Methods). In this condition, we omitted the feature names and replaced the target name with a generic label, keeping the numerical values and all other settings unchanged.

Under the \textit{Anonymised} condition at $n = 0$, the LLMs had no information about the task, and their median $R^2$ was $-13.4$ (Fig.~\ref{fig:fig3}a, left).
Without variable names, the LLMs improved steadily as $n$ increased from 5 to 100 (Fig.~\ref{fig:fig3}a, left). They performed comparably to the supervised ML models with five or ten examples but fell below TabPFN from $n=25$ onwards.
Variable names, by contrast, improved predictions even without examples, raising the median $R^2$ at $n = 0$ to $-0.335$.
With the same examples, variable names further increased the median $R^2$ by 0.343 over examples alone at $n = 5$, but this gain diminished as the number of examples increased, to 0.04 at $n = 100$ (Fig.~\ref{fig:fig3}a, right).
We therefore focused subsequent analyses on $n = 5$, where the contribution of variable names was largest.
At this number of examples, variable names improved the median $R^2$ of every LLM by 0.25 to 0.51 (Fig.~\ref{fig:fig3}b, rightmost column) and increased $R^2$ in 93 of the 100 LLM--dataset pairs.
The remaining seven pairs showed decreases of 0.02 to 0.18 (Fig.~\ref{fig:fig3}b, Supplementary Note~2 and Supplementary Fig.~5).

Together, these results indicate that both numerical values and variable names contribute to the predictions of LLMs, and that variable names allow them to draw on pretrained knowledge.
Variable names, in particular, substantially improve predictions when examples are scarce while requiring no additional data, making them especially valuable in scientific domains where data are difficult to obtain.

\newcommand{\capFigThree}{\textbf{Effect of variable names on the numerical predictions of large language models (LLMs) across numbers of examples.}
Five LLMs are evaluated on 20 datasets under the \textit{Anonymised} and \textit{Correct variable names} conditions over five random seeds, and $n$ is the number of examples per model. Condition summaries are medians over seeds and LLMs within each dataset and then across datasets. Improvement is the increase in the coefficient of determination ($R^2$) from \textit{Anonymised} to \textit{Correct variable names}.
\textbf{a}, Left, $R^2$ of the two conditions across $n$. Thick lines show condition summaries and thin dashed lines show individual LLMs (squares for \textit{Correct variable names} and circles for \textit{Anonymised}). Grey lines show three supervised machine learning (ML) models, which are not shown at $n=0$. TabPFN, the strongest at most $n$, is shown in dark grey with triangles, and tuned XGBoost (diamonds) and tuned random forest (inverted triangles) are shown in lighter grey dashed lines. Bands show nominal 95\% confidence intervals (CIs) for the condition summaries and TabPFN. Right, improvement across $n$. Bars show the improvement, labelled above each bar, with 95\% CIs from 2,000 paired bootstrap resamples of datasets, and points show individual LLMs. In both plots, the vertical axis is broken to show values at $n=0$, with a logarithmic segment below $-0.45$ (left) or in the upper part (right).
\textbf{b}, Improvement at $n=5$ for each LLM and dataset, computed from medians over seeds. Columns are datasets in the order of Fig.~\ref{fig:fig1}a. The rightmost column shows each LLM's overall improvement, which equals its point at $n=5$ in the right plot of \textbf{a}. Negative values are outlined in red and labelled, and stars mark post-cutoff datasets.}

\ifsubmission\else
\begin{figure}[t!]   %
\centering
\includegraphics[width=152.62mm]{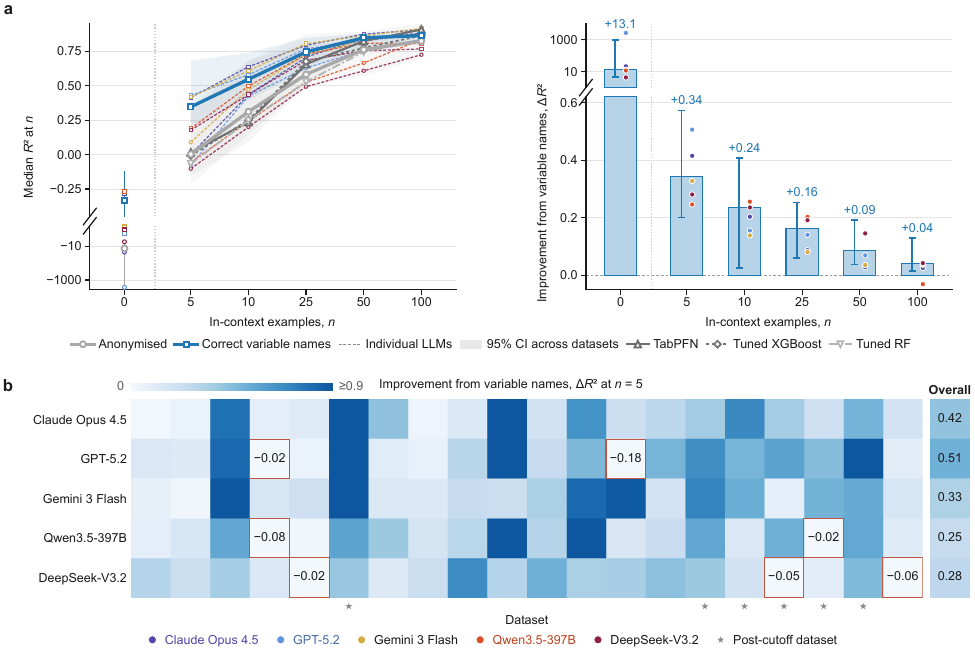}   %
\caption{\capFigThree}
\label{fig:fig3}
\end{figure}
\fi

\subsection{Variable names improve predictions most when they match the values they describe}

\newcommand{\capFigNames}{\textbf{Sources of the improvement from variable names in large language models (LLMs).}
Five LLMs are evaluated on 20 datasets at $n=5$ examples per model over five random seeds. Improvement is the increase in the coefficient of determination ($R^2$) over the corresponding \textit{Anonymised} condition, and summaries are medians computed as in Fig.~\ref{fig:fig3}. Coloured points and ribbons denote individual LLMs.
\textbf{a}, Bars show the improvement for each LLM from the target name, the feature names or all variable names. Error bars show 95\% confidence intervals (CIs) from 2,000 paired bootstrap resamples of datasets.
\textbf{b}, Median $R^2$ without and with feature names in the original prompt format and in a standardised format, in which the features are given as codes and both conditions include the target name and units (Methods). Bars show condition summaries with 95\% CIs across datasets, points show individual LLMs and brackets give the improvement.
\textbf{c}, Improvement under \textit{Correct variable names}, \textit{Rescaled values}, \textit{Shuffled names} and \textit{Cross-domain names}. Bars show the improvement, and error bars show 95\% CIs as in \textbf{a}.
\textbf{d}, $R^2$ under prompt guidance. Bars show condition summaries, with values given on the right, and points show individual LLMs. CIs are not shown.
\textbf{e}, Summary of the improvement contrasts in \textbf{a} and \textbf{c}. Numerical magnitude denotes the difference in improvement between \textit{Correct variable names} and \textit{Rescaled values} in \textbf{c}; variable identity denotes that between \textit{Rescaled values} and \textit{Shuffled names}; and target identity denotes the gain from adding the target name to the feature names in \textbf{a}. Ribbon width indicates the size of each contrast, and dashed ribbons indicate negative contrasts.}

\ifsubmission\else
\begin{figure}[t!]   %
\centering
\includegraphics[width=156.88mm]{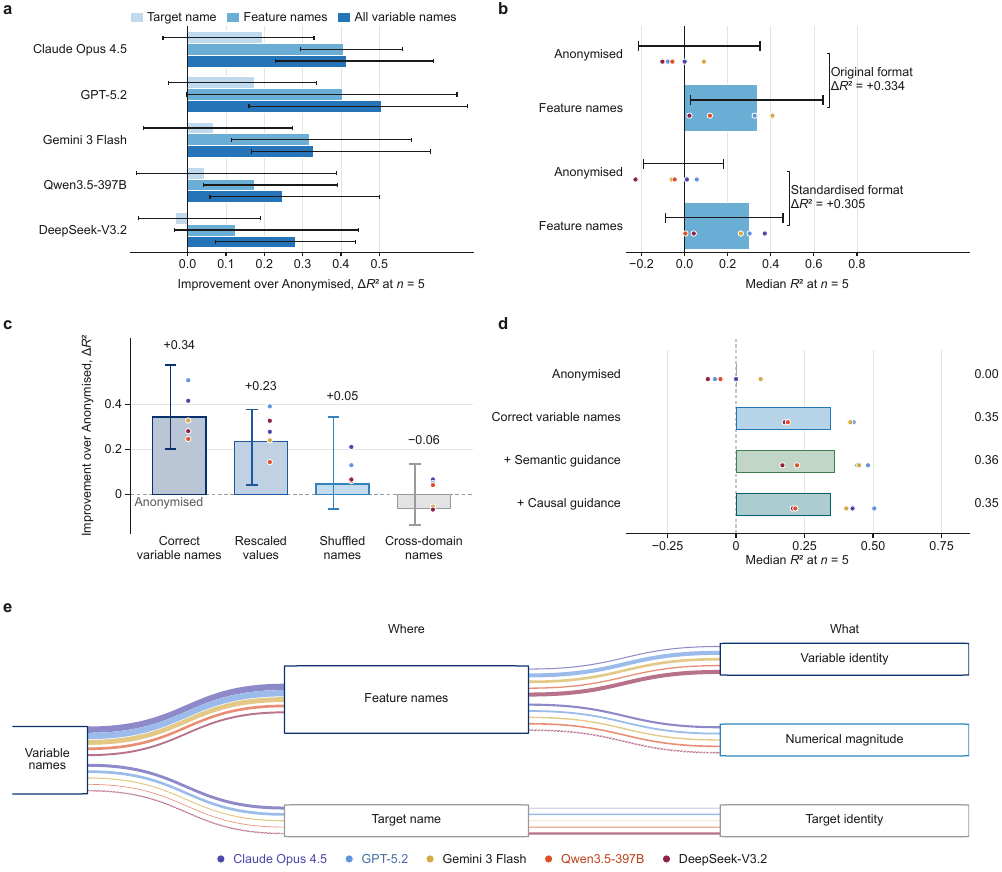}   %
\caption{\capFigNames}
\label{fig:names}
\end{figure}
\fi

We next asked under which conditions variable names improve predictions.
We first separated the contributions of the two components of variable names, namely the feature names, which specify what the input values represent, and the target name, which identifies the quantity to be predicted (Fig.~\ref{fig:names}a).
Although the magnitude of improvement varied across LLMs, feature names improved predictions in all five LLMs, and the target name in four of the five (Supplementary Fig.~6a,b and Supplementary Table~6).
Across the five LLMs, feature names alone increased the median $R^2$ by 0.334 and the target name alone by 0.063.
Once feature names were present, however, adding the target name increased the median $R^2$ by only a further 0.009, suggesting that much of the information it conveys is already captured by the feature names.
In a standardised prompt format, feature names produced a similar improvement (0.305; Fig.~\ref{fig:names}b, Methods, Supplementary Note~3 and Supplementary Table~7).
Both components therefore contributed to the improvement, so the following analyses treated the variable names as a whole.

We then perturbed the match between names and values in three ways that kept the names meaningful (Fig.~\ref{fig:names}c and Methods).
With \textit{Correct variable names}, the improvement over the \textit{Anonymised} condition was 0.343.
With \textit{Rescaled values}, which preserved the link between each name and its feature but altered the scales, the improvement fell to 0.234, indicating that names remained informative even when the magnitudes no longer matched.
With \textit{Shuffled names}, which kept domain-relevant names but attached them to the wrong features, the improvement nearly vanished (0.046), showing that relevant names help little when they describe the wrong values.
With \textit{Cross-domain names}, which replaced the names with those from unrelated domains, the improvement disappeared, falling to $-0.063$ (Supplementary Note~3, Supplementary Figs.~6c,d and 7 and Supplementary Table~22).
Thus, the improvement from variable names stems not from meaningful names alone, but from the match between the names and the values they describe, which we term \textit{semantic grounding}.

We then tested whether explicit instructions about the variable names could further improve predictions (Fig.~\ref{fig:names}d and Methods). Two types of instruction were added to the \textit{Correct variable names} prompts.
The first, \textit{+ Semantic guidance}, asked the LLMs to treat the feature names as informative.
The second, \textit{+ Causal guidance}, additionally asked them to consider causal relationships between the features and the target.
At $n = 5$, \textit{+ Semantic guidance} improved the median $R^2$ by only 0.014 relative to \textit{Correct variable names}, and \textit{+ Causal guidance} by less than 0.001.
Neither provided substantial additional improvement at any number of examples (Supplementary Note~3, Supplementary Fig.~8 and Supplementary Tables~9 and 10).
Variable names therefore improved predictions without the LLMs being instructed to use them.

Figure~\ref{fig:names}e summarises the analyses in this section, splitting the improvement from variable names by the names that provide it and by the information these names convey.
Most of the improvement was obtained with feature names alone. The difference in improvement between \textit{Rescaled values} and \textit{Shuffled names} was 0.189, compared with 0.109 between \textit{Correct variable names} and \textit{Rescaled values}.
The target name conveyed target identity but contributed only 0.009 once feature names were present.

Together, these results show that variable names improve predictions most under semantic grounding, the match between names and the values they describe.

\subsection{Variable names allow LLMs to draw on pretrained domain-specific knowledge}

We next investigated what pretrained knowledge the LLMs draw on from variable names.
To characterise this knowledge, we collected the reasoning traces that the LLMs produced before answering, with extended reasoning enabled, under both \textit{Correct variable names} and \textit{Anonymised} at $n = 5$.
Enabling extended reasoning brought no substantial improvement in accuracy (Supplementary Note~4).
We first removed from the traces all words that appeared in the prompt, so that the remaining content reflected knowledge elicited by the prompt rather than copied from it (Methods).

\newcommand{\capFigTraces}{\textbf{Effect of variable names on the reasoning traces of large language models (LLMs).}
\textbf{a}, t-distributed stochastic neighbour embedding (t-SNE) maps of LLM reasoning traces at $n=5$ examples per model on 20 datasets over five random seeds, embedded after removing all words that appear in the prompt (Methods). Coloured points show traces under \textit{Correct variable names}, by domain, and grey points show traces under \textit{Anonymised}. The large map shows Qwen3.5-397B and DeepSeek-V3.2, which return their full reasoning (150 traces per condition and dataset), and the list gives the words most over-represented in each domain and under \textit{Anonymised}. The small maps show each LLM separately (60 traces per condition and dataset), using reasoning summaries for Claude Opus 4.5, GPT-5.2 and Gemini 3 Flash, which do not return full reasoning.
\textbf{b}, Assignment of a trace to a domain, illustrated with a sentence from a DeepSeek-V3.2 trace on the Shear Force dataset. Words that appear in the prompt (struck out) are removed, and the trace is embedded with a sentence encoder. A classifier trained on traces with all variable names from other seeds then gives the probability of each domain, and $p$ is the probability of the trace's own domain. Bar heights are illustrative.
\textbf{c}, Reasoning and prediction accuracy of DeepSeek-V3.2 and Qwen3.5-397B as variable names are added step by step, from no names and the target name only (0\% of feature names) to all feature names (100\%), on 20 datasets at $n=5$ with one seed (Methods). First plot, distribution of $p$ pooled over both LLMs, shown as stacked density curves. Second plot, $\bar{p}$, the mean of $p$ over the traces of each dataset, for each LLM, with the dashed line marking chance (1/9) and values labelled at no names and 100\%. Third plot, number of domain terms per trace (Methods). Fourth plot, change in the coefficient of determination ($R^2$) relative to no names, paired within each dataset before taking the median across datasets. In the second to fourth plots, points are medians across datasets with 95\% confidence intervals from 2,000 bootstrap resamples of datasets, and the dotted line marks where the target name is added.}
\ifsubmission\else
\begin{figure}[p]   %
\centering
\includegraphics[width=\textwidth]{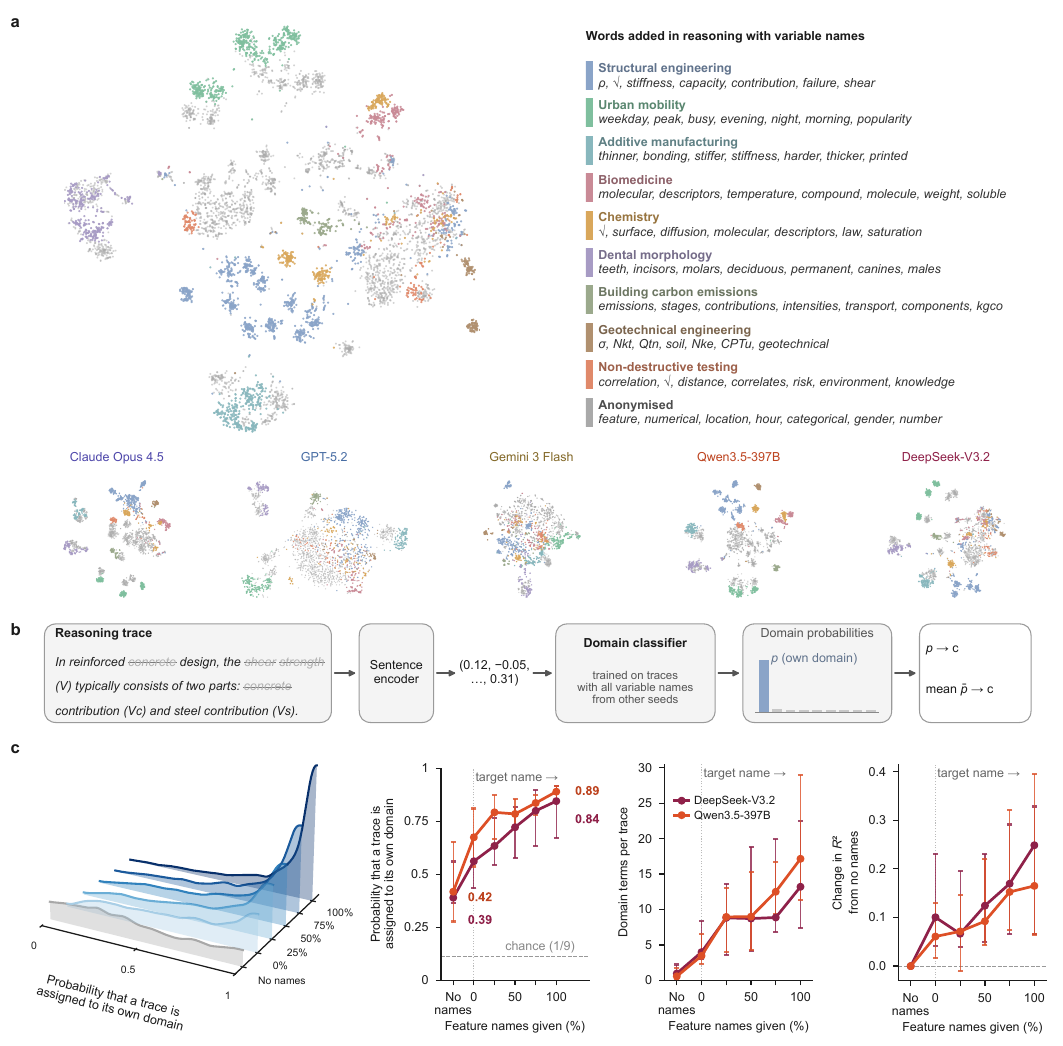}   %
\caption{\capFigTraces}
\label{fig:traces}
\end{figure}
\fi

To visualise how the content of the reasoning differed between the two conditions, we embedded each trace with a sentence encoder and projected the embeddings into two dimensions using t-distributed stochastic neighbour embedding (t-SNE; Fig.~\ref{fig:traces}a and Methods).
Under \textit{Correct variable names} (coloured points), the traces formed clusters by domain, whereas under \textit{Anonymised} (grey points) they overlapped across domains.
The same pattern appeared in each of the five LLMs.
Under \textit{Correct variable names}, the reasoning drew on domain-specific knowledge absent from the prompt, such as stiffness and failure in structural engineering, incisors and molars in dental morphology, and the cone factor $N_{kt}$ in geotechnical engineering.
Under \textit{Anonymised}, by contrast, the most characteristic words described the format of the data rather than a domain, such as feature, numerical and categorical (Supplementary Note~4 and Supplementary Table~12).
Variable names thus allow LLMs to draw on domain-specific knowledge for numerical prediction.

We next asked whether providing more variable names allows LLMs to draw on more domain-specific knowledge.
We analysed the reasoning traces of DeepSeek-V3.2 and Qwen3.5-397B, the only two of the five LLMs that return their full reasoning traces.
The other three return only brief summaries, which omit most of the reasoning traces and therefore cannot reveal how much domain-specific knowledge was used.
Starting from prompts without any names, we added the target name and then the feature names progressively, from 25\% to 100\% (Methods).
To quantify how domain-specific each trace was, we trained a classifier on traces generated with all variable names in the other random seeds and recorded the probability it assigned to the trace's own domain (Fig.~\ref{fig:traces}b,c and Methods).
Without any variable names, the classifier assigned a trace to its own domain with a median probability of 0.39 for DeepSeek-V3.2 and 0.42 for Qwen3.5-397B.
With all variable names, this probability rose to 0.84 and 0.89, respectively (Fig.~\ref{fig:traces}c, second plot).
Over the same range, the number of domain terms per trace rose from 1.0 to 13.2 and from 0.6 to 17.2, respectively (Fig.~\ref{fig:traces}c, third plot, and Supplementary Table~13).
More variable names thus allowed LLMs to draw on more domain-specific knowledge.

We then asked whether the greater use of domain-specific knowledge was accompanied by more accurate predictions.
From prompts without any variable names to prompts with all variable names, $R^2$ increased by a median of 0.25 for DeepSeek-V3.2 and 0.17 for Qwen3.5-397B, and improved in 17 of the 20 datasets for both LLMs (Fig.~\ref{fig:traces}c, fourth plot).
Thus, as the LLMs drew on more domain-specific knowledge, their predictions also became more accurate.

Together, these results show that variable names allow LLMs to draw on pretrained domain-specific knowledge.

\newcommand{\capFigFour}{
\textbf{Relative influence of pretrained domain-specific knowledge and in-context examples on the predictions of large language models (LLMs).}
\textbf{a}, Design of the controlled relation tasks, illustrated with beam deflection. In-context examples use $x_1=x_2$ and are therefore consistent with either assignment of the feature names, whereas prediction inputs use unequal values, for which the two assignments imply different targets. The relation score compares each prediction with that for the same input with $x_1$ and $x_2$ swapped. A score of $+1$ corresponds to the Original relation, $-1$ to the Swapped relation and 0 to no preference.
\textbf{b}, Relation scores for 25 tasks from seven fields, grouped by functional form, under the three conditions. Each task uses five in-context examples and 20 instances, evaluated with five LLMs. Numbers in parentheses give the number of tasks. Small points show task medians across LLMs, coloured by field, and open circles show the median of each form pooled across tasks, instances and LLMs, with 95\% confidence intervals (CIs) from 2,000 bootstrap resamples of task instances. Bold vertical lines mark the score expected in each condition.
\textbf{c}, Relation scores when $k$ of the five in-context examples are replaced, for 12 tasks (two per form) with ten instances each. Left, \textit{Anonymous} (dashed lines, open points) with $k$ examples supporting (upper branch) or contradicting (lower branch) the Original relation. Right, \textit{Original assignment} (solid lines, filled points) with $k$ examples from the Swapped relation, which contradict the names. Thin lines show individual LLMs and thick lines their median.
\textbf{d}, Schematic of a Bayesian observer at $k = 0$, 2 and 5 of five examples. The prior is set by the domain-specific knowledge elicited by the variable names, the likelihood by the contradicting examples, and the posterior gives the prediction. Curves are illustrative and not fitted to data.
\textbf{e}, Prior weight (left axis) and weighted absolute percentage error (WAPE, right axis) when $k$ of 20 in-context examples contradict the names, for the same 12 tasks with ten instances each. The prior weight is the share of the prediction attributable to the prior rather than to the examples. Points show observed prior weights, and translucent curves show the prior weights predicted by this observer, fitted only to the shaded range ($k\le5$, Methods). Squares show the WAPE relative to the Original relation (median across tasks). Right axes are scaled separately for each LLM.
\textbf{f}, Goodness of fit of the observer. For each LLM, the median relation scores for 12 tasks, two conditions and different values of $k$ were compared with the observer's predictions, and the squared differences were summed into a single value (vertical line). Histograms show the same value for 1,000 datasets simulated from the observer, using the parameters fitted to that LLM and noise of the size observed among its task instances. Dark tails mark the largest 5\% of simulated values, and $P$ is the proportion of simulated values greater than or equal to that of the LLM.}

\ifsubmission\else
\begin{figure}[p]   %
\centering
\includegraphics[width=157.39mm]{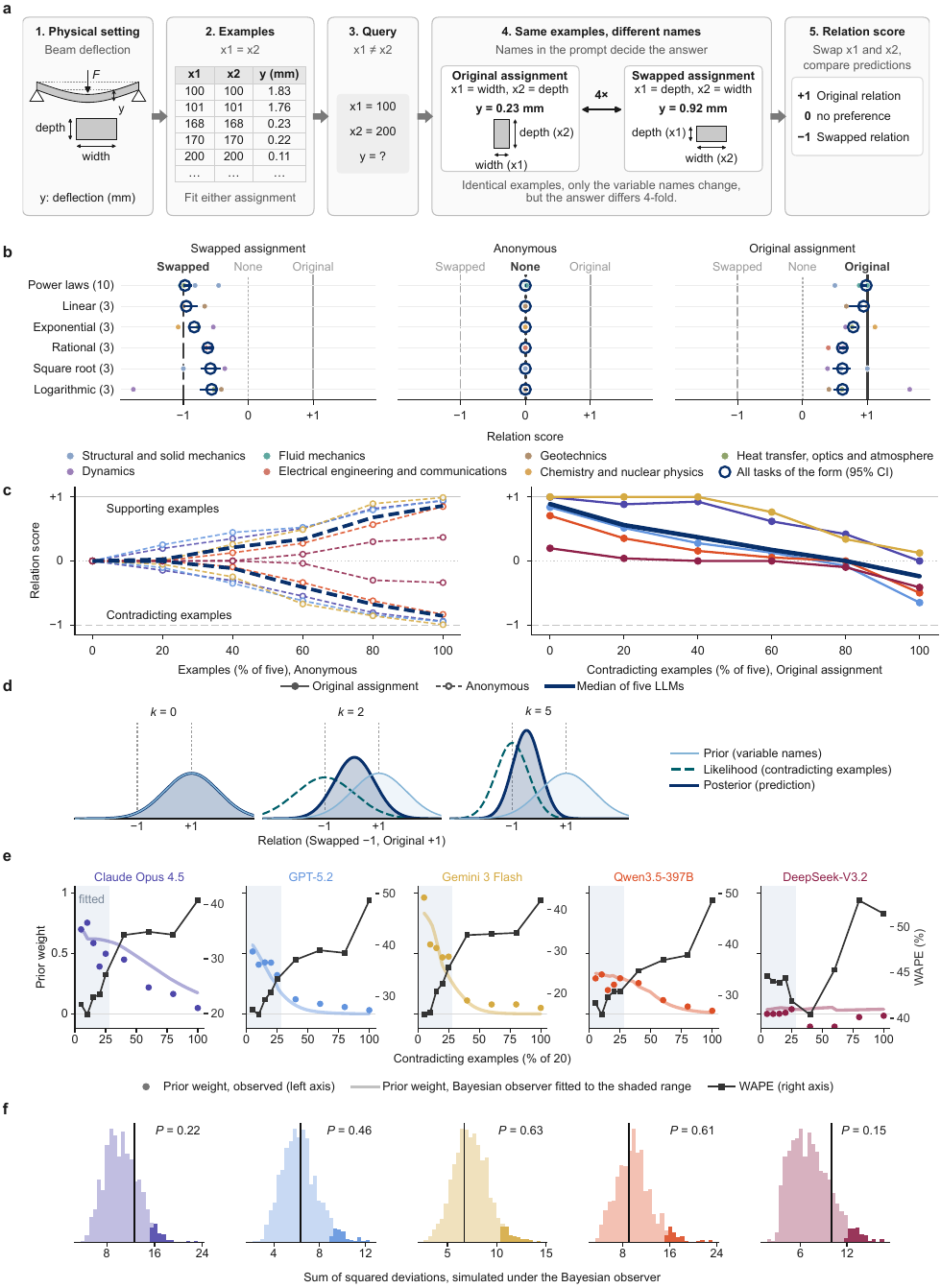}
\end{figure}
\begin{figure}[t]
\caption{\capFigFour}
\label{fig:fig4}
\end{figure}
\fi

\subsection{Pretrained domain-specific knowledge behaves as a prior that in-context examples update}

Having shown that variable names allow LLMs to draw on pretrained domain-specific knowledge, we next asked how this knowledge is integrated with the in-context examples.
In real-world datasets, the variable names and the examples reflect the same relation, so it is unclear whether the relation applied by the LLMs comes from the names or from the examples.
We therefore designed 25 controlled relation tasks, each based on an established formula in which two inputs play different roles.
The tasks covered power-law, linear, exponential, square-root, logarithmic and rational relations (Methods).
We illustrate the design with beam deflection (Fig.~\ref{fig:fig4}a), a power-law relation in which the depth of the cross-section affects the deflection far more strongly than its width.
In the in-context examples, the deflection was computed from the formula with equal depth and width, so the examples could not reveal the roles of the two inputs.
In the prediction inputs, by contrast, the two values differed, so a correct prediction required knowing how each input affects the deflection.
\textit{Original assignment} named the two inputs as in the formula.
To test whether the predictions followed the names, \textit{Swapped assignment} exchanged the two names.
\textit{Anonymous} gave the two inputs neutral descriptions and served as a baseline.
To summarise the predictions, we computed a relation score, which compared the prediction for each input with that obtained after exchanging its two values (Methods). A score of $+1$, $-1$ or 0 indicated that the predictions followed \textit{Original assignment}, \textit{Swapped assignment} or neither.

The predictions of LLMs followed the roles implied by the feature names (Fig.~\ref{fig:fig4}b).
Under \textit{Original assignment}, the median relation scores were positive for every functional form, ranging from 0.61 for rational to 0.98 for power-law relations.
Under \textit{Swapped assignment}, the relation scores consistently flipped to negative: across all 25 tasks, the median score shifted from positive to negative. In 92.1\% of 2,500 paired evaluations (comprising the same task, LLM, and instance), scores were higher under \textit{Original assignment}, with 68.6\% exhibiting the expected sign reversal.
Under \textit{Anonymous}, by contrast, the predictions showed no preference, with a median relation score of 0 for every functional form.
The predicted values also matched the relation implied by the names, with 79.0\% of 29,941 valid predictions closer to the target computed with the assigned names (Supplementary Table~15).
Because the three conditions differed only in how the two inputs were described, the differences between their predictions reflected the domain-specific knowledge elicited by the feature names (Supplementary Note~5 and Supplementary Tables~14 and 15).

We next investigated how predictions were affected by the number of in-context examples that contradict this knowledge.
In 12 randomly selected tasks, two per functional form, we replaced $k$ of the five examples with contradicting ones. In these examples, the two inputs differed, and the target followed the relation implied by \textit{Swapped assignment}. The total number of examples thus remained five, while the proportion contradicting the knowledge increased with $k$ (Fig.~\ref{fig:fig4}c).
Under \textit{Anonymous}, where only the examples could guide the predictions, the median relation score pooled across the five LLMs started at 0 and fell to $-0.86$ at $k = 5$ (lower branch in Fig.~\ref{fig:fig4}c, left).
When the replaced examples instead followed the original roles, the score rose symmetrically to 0.86 (upper branch in Fig.~\ref{fig:fig4}c, left).
Without names, the LLMs therefore had no default preference and followed the examples.
Under \textit{Original assignment}, the predictions started close to the roles implied by the names (0.89 at $k = 0$) and showed a slower shift towards the swapped roles than without names, reaching only $-0.24$ at $k = 5$ compared with $-0.86$ (Fig.~\ref{fig:fig4}c, right).

We attribute this slower shift to the domain-specific knowledge that the LLMs acquired during pretraining.
In this interpretation, the knowledge elicited by the variable names acts as a prior on the predictions and the in-context examples as the likelihood.
The more examples contradict the variable names, the further predictions shift from the prior towards the examples (Fig.~\ref{fig:fig4}d).
To quantify the contribution of the prior, we computed the prior weight, the share of the prediction attributable to the prior rather than to the examples (Methods).
It equals 0 when the names have no effect on the predictions and 1 when the predictions ignore the examples.
The prior weight (pooled across the five LLMs) differed widely across tasks, from 0.17 for Equivalent bending moment to 1.00 for Soil shear strength, whose predictions followed the names even when all five examples contradicted them (Supplementary Table~16).
To examine this weight over a wider range of contradicting examples, we increased the number of examples to 20 and varied the number $k$ contradicting the knowledge from 0 to 20.
As $k$ increased, the prior weight declined towards 0 (Fig.~\ref{fig:fig4}e). The predictions thus moved away from the relation implied by the names, and for the four LLMs other than DeepSeek-V3.2, the WAPE relative to this relation rose from 17--23\% at $k = 0$ to 41--49\% at $k = 20$.
When the examples could not distinguish the two relations ($k = 0$), the LLMs with a larger prior weight predicted more accurately. The WAPE ranged from 17\% for Gemini 3 Flash, which had the largest prior weight in the five-example experiment (Supplementary Table~17), to 46\% for DeepSeek-V3.2, which had the smallest (Fig.~\ref{fig:fig4}e).

To test whether the decline in the prior weight followed Bayesian updating, we fitted a Bayesian observer to the LLM predictions (Methods).
When fitted only to $k\le5$, the observer predicted prior weights at $k=8,12,16,20$. The predicted and observed prior weights at these four held-out values of $k$ were correlated at $r=0.870$--$0.997$ for the four LLMs other than DeepSeek-V3.2 (translucent curves in Fig.~\ref{fig:fig4}e; Supplementary Fig.~11).
When fitted to all values of $k$, the observer was not rejected by a goodness-of-fit test for any of the five LLMs, although it showed the poorest fit for DeepSeek-V3.2 ($P = 0.15$; Fig.~\ref{fig:fig4}f).
For the other four LLMs, the effect of each contradicting example did not differ detectably with versus without variable names, consistent with the observer's assumption that variable names change only the prior (Supplementary Note~5).
DeepSeek-V3.2, by contrast, had a prior weight of only 0.04 (Supplementary Table~17), and the observer fitted to $k \le 5$ deviated from its relation scores at larger $k$ by $-0.40$ on average, on a scale from $-1$ to $+1$ (Supplementary Note~5).
Together, these results indicate that, in four of the five LLMs, the knowledge elicited by variable names behaved as a prior that the in-context examples updated, consistent with Bayesian updating.

\section{Discussion}

Our results show that LLMs can use both variable names and in-context examples for numerical prediction, matching or outperforming supervised ML models across diverse scientific domains, particularly when examples are scarce.
Compared with numerical values alone, adding variable names improved the predictions, most strongly under semantic grounding, the match between the names and the values they describe.
Reasoning traces further showed that variable names allow LLMs to draw on domain-specific knowledge acquired during pretraining.
Using controlled tasks, we also found that four of the five LLMs integrate this knowledge with in-context examples in a way consistent with Bayesian updating. This finding agrees with theoretical accounts of in-context learning as implicit Bayesian inference\cite{xie2022explanation,panwar2024context}.
In this view, the domain-specific knowledge elicited by variable names can be described as the prior, the in-context examples as the likelihood, and the predictions as following the posterior that combines the two.

This view accounts for how performance changed with the number of examples.
Without examples, predictions depend on the prior alone, which specifies the likely form and direction of the relation but not its precise parameters, so performance at $n = 0$ was poor.
With a few examples, the likelihood is still weak, and the prior elicited by the names supplies much of the information needed for prediction. Consistent with this, the improvement from variable names was largest when examples were scarce.
With more examples, the likelihood increasingly determines the posterior, leaving less room for the prior to improve predictions.
Recent work reported that in-context examples suppress the recall of scientific knowledge in LLMs\cite{jang2026incontext}.
Our results suggest that this suppression is consistent with Bayesian updating, in which the contribution of the prior declines as examples accumulate (Figs.~\ref{fig:fig3}a and \ref{fig:fig4}e).

This Bayesian view also links LLMs to meta-learning and to prior-data fitted networks (PFNs) such as TabPFN\cite{hollmann2025accurate}, in which knowledge acquired across tasks acts as a prior for a new task\cite{grant2018recasting,muller2022transformers}.
LLMs and PFNs differ, however, in the source of the prior.
In PFNs, the prior is learned from synthetic data and is independent of what the variables represent.
In LLMs, the prior is domain-specific knowledge acquired during pretraining and elicited by variable names.
All five LLMs drew on this knowledge, and in four of them it behaved as a prior that the in-context examples updated, suggesting that this way of integrating knowledge with examples is not specific to one model or developer.

Beyond its source, the weight of the prior also differed across LLMs, from 0.04 for DeepSeek-V3.2 to 0.81 for Gemini 3 Flash (Supplementary Table~17).
These differences may reflect distinct inductive biases\cite{alampara2026less} arising from the pretraining data or post-training of each LLM.
A larger prior weight improved predictions when the examples could not reveal the roles of the two inputs but reduced the shift towards contradicting examples (Fig.~\ref{fig:fig4}c,e).
Because new observations in scientific research are often few and depart from established knowledge, a strong prior may lead LLMs to discount them.
When such departures are expected, LLMs with a smaller prior weight may therefore be more suitable.

Our study has several limitations.
We evaluated five LLMs and two prompt formats on relatively low-di\-men\-sion\-al datasets with at most 100 in-context examples, and whether the findings extend beyond these settings remains to be established.
Only six datasets were released after the documented training cutoffs of all five LLMs, so contamination cannot be fully excluded for the others.
The controlled relation tasks involved two inputs and simple functional forms, and how knowledge of more complex relations is integrated with the examples requires further investigation.
The reasoning traces were collected with a single, modest reasoning budget (Methods) and describe what LLMs write when reasoning rather than the computation behind the predictions made without reasoning.
Finally, our experiments provide behavioural evidence consistent with Bayesian updating.
Mechanistic analyses are needed to test whether LLMs perform such updating internally and how semantic grounding is represented.

At a theoretical level, our findings are consistent with the view of in-context learning as implicit Bayesian inference and extend it to scientific numerical prediction.
At a practical level, we show that with semantic grounding, LLMs can serve as data-efficient numerical predictors across diverse scientific domains without task-specific training.

\section{Methods}
\label{sec:methods}

\subsection{Datasets}

We evaluated 20 real-world tabular datasets from nine scientific domains. Supplementary Table~18 lists the domain assignments, sample sizes, feature counts and target variables, and Supplementary Table~19 reports the sources, DOIs, licences and release dates.

Of the 20 datasets, six were first released after the training cutoffs of all five LLMs and formed the post-cutoff set (Supplementary Table~19). For each LLM, we took the training cutoff from developer or provider documentation or, where none was documented, the release date of the LLM.

\subsection{Data splitting and sampling}

Five datasets were split by group so that no group contributed rows to both the in-context examples and the test set. Groups were individuals for the two dental morphology datasets, process combinations for the two fused deposition modelling (FDM) datasets and isotherms for Hydrogen Uptake. The remaining 15 datasets were split by row. Supplementary Table~18 gives the group counts.

Sampling was repeated with five random seeds (42, 123, 456, 789 and 1024). For row-based splits, 200 rows, or half of the rows in datasets with fewer than 250 rows, were first reserved for the example pool and excluded from the test set, and 50 test rows were then drawn from the remaining rows. Examples were sampled from all non-test rows. For grouped splits, groups were randomly ordered and added to the test set until it contained at least 50 rows, and all rows of those groups were excluded from the example pool. Within each dataset and seed, a fixed random ordering of eligible example rows produced nested subsets as the number of examples increased.

LLMs and supervised ML models were compared at $n=5$, 10, 25, 50 and 100 examples, and LLM experiments also included $n=0$, with the example section omitted from the prompt. Within each dataset and seed, all models and prompt conditions used the same test rows and, at each $n>0$, the same in-context examples. When fewer than $n$ rows remained after test-set selection, all eligible rows were used for both model groups. This occurred for FRP--Steel Bond, Hydrogen Uptake and the two dental morphology datasets at $n=100$ (Supplementary Table~18).

\subsection{Prompt design}

We compared four prompt conditions, namely \textit{Anonymised}, \textit{Correct variable names}, \textit{+ Semantic guidance} and \textit{+ Causal guidance}. \textit{Anonymised} omitted the feature names and replaced the target name with a generic label. \textit{Correct variable names} restored both feature and target names. \textit{+ Semantic guidance} added to \textit{Correct variable names} a generic instruction to treat the feature names as informative variables rather than anonymous numerical dimensions. \textit{+ Causal guidance} replaced this instruction with one that made the same request and also asked the model to consider functional or causal relationships between the features and the target. Neither instruction supplied task-specific definitions, causal relationships or physical constraints. Numerical examples and their order were identical across conditions.

The features were given in the original column order of the dataset, on one line in each in-context example and one per line in the row to be predicted. In-context examples were listed in their original row order, and missing feature values were omitted from the prompt. Variable names were used as supplied by each dataset. Prompts with variable names presented each feature as a name--value pair, with units where the dataset recorded them, whereas anonymised prompts presented the values without feature names. The model was asked to return a single number, which was extracted by a fixed deterministic rule. Responses without an extractable number were recorded as unparseable. Supplementary Methods provides the exact templates, serialisation rules and extraction rule.

An evaluation cell comprised one dataset, seed, model, number of examples and prompt condition. Across the four prompt conditions, 603{,}120 predictions were requested in 12{,}000 cells. Ten predictions (0.0017\%) were unparseable and were excluded from the metrics of their cells. Including the semantic perturbations, the factorial design and the rescaling ablation (Supplementary Note~3), 754{,}001 predictions were requested in 15{,}000 cells.

\subsection{Semantic perturbations and ablations}

To test whether the contribution of variable names depends on their match to the numerical values, we used \textit{Shuffled names} and \textit{Cross-domain names}. \textit{Shuffled names} reassigned feature names across columns by derangement, so that no feature retained its original name, while the target name remained correct. \textit{Cross-domain names} replaced both feature and target names with names from a dataset in another group of related datasets. Supplementary Methods gives the donor rules.

To test whether the names contribute through the plausible magnitude of each quantity, we used \textit{Rescaled values}, which retained all variable names but multiplied each feature and the target by its own factor. Each column had a single factor, shared by all rows and redrawn for each seed. Each factor was $f=10^{su}$, where $u\sim U(0.70,1.50)$ and $s\in\{-1,+1\}$ with equal probability. This gave $f\in[0.0316,0.2]\cup[5,31.6]$. The same undisclosed factors were applied to the matched \textit{Anonymised} prompts.

To separate the contributions of feature and target names, we added \textit{Target name only}, which omitted the feature names but kept the target name, and \textit{Feature names only}, which kept the feature names but replaced the target name with a generic label. With \textit{Correct variable names} and \textit{Anonymised}, these formed a $2\times2$ factorial design. All perturbation, rescaling and factorial conditions were evaluated at $n=5$ with all five seeds and LLMs.

\subsection{Standardised prompt format}

To check that the effect of feature names did not depend on the prompt layout, we built a standardised prompt format with two conditions, \textit{Anonymised} and \textit{Feature names}. Both presented the features as the codes $x_1$ to $x_k$ in the original column order and supplied the same target name and recorded units. The two conditions differed only in a field-definition block with one line per code. \textit{Anonymised} described each feature by its position, whereas \textit{Feature names} gave its original name. Because the two conditions differed only in the feature names, their difference was compared with the improvement of \textit{Feature names only} over \textit{Anonymised} in the original format.

In both conditions, text values were replaced by codes $v_{j,m}$, the $m$th distinct value of column $j$ in order of first appearance, so that text values could not reveal the meaning of a column and the two conditions presented identical values (Supplementary Methods). The comparison used $n=5$ and the example rows, test rows and inference settings of the main experiment, giving 500 cells per condition. Supplementary Methods gives the template, and Supplementary Note~3 and Supplementary Tables~7 and 8 report the results.

\subsection{Reasoning traces}

To further explain the improvement from variable names, we analysed the reasoning traces from the run in which extended reasoning was enabled (Models and inference). Qwen3.5-397B and DeepSeek-V3.2 return their full reasoning text, whereas Claude Opus 4.5, GPT-5.2 and Gemini 3 Flash return summaries generated by their providers. These summaries are less informative than the full reasoning and were used only for the small maps in Fig.~\ref{fig:traces}a. The word-level and step-by-step analyses used only Qwen3.5-397B and DeepSeek-V3.2. For GPT-5.2, 13\% of responses returned only encrypted reasoning and were excluded from the analysis of traces.

Each trace was lower-cased and stripped of English stop words and the words of the prompt template. Every word occurring in the variable names or text values of its dataset, including plural forms and abbreviations, was also removed, so that only words introduced by the model remained. The remaining text was embedded with the all-MiniLM-L6-v2 sentence encoder\cite{reimers2019sentence,wang2020minilm}, with embeddings averaged over chunks of at most 254 tokens, and projected with t-SNE\cite{vandermaaten2008tsne} using cosine distance. The large map in Fig.~\ref{fig:traces}a used 150 randomly drawn traces per condition and dataset from Qwen3.5-397B and DeepSeek-V3.2 (perplexity 40), and the small maps used 60 per LLM, condition and dataset (perplexity 30). Words distinguishing the two conditions were identified within each domain using the log-odds ratio with an informative Dirichlet prior\cite{monroe2008fightin}, and the words with the largest $z$-scores are shown after excluding a fixed list of generic words (Supplementary Note~4).

To test whether the reasoning draws increasingly on domain concepts as feature names are added (Fig.~\ref{fig:traces}b,c), we reran DeepSeek-V3.2 and Qwen3.5-397B with the same reasoning budget at $n=5$ for seed 42. The prompts used the \textit{Correct variable names} template with 0\%, 25\%, 50\% or 75\% of the feature names. Features without a name were labelled \textit{feature 1} to \textit{feature k} by column position, and the target name was given at every level. The word \textit{feature} was removed from these traces before analysis. A further level without any names also replaced the target name with a generic label. The names were added in two random orders per dataset, nested so that each level kept the names of the previous one, and intermediate levels gave at least one and at most $k-1$ names. Each level and order used the test rows of the main experiment, giving 16{,}112 predictions, and the 100\% level used the traces of the reasoning comparison, whose prompts were identical.

A multinomial logistic regression trained on the embeddings of \textit{Correct variable names} traces from the other four seeds assigned each trace probabilities over the nine domains (Fig.~\ref{fig:traces}b). For each dataset and level, $\bar{p}$ was the mean over traces of $p$, the probability assigned to the trace's own domain. The first plot of Fig.~\ref{fig:traces}c shows Gaussian kernel density estimates of $p$ for individual traces, reflected at 0 and 1. Domain terms were the 20 words per domain with the largest log-odds $z$-scores between \textit{Correct variable names} and \textit{Anonymised} traces from the other four seeds, and their occurrences in each trace were counted after the same word removal. The values of $\bar{p}$ and the number of domain terms per trace in Fig.~\ref{fig:traces}c and Supplementary Table~13 are medians across datasets, and the confidence intervals in Fig.~\ref{fig:traces}c come from 2,000 bootstrap resamples of datasets.

\subsection{Controlled relation tasks}

To test directly whether the knowledge elicited by variable names guides predictions, we constructed 25 controlled relation tasks independent of the 20 datasets. Each task had three conditions. \textit{Original assignment} and \textit{Swapped assignment} were two assignments of the feature names that were both consistent with the same in-context examples. \textit{Anonymous} described the two columns generically without stating which variable each held. Each task used an established relation $y=f(x_1,x_2)$ in which the two inputs play different roles, and \textit{Swapped assignment} exchanged the roles of the two columns, giving $y=f(x_2,x_1)$. Because every example was generated with $x_1=x_2$, both assignments give the same values on the examples and cannot be distinguished from the numerical values alone.

The tasks were chosen so that only the names could distinguish the two relations. Of the 25 tasks, 10 used power laws, $y=Cx_1^{p}x_2^{q}$, admitted only if $p\ne q$, so that exchanging the names changes the prediction, and $p+q\ne 0$, so that the examples are not constant. For these tasks, $x_1=x_2$ also had to be physically admissible and $C$ could not depend on the ratio of the two inputs. The other 15 tasks used linear, exponential, rational, square-root and logarithmic relations, with non-constant examples, a physically admissible $x_1=x_2$ and targets that changed by at least 1.25-fold when the two inputs were exchanged. In all 25, outputs were positive and finite, and both inputs shared a single unit, so that units could not reveal the roles under \textit{Anonymous}. All 25 are standard textbook relations from seven fields and are listed with their constants and units in Supplementary Table~14.

Each task had 20 instances, each with its own set of examples. For power laws, $C$ was drawn on a logarithmic scale for each instance and was not disclosed. For the other relations, constants were fixed across instances. The five examples used the two endpoints of a task-specific range (100 and 200 for power laws) and three distinct values drawn from within it, listed in a random order shared by all conditions. In 10 of the 20 instances of each task, the two columns, with their names and values, were listed in the opposite order, so that a preference for a column position alone would average to zero. Each instance had six prediction inputs, formed from three pairs of unequal values and their exchanges. The three conditions shared the physical setting, examples, prediction inputs, units and output instruction and differed only in the two lines defining the variables. Unlike the dataset prompts, these prompts stated a physical setting to define the task, but no formula, exponent, monotonic direction or worked calculation. Each prediction was requested in an independent session, with conditions interleaved, using the inference settings of the main experiment. The design gave 45{,}000 predictions across five LLMs.

For each pair of exchanged prediction inputs $(u,v)$ and $(v,u)$, we summarised the predictions by the relation score
\begin{equation*}
T=\frac{\log\left[\hat{y}(u,v)/\hat{y}(v,u)\right]}{\log\left[f(u,v)/f(v,u)\right]},
\end{equation*}
where $f$ is the relation under \textit{Original assignment}, so that the denominator equals $(p-q)\log(u/v)$ for power laws. $T$ equals $+1$ when the predictions follow \textit{Original assignment} exactly and $-1$ when they follow \textit{Swapped assignment}. Scores were not truncated to $[-1,1]$. Pairs with a non-positive or unparseable prediction were excluded. For each LLM, task instance and condition, $T$ was the median over the three exchanged pairs. For each LLM and task instance, we recorded whether the relation score was higher under \textit{Original assignment} than under \textit{Swapped assignment}. We also recorded whether the score changed from positive under \textit{Original assignment} to negative under \textit{Swapped assignment}. Supplementary Note~5 reports how closely the predicted values matched the implied targets.

\subsection{Contradicting examples}

To measure how predictions change when the in-context examples contradict the feature names, we randomly selected two controlled relation tasks per functional form, 12 in total (Supplementary Table~16). Each task contributed ten instances, five in each column order. In each instance, $k\in\{1,2,3,4,5\}$ of the five examples with $x_1=x_2$ were replaced by contradicting examples, which had unequal inputs and targets from the Swapped relation. The middle examples were replaced before the two endpoints, so that the range of the examples changed as little as possible, and the replaced and added examples were nested across $k$. In each contradicting example, one input came from the lower 40\% and the other from the upper 40\% of the task range. Its targets under the two assignments differed by at least 1.25-fold, and it did not coincide with a prediction input or its exchange. The prediction inputs, prompt template and inference settings were those of the controlled relation tasks, and $k=0$ used their results. \textit{Original assignment} and \textit{Anonymous} were evaluated, giving 36{,}000 predictions across five LLMs, and relation scores used the Original relation as reference.

To check that predictions under \textit{Anonymous} respond to supporting examples as they do to contradicting ones, \textit{Anonymous} was also evaluated at $k=1$--5 with the same added inputs but with targets from the Original relation, giving 18{,}000 further predictions. Exclusions in both experiments are reported in Supplementary Note~5.

The prior weight, the share of the prediction attributable to the prior rather than to the examples, was $(T_{\mathrm{Original}}-T_{\mathrm{Anonymous}})/(1-T_{\mathrm{Anonymous}})$, computed from the median relation scores at each $k$. It is the fraction of the gap between \textit{Anonymous} and the Original relation ($+1$) that the names close. It therefore equals 0 when the names have no effect on the predictions and 1 when the predictions ignore the examples. In the five-example experiment, the prior weight of each LLM was averaged over $k=1$--5, with 95\% confidence intervals from 2,000 bootstrap resamples of task instances that retained all LLMs for each sampled instance (Supplementary Table~17). The number of contradicting examples without names that reached the same relation score as five with names was obtained by linear interpolation of the \textit{Anonymous} medians.

In an extension, each prompt contained 20 in-context examples. Each instance received 15 further examples with $x_1=x_2$, drawn without replacement from the grid of input values of its task and excluding the prediction inputs. Then $k\in\{0,1,2,3,4,5,8,12,16,20\}$ of the 20 examples were replaced by contradicting examples generated as above, giving 72{,}000 predictions across five LLMs. Prior weights were computed at each $k$ as above. WAPE under \textit{Original assignment} was computed against the targets of the Original relation, pooled over the instances and prediction inputs of each task, and summarised as the median across tasks.

We modelled the relation scores with a Bayesian observer that treated the two assignments as hypotheses. With prior log odds $a_t$ for the Original relation in task $t$ and log likelihood ratio $b_t$ per contradicting example, the posterior probability of the Original relation after $k$ contradicting examples is $\sigma(a_t-kb_t)$, where $\sigma$ is the logistic function. The predicted relation score is $2\sigma(a_t-kb_t)-1$, and $a_t=0$ under \textit{Anonymous}. This conversion assumes that each prediction mixes the two relations in log space, $\log\hat{y}(u,v)=\pi\log f(u,v)+(1-\pi)\log f(v,u)$, where $\pi$ is the posterior probability of the Original relation, which gives $T=2\pi-1$. Mixing in linear space gave nearly identical posterior probabilities ($r=0.97$ across 11{,}992 instances), so the choice between the two mixing rules has little effect on the fitted observer (Supplementary Note~5). For each LLM and task, $a_t$ and $b_t$ were fitted by least squares to the median relation scores of both conditions, with $b_t$ shared between them and $a_t=0$ under \textit{Anonymous}, so that the knowledge elicited by the names could change only the prior. We assessed the observer in two ways. A parametric-bootstrap goodness-of-fit test compared the fitted observer with the data\cite{wichmann2001psychometric}. Its statistic was the sum of squared differences between the median relation scores and the observer's predictions. To simulate data from a correct observer, we took the deviation of each task instance from the median relation score at every $k$ under both conditions. We then generated 1,000 simulated datasets, each by resampling task instances with replacement and adding their deviations to the observer's predictions. The deviations of one instance stayed together at all values of $k$ in both conditions, preserving their correlation across $k$. The observer was refitted to each simulated dataset to give the distribution of the statistic. $P$ was the proportion of simulated statistics at least as large as the observed one. It was computed for each LLM, after summing the statistic over its 12 tasks, and for each task, with Holm correction across the 12 tasks of each LLM. An out-of-sample test fitted $a_t$ and $b_t$ to both conditions at $k\le5$ only and predicted both conditions at $k=8$--20. Predicted prior weights were computed from the median scores predicted by this fit across tasks. The fitted curves and the prior and likelihood ratio of each LLM are reported in Supplementary Note~5 and Supplementary Figs.~9 and 10.

\subsection{Models and inference}

We evaluated Claude Opus 4.5\cite{anthropic2025opus45}, GPT-5.2\cite{openai2026gpt52}, Gemini 3 Flash\cite{google2026gemini3flash}, Qwen3.5-397B\cite{qwen2026qwen35} and DeepSeek-V3.2\cite{deepseek2025v32} through OpenRouter without parameter updates. The five LLMs came from five developers and included three proprietary and two open-weight models. Temperature was set to 0, and each test row was predicted once per evaluation cell. Extended reasoning, web search and other tools were disabled. To collect reasoning traces, a separate run enabled reasoning for \textit{Anonymised} and \textit{Correct variable names} at $n=5$ with all five seeds, using a budget of 2,000 reasoning tokens. GPT-5.2 accepts only effort levels and used low effort. Supplementary Table~11 reports its accuracy and cost by LLM, and Supplementary Table~20 lists the resolved model identifiers and generation settings.

We evaluated seven supervised ML models on the same examples, covering linear (ridge regression and ordinary least squares), kernel (Gaussian process), tree-based (tuned random forest and tuned XGBoost) and pretrained tabular (TabPFN and CARTE) approaches. The main text reports the three with the highest $R^2$ at every number of examples, namely TabPFN\cite{hollmann2025accurate}, tuned XGBoost\cite{chen2016xgboost} and tuned random forest\cite{breiman2001random}. TabPFN used implementation version 8.3.0 and the TabPFN-3 regression checkpoint\cite{grinsztajn2026tabpfn3}, predicting from labelled examples without task-specific parameter updates. XGBoost and random forest were tuned with Optuna\cite{akiba2019optuna} using 50 trials for each dataset, seed and number of examples. The objective was mean squared error from shuffled $K$-fold cross-validation within the training set, with $K=\min(5,n)$. The software environment used XGBoost 3.2.0, scikit-learn 1.8.0\cite{pedregosa2011scikit} and Optuna 4.8.0, and Supplementary Table~21 gives the search spaces and tuning procedures. For all supervised ML models, features with text values were label-encoded and features were standardised using training-set statistics.

Ridge regression\cite{hoerl1970ridge}, Gaussian process regression\cite{rasmussen2006gaussian} and ordinary least squares were included as three additional supervised ML models without Optuna tuning. For these three models, missing feature values were imputed with training-set means. Supplementary Table~21 gives their fitting procedures.

The seventh supervised ML model was CARTE\cite{kim2024carte}, which takes variable names as input and was fitted with and without them. Supplementary Table~21 gives its settings, and Supplementary Note~2 and Supplementary Table~5 report its results.

\subsection{Metrics and aggregation}

Performance was measured with $R^2=1-\mathrm{SS}_{\mathrm{res}}/\mathrm{SS}_{\mathrm{tot}}$, which can be negative, and $\mathrm{WAPE}=\sum_i|y_i-\hat{y}_i|/\sum_i|y_i|\times100\%$, where $y_i$ and $\hat{y}_i$ are the observed and predicted targets. Both were computed from the test predictions of each evaluation cell.

Results were summarised in two stages so that each dataset had equal weight. For each model, we took the median across the five seeds within each dataset and then the median across the 20 datasets. For a group of models, such as the five LLMs or the three best-performing supervised ML models, the first stage pooled the seeds and models of each dataset, so a group summary is not the median of the individual-model summaries.

Prompt conditions were summarised in the same way, pooling the seeds and LLMs of each dataset (Figs.~\ref{fig:fig3}a and \ref{fig:names}b,d). The improvement of one condition over another was the difference between their summaries. In Fig.~\ref{fig:fig3}b, each cell gives this difference for one dataset and LLM at $n=5$, using medians over seeds, and the rightmost column gives it for each LLM across datasets.

Figure~\ref{fig:names}e summarises contrasts between naming conditions and between perturbation conditions. The feature-name and target-name paths are the improvements of \textit{Feature names only} and \textit{Target name only} over \textit{Anonymised}, and the target-identity path is the difference between \textit{Correct variable names} and \textit{Feature names only}. The perturbation contrasts give the algebraic decomposition $(C-A)=[(C-A)-(R-A_R)]+[(R-A_R)-(S-A)]+(S-A)$, where $C$, $R$ and $S$ are the condition medians of \textit{Correct variable names}, \textit{Rescaled values} and \textit{Shuffled names}, and $A$ and $A_R$ are those of \textit{Anonymised} without and with rescaling. Numerical magnitude denotes the contrast between \textit{Correct variable names} and \textit{Rescaled values}, and variable identity denotes the contrast between \textit{Rescaled values} and \textit{Shuffled names}. The final term is the improvement that persisted under \textit{Shuffled names}.

\subsection{Statistics}

We used two kinds of 95\% confidence interval. For the median of a single condition, the interval ran from the sixth to the fifteenth of the 20 dataset-level values ranked in order, which covers the median with at least 95\% probability without assuming any distribution (Figs.~\ref{fig:fig1}b, \ref{fig:fig3}a, left, and \ref{fig:names}b). For the difference between two conditions, we resampled the 20 datasets 2,000 times with replacement, keeping all seeds, models and conditions of each dataset together, and took the 2.5th and 97.5th percentiles of the recalculated difference (Figs.~\ref{fig:fig3}a, right, and \ref{fig:names}a,c)\cite{efron1979bootstrap}. Intervals for individual LLMs used the same procedure. Paired tests are described in Supplementary Methods.

\section{Data availability}

The 20 evaluation datasets fall into two groups. Eleven have not previously been made public. Eight engineering datasets are original resources of the authors' research group, and three, Carbon Intensity, Drug--Polymer Solubility and Drug--Water Solubility, were provided by collaborators. All eleven are deposited with the paper under a CC BY 4.0 licence.
The other nine derive from public sources: Formation Enthalpy from the supporting-information database of Yalamanchi \emph{et al.}\cite{yalamanchi2019enthalpy}, the two dental morphology datasets from a Zenodo deposit, and the six post-cutoff datasets from five public sources. Their DOIs or access links, licences and first-release dates are given in Supplementary Table~19. For the post-cutoff datasets, the first-release date is the timestamp of the deposit on Mendeley Data or on the data provider's open-data portal.
Formation Enthalpy and the three urban-mobility datasets are not redistributed, in line with the terms of their sources, and code in the Supplementary Software rebuilds them from the original releases. These four datasets are provided to editors and reviewers in the confidential submission package.
Source data for Figs.~\ref{fig:fig1} to \ref{fig:fig4} are provided with this paper, together with the
performance metrics of every evaluation cell. Figure~\ref{fig:overview} is a schematic overview and has no source data.

\section{Code availability}
The code used to run the experiments and to compute the metrics and statistics is provided as Supplementary Software with this paper. The accompanying data package contains historical analysis input data, and the software documentation clearly distinguishes between `recalculating results' and `rerunning model inference'. It also specifies the software dependencies, model checkpoint and commands for offline numerical validation. Rerunning the LLM experiments requires access to the specified models and API keys, which are not included in the package.
\newpage
\bibliographystyle{naturemag}
\bibliography{references}

\section{Acknowledgements}

We thank Prof.~Naixu Li and Prof.~Yuanhui Ji (School of Chemistry and Chemical Engineering, Southeast University) and Prof.~Mingzhu Zheng (School of Medicine, Southeast University) for providing data and for their support.
We also thank Prof.~Jingfeng Yuan (School of Civil Engineering, Southeast University) and Prof.~Linfeng Zhang (School of Transportation, Southeast University) for their support.
This work builds on methodological ideas from the monograph \textit{Machine Learning--Enabled Intelligent Structural Computing} by Prof.~D. Feng, Prof.~S. Chen and Prof.~G. Wu (Southeast University Press, in Chinese).

\section{Author contributions}

X.M., S.C. and D.F. conceived the study and designed the experiments. X.M.
implemented the experimental pipeline, conducted the evaluations, performed
the comparisons with supervised ML models and produced the figures. H.J. performed data
verification. D.F. curated datasets and contributed domain expertise. S.C. and D.F.
supervised the project. M.B. provided critical feedback on the manuscript.
X.M. wrote the initial draft. All authors
analysed the results, discussed the implications and contributed to manuscript
preparation.

\section{Competing interests}

The authors declare no competing interests.

\end{document}


\maketitle
\thispagestyle{empty}

\clearpage   %

{\footnotesize
\hypersetup{linkcolor=black}
\setlength{\tabcolsep}{5pt}
\renewcommand{\arraystretch}{1.0}
\begin{tabular}{@{}l L{13.6cm} r@{}}
\toprule
Item & Title & Page \\
\midrule
\multicolumn{3}{@{}l@{}}{\hyperref[si:note1]{\textbf{Note 1} $\vert$ \textbf{LLMs match or outperform supervised ML models across scientific domains}} \hfill \pageref{si:note1}} \\
\hyperref[si:fig1]{Figure 1} & \hyperref[si:fig1]{Performance on the post-cutoff datasets at $n=100$} & \pageref{si:fig1} \\
\hyperref[si:tab1]{Table 1} & \hyperref[si:tab1]{Method-level performance on the post-cutoff datasets at $n=100$} & \pageref{si:tab1} \\
\hyperref[si:fig2]{Figure 2} & \hyperref[si:fig2]{$R^2$ of each dataset across numbers of examples} & \pageref{si:fig2} \\
\hyperref[si:fig3]{Figure 3} & \hyperref[si:fig3]{WAPE of each dataset across numbers of examples} & \pageref{si:fig3} \\
\hyperref[si:tab2]{Table 2} & \hyperref[si:tab2]{$R^2$ of every method across numbers of examples} & \pageref{si:tab2} \\
\hyperref[si:tab3]{Table 3} & \hyperref[si:tab3]{WAPE (\%) of every method across numbers of examples} & \pageref{si:tab3} \\
\hyperref[si:fig4]{Figure 4} & \hyperref[si:fig4]{Differences between LLMs and supervised ML models for each dataset and number of examples} & \pageref{si:fig4} \\
\hyperref[si:tab4]{Table 4} & \hyperref[si:tab4]{Differences between LLMs and supervised ML models for each dataset at $n=5$ and $n=100$} & \pageref{si:tab4} \\
\midrule
\multicolumn{3}{@{}l@{}}{\hyperref[si:note2]{\textbf{Note 2} $\vert$ \textbf{LLMs rely on variable names for numerical prediction}} \hfill \pageref{si:note2}} \\
\hyperref[si:tab5]{Table 5} & \hyperref[si:tab5]{Effect of variable names on CARTE and the LLMs across numbers of examples} & \pageref{si:tab5} \\
\hyperref[si:fig5]{Figure 5} & \hyperref[si:fig5]{Effect of variable names on each dataset at $n=5$} & \pageref{si:fig5} \\
\midrule
\multicolumn{3}{@{}l@{}}{\hyperref[si:note3]{\textbf{Note 3} $\vert$ \textbf{Variable names improve predictions most when they match the values they describe}} \hfill \pageref{si:note3}} \\
\hyperref[si:fig6]{Figure 6} & \hyperref[si:fig6]{Effects of feature names, the target name and rescaled values at $n=5$} & \pageref{si:fig6} \\
\hyperref[si:tab6]{Table 6} & \hyperref[si:tab6]{Feature-name and target-name contributions at $n=5$} & \pageref{si:tab6} \\
\hyperref[si:tab7]{Table 7} & \hyperref[si:tab7]{Improvement from feature names for each dataset in the standardised prompt format at $n=5$} & \pageref{si:tab7} \\
\hyperref[si:tab8]{Table 8} & \hyperref[si:tab8]{Improvement from feature names and from text values in the nine datasets with text values at $n=5$} & \pageref{si:tab8} \\
\hyperref[si:fig7]{Figure 7} & \hyperref[si:fig7]{Effects of perturbing variable names at $n=5$} & \pageref{si:fig7} \\
\hyperref[si:fig8]{Figure 8} & \hyperref[si:fig8]{Prompt conditions and the improvement from variable names across numbers of examples} & \pageref{si:fig8} \\
\hyperref[si:tab9]{Table 9} & \hyperref[si:tab9]{$R^2$ under the four prompt conditions} & \pageref{si:tab9} \\
\hyperref[si:tab10]{Table 10} & \hyperref[si:tab10]{WAPE (\%) under the four prompt conditions} & \pageref{si:tab10} \\
\midrule
\multicolumn{3}{@{}l@{}}{\hyperref[si:note4]{\textbf{Note 4} $\vert$ \textbf{Variable names allow LLMs to draw on pretrained domain-specific knowledge}} \hfill \pageref{si:note4}} \\
\hyperref[si:tab11]{Table 11} & \hyperref[si:tab11]{Extended reasoning at $n=5$, by LLM} & \pageref{si:tab11} \\
\hyperref[si:tab12]{Table 12} & \hyperref[si:tab12]{Words most characteristic of the reasoning traces with and without variable names} & \pageref{si:tab12} \\
\hyperref[si:tab13]{Table 13} & \hyperref[si:tab13]{Reasoning traces of DeepSeek-V3.2 and Qwen3.5-397B as feature names are added at $n=5$} & \pageref{si:tab13} \\
\midrule
\multicolumn{3}{@{}l@{}}{\hyperref[si:note5]{\textbf{Note 5} $\vert$ \textbf{Pretrained domain-specific knowledge behaves as a prior that in-context examples update}} \hfill \pageref{si:note5}} \\
\hyperref[si:tab14]{Table 14} & \hyperref[si:tab14]{Controlled relation tasks, by task} & \pageref{si:tab14} \\
\hyperref[si:tab15]{Table 15} & \hyperref[si:tab15]{Controlled relation tasks, by LLM} & \pageref{si:tab15} \\
\hyperref[si:tab16]{Table 16} & \hyperref[si:tab16]{Contradicting examples, by task} & \pageref{si:tab16} \\
\hyperref[si:tab17]{Table 17} & \hyperref[si:tab17]{Contradicting examples, by LLM} & \pageref{si:tab17} \\
\hyperref[si:fig9]{Figure 9} & \hyperref[si:fig9]{Relation scores with 20 in-context examples and the Bayesian observer} & \pageref{si:fig9} \\
\hyperref[si:fig10]{Figure 10} & \hyperref[si:fig10]{Prior and likelihood ratio of each LLM} & \pageref{si:fig10} \\
\hyperref[si:fig11]{Figure 11} & \hyperref[si:fig11]{Out-of-sample prediction of the Bayesian observer} & \pageref{si:fig11} \\
\midrule
\multicolumn{3}{@{}l@{}}{\hyperref[si:methods]{\textbf{Supplementary Methods}} \hfill \pageref{si:methods}} \\
\hyperref[si:tab18]{Table 18} & \hyperref[si:tab18]{The 20 real-world tabular datasets} & \pageref{si:tab18} \\
\hyperref[si:tab19]{Table 19} & \hyperref[si:tab19]{Provenance of the 20 datasets} & \pageref{si:tab19} \\
\hyperref[si:tab20]{Table 20} & \hyperref[si:tab20]{Inference settings for the five evaluated LLMs} & \pageref{si:tab20} \\
\hyperref[si:tab21]{Table 21} & \hyperref[si:tab21]{Hyperparameter search spaces and tuning procedures for the supervised ML models} & \pageref{si:tab21} \\
\midrule
\multicolumn{3}{@{}l@{}}{\hyperref[si:statistics]{\textbf{Supplementary Statistical Results}} \hfill \pageref{si:statistics}} \\
\hyperref[si:tab22]{Table 22} & \hyperref[si:tab22]{Paired comparisons} & \pageref{si:tab22} \\
\midrule
\multicolumn{3}{@{}l@{}}{\hyperref[si:references]{\textbf{Supplementary References}} \hfill \pageref{si:references}} \\
\bottomrule
\end{tabular}\par}

\clearpage

\section*{Supplementary Note 1 $\vert$ LLMs match or outperform supervised ML models across scientific domains}
\label{si:note1}

This note supports Fig.~2 of the main text and reports performance on the post-cutoff datasets and the advantage of the large language models (LLMs) across numbers of examples.

\subsection*{Performance of LLMs on datasets released after the training cutoffs}
\label{si:postcutoff}

On the six post-cutoff datasets at $n=100$, the five LLMs performed comparably to the three best-performing supervised machine learning (ML) models, suggesting that memorisation of the datasets alone does not explain their performance. Three of these datasets were experimental datasets first published after the cutoffs. The other three were urban-mobility datasets built from June 2026 records of bike-share and bus services that have published data for years (Supplementary Fig.~\sifign{1}a). The LLMs reached $R^2$ of 0.761 to 0.902 and the supervised ML models 0.828 to 0.851. The weighted absolute percentage errors (WAPE) were 14.8\% to 21.3\% for the LLMs and 18.8\% to 22.3\% for the supervised ML models (Supplementary Fig.~\sifign{1}b,c and \sitab{1}).

\begin{figure}[H]
  \centering
  \includegraphics[width=172.66mm]{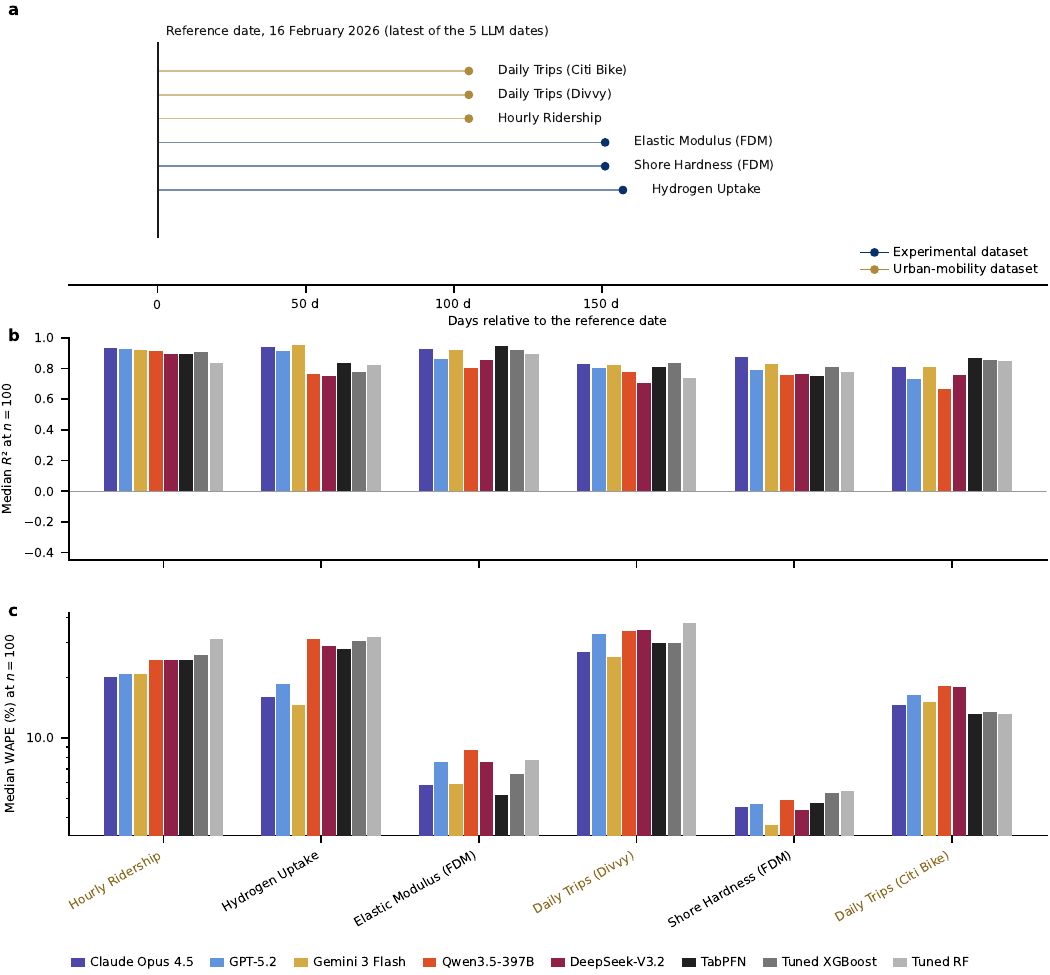}
  \caption{\textbf{\boldmath Performance on the post-cutoff datasets at $n=100$.}
  \textbf{a}, Release dates of the six datasets relative to the latest cutoff or release date of the five LLMs, 16 February 2026 (\sitab{19}). Blue marks the experimental datasets, and gold marks the urban-mobility datasets.
  \textbf{b}, $R^2$ for each dataset and method.
  \textbf{c}, WAPE on a logarithmic scale. Bars are medians over five seeds, and medians across datasets are given in \sitab{1}. RF denotes random forest.}
  \label{fig:si_temporal}\label{si:fig1}
\end{figure}

\clearpage

\begin{table}[H]
\centering
\scriptsize
\caption*{\textbf{\boldmath Supplementary Table 1 $\vert$ Method-level performance on the post-cutoff datasets at $n=100$.}}
\label{si:tab1}
\begin{tabular}{@{}lrr@{}}
\toprule
Method & All six datasets & Three experimental datasets \\
\midrule
\multicolumn{3}{@{}l}{\textit{$R^2$ at $n=100$}} \\
\quad Claude Opus 4.5 & $0.902 \pm 0.006$ & $0.930 \pm 0.012$ \\
\quad GPT-5.2 & $0.833 \pm 0.058$ & $0.863 \pm 0.047$ \\
\quad Gemini 3 Flash & $0.877 \pm 0.027$ & $0.922 \pm 0.023$ \\
\quad Qwen3.5-397B & $0.770 \pm 0.081$ & $0.762 \pm 0.049$ \\
\quad DeepSeek-V3.2 & $0.761 \pm 0.034$ & $0.766 \pm 0.053$ \\
\quad TabPFN & $0.851 \pm 0.021$ & $0.837 \pm 0.060$ \\
\quad Tuned XGBoost & $0.847 \pm 0.035$ & $0.808 \pm 0.055$ \\
\quad Tuned RF & $0.828 \pm 0.030$ & $0.821 \pm 0.036$ \\
\midrule
\multicolumn{3}{@{}l}{\textit{WAPE (\%) at $n=100$}} \\
\quad Claude Opus 4.5 & $15.3 \pm 1.5$ & $5.8 \pm 0.7$ \\
\quad GPT-5.2 & $17.5 \pm 1.6$ & $7.6 \pm 0.3$ \\
\quad Gemini 3 Flash & $14.8 \pm 2.2$ & $5.9 \pm 0.6$ \\
\quad Qwen3.5-397B & $21.3 \pm 1.4$ & $8.7 \pm 1.1$ \\
\quad DeepSeek-V3.2 & $21.3 \pm 2.5$ & $7.6 \pm 1.2$ \\
\quad TabPFN & $18.8 \pm 4.3$ & $5.2 \pm 0.7$ \\
\quad Tuned XGBoost & $19.7 \pm 1.9$ & $6.6 \pm 1.8$ \\
\quad Tuned RF & $22.3 \pm 2.7$ & $7.7 \pm 1.0$ \\
\bottomrule
\end{tabular}

\vspace{0.4em}
\begin{minipage}{\linewidth}
{\footnotesize Each value is the median across datasets, taken over all six post-cutoff datasets in the first column and over the three experimental datasets in the second. Medians are computed in two stages, first across seeds within each dataset and then across datasets (Methods). The value after $\pm$ is the standard deviation (s.d.) of this median across the five seeds.}
\end{minipage}
\end{table}

\clearpage

\subsection*{Performance of LLMs across numbers of examples}
\label{si:context}

Supplementary Figs.~\sifign{2} and \sifign{3} show the $R^2$ and WAPE of each dataset from $n=5$ to $n=100$, and Supplementary Tables~\sitabn{2} and \sitabn{3} summarise each method, including the LLMs at $n=0$. At $n=0$, the predictions of the LLMs were not merely on the wrong scale but poorly correlated with the targets. The squared correlation between predictions and targets, which equals $R^2$ after optimal linear rescaling, was only 0.107 (Supplementary Table~\sitabn{2}b).

The advantage of the LLMs was largest with few examples. At $n=5$, they had higher $R^2$ on 17 of the 20 datasets in comparisons paired by seed (Supplementary Fig.~\sifign{4}). By $n=100$, the differences were close to zero on most datasets (\sitab{4}).

\begin{figure}[H]
  \centering
  \includegraphics[width=174.26mm]{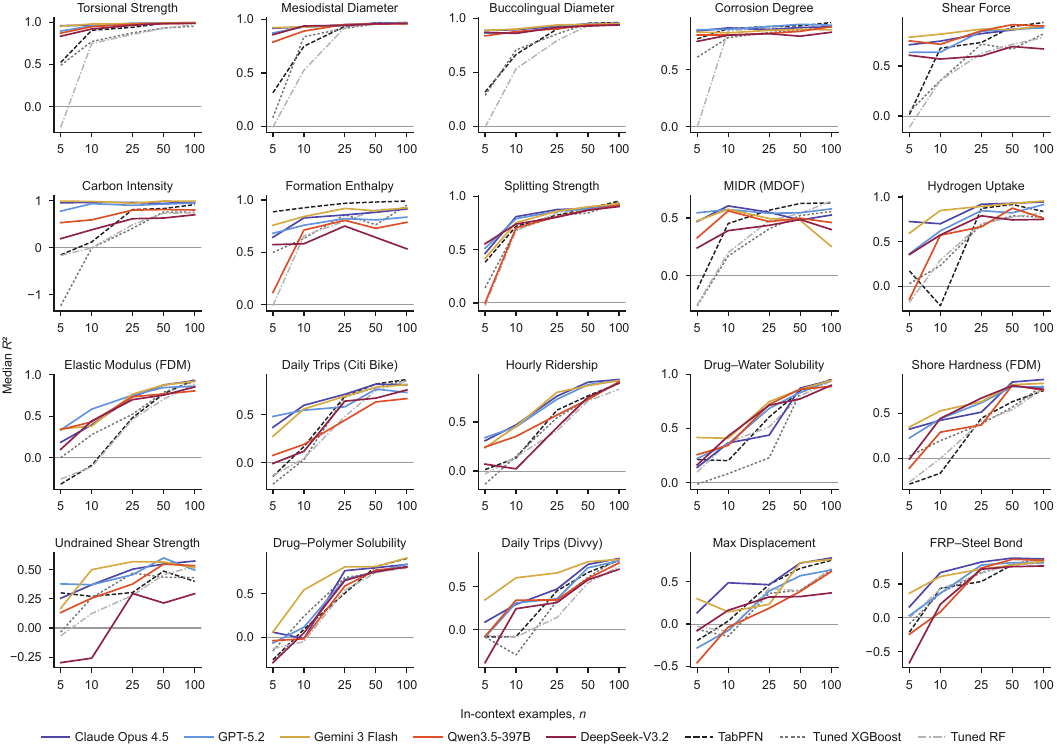}
  \caption{\textbf{\boldmath $R^2$ of each dataset across numbers of examples.}
  Curves show the five LLMs and the three best-performing supervised ML models, and points are medians over five seeds. Vertical scales differ between panels.}
  \label{fig:si_curve_r2}\label{si:fig2}
\end{figure}

\clearpage

\begin{figure}[H]
\centering
\includegraphics[width=174.26mm]{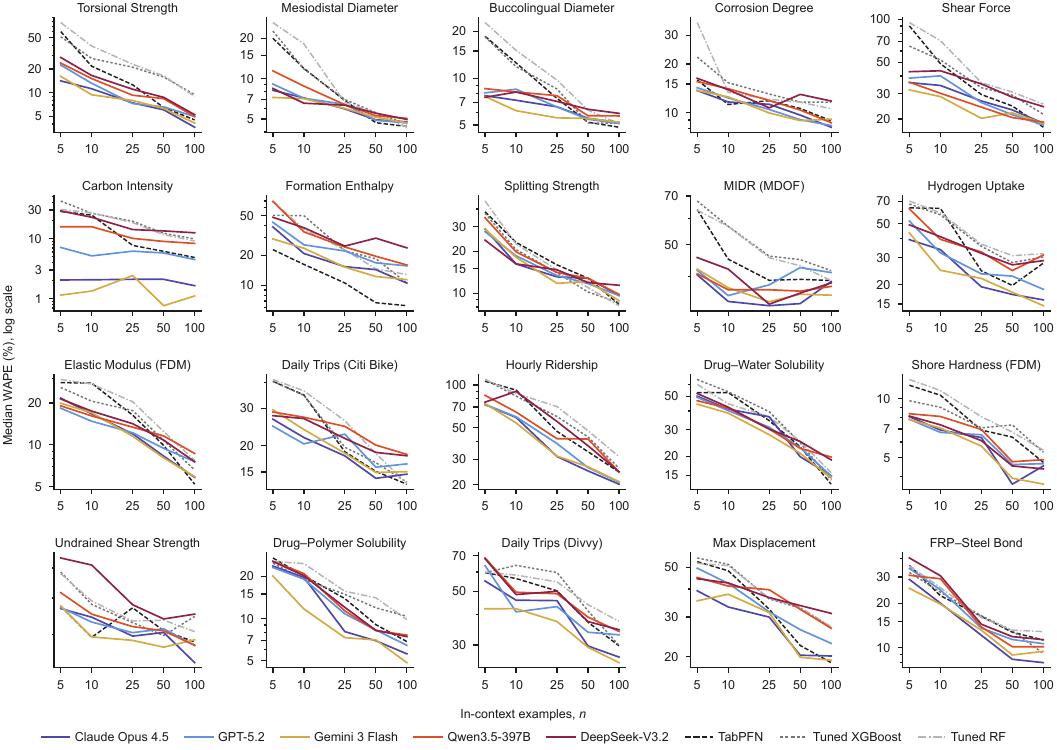}
\caption{\textbf{WAPE of each dataset across numbers of examples.}
As Supplementary Fig.~\sifign{2}, with logarithmic vertical scales.}
\label{fig:si_curve_wape}\label{si:fig3}
\end{figure}

\clearpage

\begin{table}[H]
\centering
\scriptsize
\caption*{\textbf{\boldmath Supplementary Table 2 $\vert$ $R^2$ of every method across numbers of examples.}}
\label{si:tab2}
\begin{tabular}{@{}lrrrrrr@{}}
\toprule
\multicolumn{7}{@{}l}{\textbf{a}, Method-level performance} \\
Method & $n=0$ & $n=5$ & $n=10$ & $n=25$ & $n=50$ & $n=100$ \\
\midrule
Claude Opus 4.5 & $-0.286 \pm 0.098$ & $0.415 \pm 0.059$ & $0.638 \pm 0.044$ & $0.794 \pm 0.024$ & $0.875 \pm 0.019$ & $0.907 \pm 0.009$ \\
GPT-5.2 & $-1.841 \pm 0.641$ & $0.429 \pm 0.098$ & $0.580 \pm 0.058$ & $0.769 \pm 0.024$ & $0.828 \pm 0.031$ & $0.871 \pm 0.020$ \\
Gemini 3 Flash & $-0.630 \pm 0.132$ & $0.417 \pm 0.065$ & $0.609 \pm 0.068$ & $0.806 \pm 0.027$ & $0.863 \pm 0.020$ & $0.917 \pm 0.012$ \\
Qwen3.5-397B & $-0.272 \pm 0.060$ & $0.189 \pm 0.072$ & $0.497 \pm 0.079$ & $0.728 \pm 0.017$ & $0.809 \pm 0.043$ & $0.805 \pm 0.017$ \\
DeepSeek-V3.2 & $-1.085 \pm 0.094$ & $0.177 \pm 0.156$ & $0.436 \pm 0.065$ & $0.684 \pm 0.082$ & $0.754 \pm 0.018$ & $0.767 \pm 0.021$ \\
Median of five LLMs & $-0.335 \pm 0.068$ & $0.346 \pm 0.071$ & $0.547 \pm 0.053$ & $0.745 \pm 0.036$ & $0.849 \pm 0.027$ & $0.865 \pm 0.008$ \\
\midrule
TabPFN & --- & $0.015 \pm 0.137$ & $0.236 \pm 0.123$ & $0.657 \pm 0.090$ & $0.825 \pm 0.038$ & $0.908 \pm 0.017$ \\
Tuned XGBoost & --- & $-0.008 \pm 0.055$ & $0.250 \pm 0.123$ & $0.677 \pm 0.091$ & $0.773 \pm 0.027$ & $0.865 \pm 0.009$ \\
Tuned RF & --- & $-0.066 \pm 0.055$ & $0.247 \pm 0.038$ & $0.528 \pm 0.055$ & $0.747 \pm 0.017$ & $0.828 \pm 0.017$ \\
\shortstack[l]{Median of three\\supervised ML models} & --- & $-0.081 \pm 0.049$ & $0.227 \pm 0.070$ & $0.605 \pm 0.078$ & $0.777 \pm 0.024$ & $0.846 \pm 0.008$ \\
\midrule
Ridge & --- & $-0.113 \pm 0.161$ & $0.037 \pm 0.073$ & $0.386 \pm 0.046$ & $0.509 \pm 0.038$ & $0.532 \pm 0.034$ \\
Gaussian process & --- & $-0.077 \pm 0.065$ & $0.054 \pm 0.058$ & $0.460 \pm 0.086$ & $0.697 \pm 0.040$ & $0.801 \pm 0.046$ \\
Ordinary least squares & --- & $-2.643 \pm 1.185$ & $-0.540 \pm 1.023$ & $0.130 \pm 0.182$ & $0.423 \pm 0.024$ & $0.539 \pm 0.029$ \\
\midrule
CARTE & --- & $-0.165 \pm 0.086$ & $-0.068 \pm 0.143$ & $0.263 \pm 0.066$ & $0.515 \pm 0.130$ & $0.719 \pm 0.063$ \\
\bottomrule
\end{tabular}
\par\vspace{1.6em}
\begin{tabular}{@{}lrr@{}}
\toprule
\multicolumn{3}{@{}l}{\textbf{b}, Zero-example diagnostic} \\
Summary & $n=0$ & $n=100$ \\
\midrule
Median $R^2$ across the five LLMs & $-0.335$ & $0.865$ \\
Datasets with median $R^2<0$ & 16/20 & 0/20 \\
Median squared correlation after optimal linear rescaling & $0.107$ & $0.877$ \\
Prediction vectors with defined squared correlation & 497/500 & 500/500 \\
\bottomrule
\end{tabular}

\vspace{0.4em}
\begin{minipage}{\linewidth}
{\footnotesize \textbf{a}, Each value is the median across the 20 datasets of the median across the five seeds within each dataset (Methods). The value after $\pm$ is the s.d. of this median across the five seeds. LLMs use \textit{Correct variable names}. The rows for the five LLMs and the three supervised ML models pool the seeds and models of each dataset before taking the median across datasets, so they are not medians of the rows above them. \textbf{b}, Median $R^2$ of the five LLMs and median squared Pearson correlation between predictions and targets, which equals $R^2$ after optimal linear rescaling, at $n=0$ and $n=100$. Three of the 500 cells at $n=0$ had constant predictions, for which the correlation is undefined.}
\end{minipage}
\end{table}

\clearpage

\begin{table}[H]
\centering
\scriptsize
\caption*{\textbf{\boldmath Supplementary Table 3 $\vert$ WAPE (\%) of every method across numbers of examples.}}
\label{si:tab3}
\begin{tabular}{@{}lrrrrrr@{}}
\toprule
Method & $n=0$ & $n=5$ & $n=10$ & $n=25$ & $n=50$ & $n=100$ \\
\midrule
Claude Opus 4.5 & $62.1 \pm 4.5$ & $28.9 \pm 3.4$ & $20.4 \pm 1.7$ & $14.3 \pm 1.6$ & $13.4 \pm 2.6$ & $10.1 \pm 1.6$ \\
GPT-5.2 & $78.7 \pm 2.4$ & $31.0 \pm 4.3$ & $22.5 \pm 4.1$ & $17.8 \pm 1.6$ & $13.8 \pm 1.2$ & $12.8 \pm 0.9$ \\
Gemini 3 Flash & $59.1 \pm 3.1$ & $29.0 \pm 2.7$ & $21.6 \pm 4.3$ & $14.2 \pm 1.6$ & $12.2 \pm 1.9$ & $10.5 \pm 1.3$ \\
Qwen3.5-397B & $64.2 \pm 4.1$ & $32.8 \pm 1.8$ & $28.3 \pm 3.4$ & $19.2 \pm 2.2$ & $16.2 \pm 2.0$ & $13.1 \pm 1.8$ \\
DeepSeek-V3.2 & $79.2 \pm 2.8$ & $34.7 \pm 1.8$ & $28.8 \pm 3.7$ & $18.2 \pm 2.4$ & $16.0 \pm 2.2$ & $15.2 \pm 1.1$ \\
Median of five LLMs & $66.0 \pm 3.0$ & $31.7 \pm 4.1$ & $22.9 \pm 4.3$ & $17.6 \pm 1.6$ & $14.1 \pm 1.6$ & $12.6 \pm 0.8$ \\
\midrule
TabPFN & --- & $39.3 \pm 5.1$ & $26.3 \pm 4.5$ & $16.3 \pm 3.0$ & $12.8 \pm 1.3$ & $10.1 \pm 0.7$ \\
Tuned XGBoost & --- & $46.4 \pm 5.7$ & $30.9 \pm 5.3$ & $20.4 \pm 1.9$ & $16.2 \pm 1.4$ & $11.1 \pm 1.0$ \\
Tuned RF & --- & $48.8 \pm 5.0$ & $36.4 \pm 2.9$ & $22.8 \pm 2.4$ & $15.5 \pm 1.7$ & $12.8 \pm 0.4$ \\
\shortstack[l]{Median of three\\supervised ML models} & --- & $45.4 \pm 4.6$ & $30.7 \pm 4.1$ & $19.4 \pm 2.4$ & $13.2 \pm 1.8$ & $10.9 \pm 0.9$ \\
\midrule
Ridge & --- & $47.1 \pm 5.6$ & $38.2 \pm 2.2$ & $30.8 \pm 3.4$ & $27.5 \pm 2.5$ & $25.9 \pm 1.5$ \\
Gaussian process & --- & $46.1 \pm 3.8$ & $29.5 \pm 3.3$ & $22.0 \pm 2.4$ & $18.0 \pm 2.0$ & $14.6 \pm 0.4$ \\
Ordinary least squares & --- & $67.3 \pm 8.6$ & $66.1 \pm 11.3$ & $40.5 \pm 6.8$ & $26.6 \pm 3.2$ & $23.8 \pm 2.0$ \\
\midrule
CARTE & --- & $62.7 \pm 6.6$ & $47.9 \pm 1.9$ & $35.4 \pm 3.7$ & $23.2 \pm 3.4$ & $17.1 \pm 2.3$ \\
\bottomrule
\end{tabular}

\vspace{0.4em}
\begin{minipage}{\linewidth}
{\footnotesize Same as Supplementary Table~2, but for WAPE. Each value is the median across the 20 datasets of the median across the five seeds within each dataset, and the value after $\pm$ is the s.d. across seeds. The rows for the five LLMs and the three supervised ML models pool the seeds and models of each dataset.}
\end{minipage}
\end{table}

\clearpage

\begin{figure}[H]
  \centering
  \includegraphics[width=114.01mm]{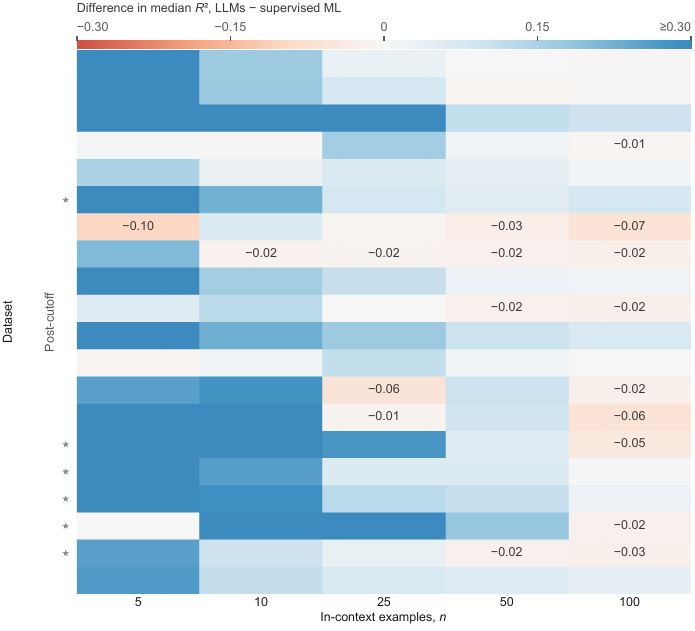}
  \caption{\textbf{\boldmath Differences between LLMs and supervised ML models for each dataset and number of examples.}
  Each cell is the median over five seeds of the $R^2$ difference, LLMs minus supervised ML models, with each group represented by its median across models within a seed. Values are shown for cells in which the supervised ML models performed better by more than 0.01. Datasets are grouped by domain, and stars mark the six post-cutoff datasets.}
  \label{si:fig4}
\end{figure}

\clearpage

\begin{table}[H]
\centering
\small
\caption*{\textbf{\boldmath Supplementary Table 4 $\vert$ Differences between LLMs and supervised ML models for each dataset at $n=5$ and $n=100$.}}
\label{si:tab4}
\begin{tabular}{@{}lrrrr@{}}
\toprule
 & \multicolumn{2}{c}{$n=5$} & \multicolumn{2}{c}{$n=100$} \\
Dataset & $R^2$ & WAPE (pp) & $R^2$ & WAPE (pp) \\
\midrule
Buccolingual Diameter & $+0.501$ & $+10.3$ & $-0.005$ & $-0.3$ \\
Carbon Intensity & $+0.938$ & $+20.5$ & $+0.095$ & $+4.6$ \\
Corrosion Degree & $+0.207$ & $+7.2$ & $-0.025$ & $+1.7$ \\
Daily Trips (Citi Bike) & $+0.255$ & $+13.5$ & $-0.026$ & $-0.6$ \\
Daily Trips (Divvy) & $+0.000$ & $-0.3$ & $-0.020$ & $+0.0$ \\
Drug--Polymer Solubility & $+0.149$ & $+3.8$ & $+0.013$ & $+3.2$ \\
Drug--Water Solubility & $+0.009$ & $+5.1$ & $-0.012$ & $-1.4$ \\
Elastic Modulus (FDM) & $+0.528$ & $+7.0$ & $-0.047$ & $-1.0$ \\
FRP--Steel Bond & $-0.008$ & $+5.2$ & $-0.001$ & $+1.1$ \\
Formation Enthalpy & $-0.096$ & $+6.3$ & $-0.065$ & $-4.6$ \\
Hourly Ridership & $+0.320$ & $+28.1$ & $+0.026$ & $+7.2$ \\
Hydrogen Uptake & $+0.609$ & $+25.4$ & $+0.080$ & $+9.4$ \\
MIDR (MDOF) & $+0.521$ & $+18.7$ & $-0.062$ & $-0.7$ \\
Max Displacement & $+0.255$ & $+8.9$ & $-0.023$ & $+3.4$ \\
Mesiodistal Diameter & $+0.775$ & $+15.6$ & $-0.006$ & $-0.3$ \\
Shear Force & $+0.758$ & $+57.6$ & $+0.072$ & $+2.0$ \\
Shore Hardness (FDM) & $+0.336$ & $+3.8$ & $+0.006$ & $+0.0$ \\
Splitting Strength & $+0.060$ & $+10.1$ & $-0.024$ & $-0.4$ \\
Torsional Strength & $+0.453$ & $+38.1$ & $+0.017$ & $+4.4$ \\
Undrained Shear Strength & $+0.264$ & $+7.3$ & $+0.047$ & $+0.4$ \\
\midrule
Datasets with LLMs better & 17/20 & 19/20 & 8/20 & 11/20 \\
\bottomrule
\end{tabular}

\vspace{0.4em}
\begin{minipage}{\linewidth}
{\footnotesize Each value is the difference between the LLMs and the three supervised ML models on one dataset, taken as the median over the five seeds. Positive values indicate an LLM advantage. WAPE differences are in percentage points (pp).}
\end{minipage}
\end{table}

\clearpage
\section*{Supplementary Note 2 $\vert$ LLMs rely on variable names for numerical prediction}
\label{si:note2}

This note supports Fig.~3 of the main text and reports how the effect of variable names changes with the number of examples, how it compares with that on CARTE and how it varies across datasets.

\subsection*{Effect of variable names on LLMs across numbers of examples}

Variable names improved the $R^2$ of the LLMs by $+0.299$ at $n=5$ but only by $+0.030$ at $n=100$ (paired Hodges--Lehmann estimates). At $n=100$, the improvement remained positive for every LLM, ranging from $+0.012$ to $+0.032$ for four LLMs and reaching $+0.092$ for DeepSeek-V3.2 (\sifig{8}c,d).

\subsection*{Effect of variable names on CARTE}

CARTE is a supervised ML model that takes variable names as input. We fitted it with and without variable names on the same data splits, seeds and numbers of examples. Unlike the LLMs, it benefited little from them (\sitab{5}). The paired Hodges--Lehmann effect of variable names on CARTE was $+0.004$ to $+0.025$ across numbers of examples, against $+0.030$ to $+0.299$ for the LLMs. For CARTE, Wilcoxon signed-rank tests detected no effect at any number of examples.

\begin{table}[H]
\centering
\caption*{\textbf{\boldmath Supplementary Table 5 $\vert$ Effect of variable names on CARTE and the LLMs across numbers of examples.}}
\label{si:tab5}
\footnotesize
\setlength{\tabcolsep}{5pt}
\begin{tabular}{@{}lrrrrrr@{}}
\toprule
\shortstack{Number of\\examples} & CARTE HL & \shortstack{Signed-rank\\95\% CI} & \shortstack{Dataset-bootstrap\\95\% CI} & $P$ & \shortstack{Datasets\\improved} & \shortstack{Five LLMs combined\\HL} \\
\midrule
$n=5$ & $+0.008$ & $-0.005$ to $+0.063$ & $-0.003$ to $+0.062$ & $0.379$ & 11/20 & $+0.299$ \\
$n=10$ & $+0.012$ & $-0.006$ to $+0.064$ & $+0.001$ to $+0.054$ & $0.145$ & 14/20 & $+0.152$ \\
$n=25$ & $+0.018$ & $-0.028$ to $+0.073$ & $-0.026$ to $+0.072$ & $0.398$ & 15/20 & $+0.092$ \\
$n=50$ & $+0.025$ & $-0.009$ to $+0.065$ & $-0.014$ to $+0.074$ & $0.149$ & 11/20 & $+0.049$ \\
$n=100$ & $+0.004$ & $-0.013$ to $+0.021$ & $-0.012$ to $+0.019$ & $0.613$ & 9/20 & $+0.030$ \\
\bottomrule
\end{tabular}

\vspace{0.4em}
\begin{minipage}{\linewidth}
{\footnotesize Effects are paired Hodges--Lehmann (HL) estimates of \textit{Correct variable names} minus \textit{Anonymised}, with confidence intervals (CIs) and tests as in Supplementary Methods. Datasets improved counts the datasets on which variable names improved CARTE. At $n=10$, two seeds of FRP--Steel Bond failed for CARTE.}
\end{minipage}
\end{table}

\clearpage

\subsection*{Effect of variable names across datasets}

The improvement from variable names varied across datasets (\sifig{5}). It was smallest for Drug--Polymer Solubility, while Drug--Water Solubility and Formation Enthalpy ranked sixth and seventh smallest. The feature names of these three datasets are molecular descriptors that convey little domain meaning.

\begin{figure}[H]
  \centering
  \includegraphics[width=169.68mm]{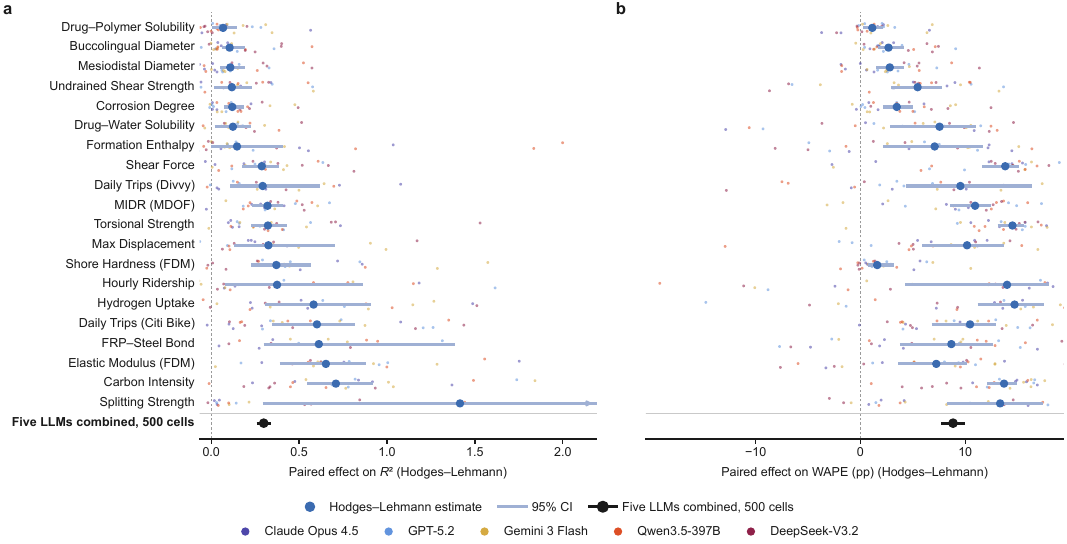}
  \caption{\textbf{\boldmath Effect of variable names on each dataset at $n=5$.}
  Each row shows \textit{Correct variable names} minus \textit{Anonymised} for one dataset. Small points show the 25 LLM--seed differences, blue dots and bars show the Hodges--Lehmann estimate and its 95\% confidence interval, and the black line indicates the estimate over all 500 cells. Rows are ordered by the $R^2$ estimate.
  \textbf{a}, $R^2$, with the axis truncated at 2.2 and an arrow marking an interval that extends to $+5.3$.
  \textbf{b}, WAPE in percentage points, with positive values denoting improvement, on a symmetric-logarithmic axis that is linear within $\pm 10$.}
  \label{si:fig5}
  \end{figure}

\clearpage

\section*{Supplementary Note 3 $\vert$ Variable names improve predictions most when they match the values they describe}
\label{si:note3}

This note supports Fig.~4 of the main text and reports the separate contributions of feature names and the target name, a check with a standardised prompt format, the semantic perturbations and the effect of prompt guidance.

\subsection*{Contributions of feature names and the target name}
\label{si:feature-target}

In a factorial design with \textit{Target name only} and \textit{Feature names only} (Fig.~4a of the main text, \sifig{6}a,b and \sitab{6}), feature names contributed more than the target name in all five LLMs. Feature names improved $R^2$ by $+0.237$ without the target name and $+0.196$ with it, whereas the target name improved it by $+0.084$ without feature names and $+0.045$ with them (paired Hodges--Lehmann estimates).

\begin{figure}[H]
\centering
\includegraphics[width=172.14mm]{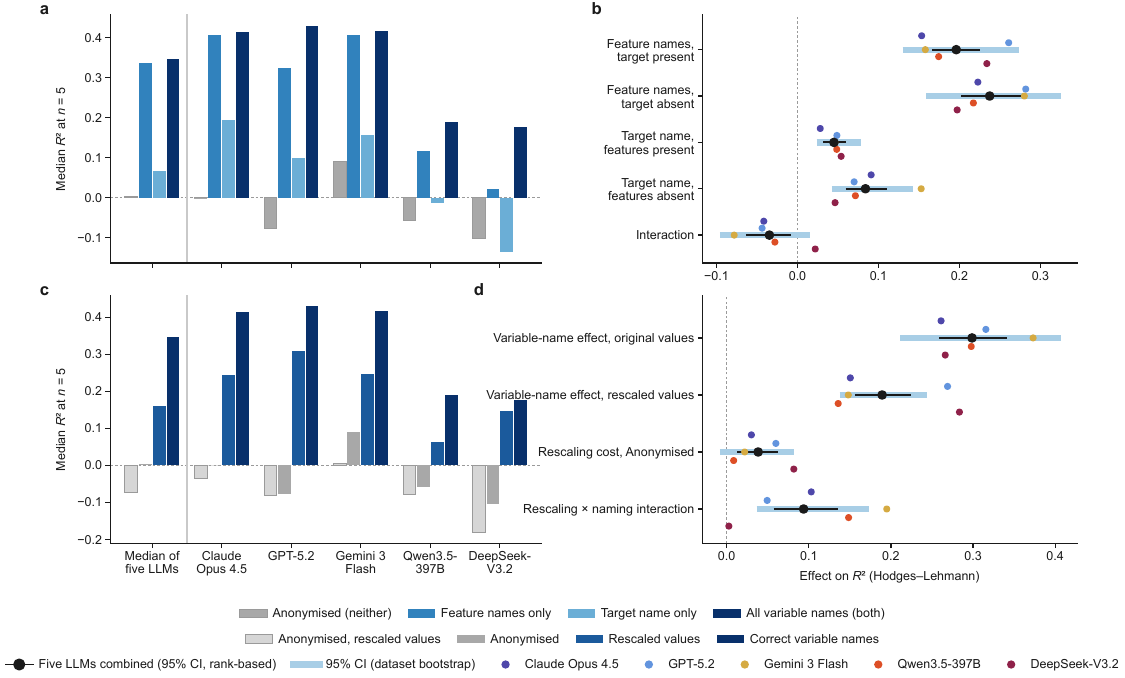}
\caption{\textbf{\boldmath Effects of feature names, the target name and rescaled values at $n=5$.}
\textbf{a}, \textbf{b}, Feature and target names given or withheld.
\textbf{c}, \textbf{d}, Numerical values rescaled by random factors of 5 to 31.6 or their reciprocals, with or without the names (Methods).
\textbf{a}, \textbf{c}, Median $R^2$ of each condition across the five LLMs and for each LLM.
\textbf{b}, \textbf{d}, Paired Hodges--Lehmann effects and their interaction, with signed-rank (black) and dataset-bootstrap (pale) 95\% confidence intervals and estimates for each LLM (dots). Numerical results are given in \sitab{6} for a,b and in \sitab{22} for c,d.}
\label{si:fig6}
\end{figure}

\begin{table}[H]
\centering
\scriptsize
\setlength{\tabcolsep}{1.8pt}
\caption*{\textbf{\boldmath Supplementary Table 6 $\vert$ Feature-name and target-name contributions at $n=5$.}}
\label{si:tab6}
\begin{tabular}{@{}lrrrrrr@{}}
\toprule
Condition & \shortstack{Five LLMs\\combined} & Claude Opus 4.5 & GPT-5.2 & Gemini 3 Flash & Qwen3.5-397B & DeepSeek-V3.2 \\
\midrule
Anonymised (neither) & $0.003 \pm 0.104$ & $0.000 \pm 0.120$ & $-0.077 \pm 0.160$ & $0.089 \pm 0.177$ & $-0.057 \pm 0.070$ & $-0.103 \pm 0.139$ \\
Feature names only & $0.337 \pm 0.113$ & $0.406 \pm 0.089$ & $0.325 \pm 0.117$ & $0.407 \pm 0.139$ & $0.117 \pm 0.109$ & $0.022 \pm 0.125$ \\
Target name only & $0.066 \pm 0.064$ & $0.194 \pm 0.057$ & $0.098 \pm 0.099$ & $0.157 \pm 0.174$ & $-0.014 \pm 0.056$ & $-0.135 \pm 0.070$ \\
Correct variable names (both) & $0.346 \pm 0.071$ & $0.415 \pm 0.059$ & $0.429 \pm 0.098$ & $0.417 \pm 0.065$ & $0.189 \pm 0.072$ & $0.177 \pm 0.156$ \\
\midrule
Contrast & HL & \shortstack{95\% CI\\(signed rank)} & \shortstack{95\% CI\\(dataset bootstrap)} & $P$ & & \\
\midrule
Feature names, target present & $0.1960$ & $0.1665$ to $0.2259$ & $0.1298$ to $0.2729$ & $1.5\times10^{-44}$ \\
Feature names, target absent & $0.2373$ & $0.2025$ to $0.2756$ & $0.1587$ to $0.3250$ & $3.3\times10^{-46}$ \\
Target name, features present & $0.0450$ & $0.0318$ to $0.0594$ & $0.0237$ to $0.0780$ & $6.1\times10^{-12}$ \\
Target name, features absent & $0.0839$ & $0.0595$ to $0.1108$ & $0.0425$ to $0.1420$ & $5.5\times10^{-14}$ \\
Interaction & $-0.0347$ & $-0.0636$ to $-0.0080$ & $-0.0959$ to $0.0148$ & $0.0095$ \\
\midrule
Contrast, per model & Claude Opus 4.5 & GPT-5.2 & Gemini 3 Flash & Qwen3.5-397B & DeepSeek-V3.2 \\
\midrule
Feature names, target present & $+0.153$ & $+0.261$ & $+0.158$ & $+0.174$ & $+0.234$ \\
Feature names, target absent & $+0.223$ & $+0.282$ & $+0.280$ & $+0.217$ & $+0.197$ \\
Target name, features present & $+0.028$ & $+0.049$ & $+0.046$ & $+0.048$ & $+0.054$ \\
Target name, features absent & $+0.091$ & $+0.070$ & $+0.153$ & $+0.071$ & $+0.046$ \\
Interaction & $-0.042$ & $-0.044$ & $-0.078$ & $-0.028$ & $+0.022$ \\
\bottomrule
\end{tabular}

\vspace{0.4em}
\begin{minipage}{\linewidth}
{\footnotesize The table compares four prompts that give no names, only the feature names, only the target name or both. The upper block gives the median $R^2$ of each prompt across the 20 datasets, after taking the median across seeds within each dataset. The middle block gives how much the feature names improved $R^2$ with and without the target name, how much the target name improved it with and without the feature names, and the interaction between the two (paired Hodges--Lehmann estimates). The bottom block gives these improvements for each LLM.}
\end{minipage}
\end{table}

\subsection*{Improvement from feature names in a standardised prompt format}

The improvement from feature names changed little with the prompt layout. In the standardised format, the prompts with and without feature names gave the same target name and units and differed only in whether each feature was described by its position or by its name (\hyperref[si:alternative-prompts]{Supplementary Methods}). Measured as the difference between condition medians, as in the main text, the improvement was 0.334 in the original format and 0.305 in the standardised one (Fig.~4b of the main text and \sitab{7}). In the standardised format, text values, such as male or female for sex in the dental datasets, were replaced by codes that do not reveal their meaning, such as \verb|v1_1| and \verb|v1_2|. In the nine datasets with text values, feature names improved $R^2$ by 0.274, and restoring the original text values improved it by a further 0.214 (medians of paired differences; \sitab{8}).

\clearpage

\begin{table}[H]
\centering
\small
\caption*{\textbf{\boldmath Supplementary Table 7 $\vert$ Improvement from feature names for each dataset in the standardised prompt format at $n=5$.}}
\label{si:tab7}
\begin{tabular}{@{}lrrrr@{}}
\toprule
Dataset & Anonymised & Feature names & Difference & Paired \\
\midrule
Buccolingual Diameter$^{\S}$ & $+0.151$ & $+0.457$ & $+0.306$ & $+0.344$ \\
Carbon Intensity$^{\S}$ & $-0.098$ & $+0.425$ & $+0.523$ & $+0.435$ \\
Corrosion Degree & $+0.776$ & $+0.801$ & $+0.025$ & $+0.028$ \\
Daily Trips (Citi Bike)$^{\S}$ & $-0.191$ & $+0.167$ & $+0.357$ & $+0.274$ \\
Daily Trips (Divvy)$^{\S}$ & $-0.099$ & $-0.088$ & $+0.011$ & $+0.014$ \\
Drug--Polymer Solubility & $-0.127$ & $-0.147$ & $-0.020$ & $+0.002$ \\
Drug--Water Solubility & $+0.179$ & $+0.297$ & $+0.118$ & $+0.121$ \\
Elastic Modulus (FDM)$^{\S}$ & $-0.195$ & $-0.161$ & $+0.034$ & $+0.063$ \\
FRP--Steel Bond & $-0.513$ & $-0.512$ & $+0.001$ & $+0.001$ \\
Formation Enthalpy & $+0.379$ & $+0.479$ & $+0.100$ & $+0.003$ \\
Hourly Ridership$^{\S}$ & $+0.127$ & $+0.301$ & $+0.173$ & $+0.109$ \\
Hydrogen Uptake$^{\S}$ & $-0.281$ & $+0.087$ & $+0.368$ & $+0.339$ \\
MIDR (MDOF) & $+0.088$ & $+0.395$ & $+0.307$ & $+0.236$ \\
Max Displacement & $-0.289$ & $-0.210$ & $+0.078$ & $+0.023$ \\
Mesiodistal Diameter$^{\S}$ & $+0.213$ & $+0.483$ & $+0.270$ & $+0.292$ \\
Shear Force & $+0.547$ & $+0.644$ & $+0.096$ & $+0.063$ \\
Shore Hardness (FDM)$^{\S}$ & $-0.265$ & $-0.217$ & $+0.048$ & $+0.068$ \\
Splitting Strength & $+0.033$ & $+0.375$ & $+0.342$ & $+0.115$ \\
Torsional Strength & $+0.855$ & $+0.879$ & $+0.024$ & $+0.010$ \\
Undrained Shear Strength & $-0.045$ & $+0.172$ & $+0.217$ & $+0.167$ \\
\midrule
Across datasets & $-0.006$ & $+0.299$ & $+0.305$ & $+0.089$ \\
\bottomrule
\end{tabular}

\vspace{0.4em}
\begin{minipage}{\linewidth}
{\footnotesize Columns give the median $R^2$ without and with feature names, the difference between these medians, and the median of the 25 paired differences across seeds and LLMs. The last row summarises the 20 datasets. Section signs mark the nine datasets with text values.}
\end{minipage}
\end{table}

\clearpage

\begin{table}[H]
\centering
\small
\caption*{\textbf{\boldmath Supplementary Table 8 $\vert$ Improvement from feature names and from text values in the nine datasets with text values at $n=5$.}}
\label{si:tab8}
\begin{tabular}{@{}lrrrrr@{}}
\toprule
 & & Feature & Original & Feature names & Text-value \\
Dataset & Anonymised & names & values & contribution & contribution \\
\midrule
Buccolingual Diameter & $+0.151$ & $+0.457$ & $+0.816$ & $+0.344$ & $+0.330$ \\
Carbon Intensity & $-0.098$ & $+0.425$ & $+0.406$ & $+0.435$ & $+0.051$ \\
Daily Trips (Citi Bike) & $-0.191$ & $+0.167$ & $+0.357$ & $+0.274$ & $+0.144$ \\
Daily Trips (Divvy) & $-0.099$ & $-0.088$ & $-0.065$ & $+0.014$ & $+0.214$ \\
Elastic Modulus (FDM) & $-0.195$ & $-0.161$ & $+0.320$ & $+0.063$ & $+0.492$ \\
Hourly Ridership & $+0.127$ & $+0.301$ & $+0.212$ & $+0.109$ & $-0.114$ \\
Hydrogen Uptake & $-0.281$ & $+0.087$ & $+0.149$ & $+0.339$ & $+0.097$ \\
Mesiodistal Diameter & $+0.213$ & $+0.483$ & $+0.821$ & $+0.292$ & $+0.338$ \\
Shore Hardness (FDM) & $-0.265$ & $-0.217$ & $+0.267$ & $+0.068$ & $+0.479$ \\
\midrule
Across datasets & $-0.099$ & $+0.167$ & $+0.320$ & $+0.274$ & $+0.214$ \\
\bottomrule
\end{tabular}

\vspace{0.4em}
\begin{minipage}{\linewidth}
{\footnotesize The nine datasets have text values, such as male or female for sex. \textit{Anonymised} and \textit{Feature names} replace these text values with codes that do not reveal their meaning, and \textit{Original values} restores them. \textit{Anonymised} also omits the feature names. The first three columns give the median $R^2$ of each prompt. Feature names contribution is the improvement of \textit{Feature names} over \textit{Anonymised}, and Text-value contribution is the further improvement of \textit{Original values} over \textit{Feature names}. Both are medians of the 25 paired differences across seeds and LLMs. All prompts use the standardised format. The last row summarises the nine datasets.}
\end{minipage}
\end{table}

\clearpage
\subsection*{Predictions when variable names do not match the values}
\label{si:perturbations}
\begingroup
\setlength{\parskip}{0.3em}

Under \textit{Shuffled names}, the small improvement that remained came almost entirely from the target name, which was still correct. \textit{Shuffled names} improved $R^2$ by $+0.097$ over \textit{Anonymised}, close to the $+0.084$ from the target name alone. Over \textit{Target name only}, it improved $R^2$ by only $+0.010$ (paired Hodges--Lehmann estimates). \textit{Cross-domain names}, which also replaced the target name, removed the improvement from the target name as well and did not differ detectably from \textit{Anonymised} ($-0.010$). The four perturbation conditions are shown for each LLM in \sifig{7}, with all paired comparisons in \sitab{22}.

Part of the improvement came from the variable names conveying the typical magnitude of each quantity. Each feature and the target were multiplied by their own undisclosed factor, between 5 and 31.6 or the reciprocal of such a value, in prompts with and without names (Methods). This rescaling reduced the improvement from variable names by 0.094 in $R^2$, about a third of the $+0.299$ improvement at $n=5$. Rescaling did not increase the improvement in any LLM (paired Hodges--Lehmann estimate, \sifig{6}c,d).

\par\endgroup
\clearpage

\begin{figure}[H]
  \centering
  \includegraphics[width=169.17mm]{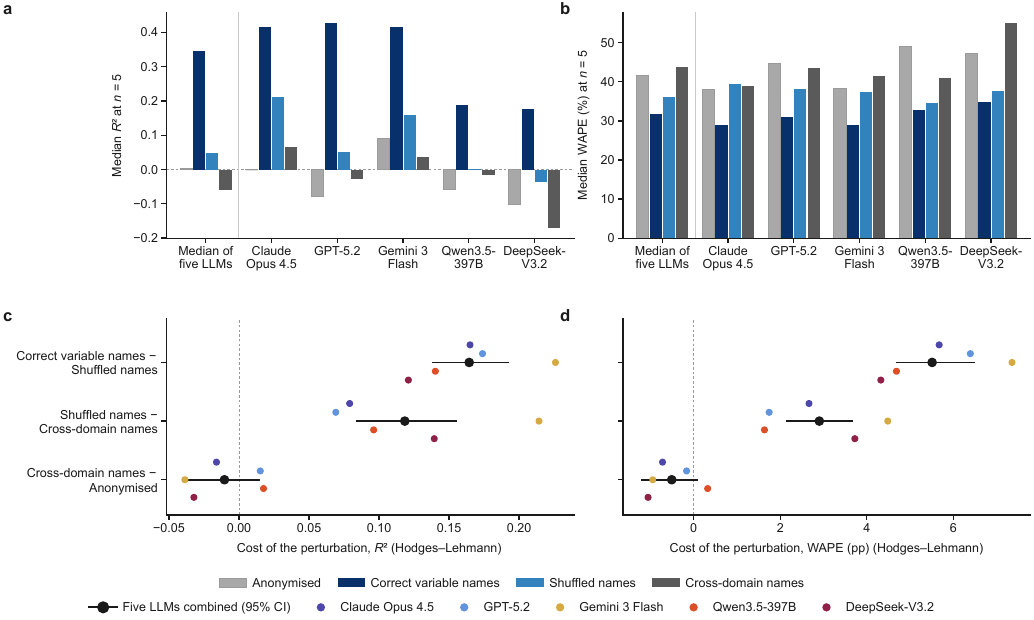}
  \caption{\textbf{\boldmath Effects of perturbing variable names at $n=5$.}
  \textbf{a}, \textbf{b}, Median $R^2$ and WAPE of each condition across the five LLMs and for each LLM.
  \textbf{c}, \textbf{d}, Loss of performance from \textit{Correct variable names} to \textit{Shuffled names}, from \textit{Shuffled names} to \textit{Cross-domain names} and from \textit{Cross-domain names} to \textit{Anonymised} (paired Hodges--Lehmann estimates, positive values denote a loss), with 95\% confidence intervals and estimates for each LLM (dots). Numerical values are given in \sitab{22}.}
  \label{si:fig7}
  \end{figure}
\clearpage

\clearpage

\subsection*{Effect of explicit guidance in the prompt}
\label{si:guidance}

Explicit guidance in the prompt barely changed the predictions. \textit{+ Semantic guidance} stated that the feature names carry domain meaning, and \textit{+ Causal guidance} also asked the LLMs to reason about the relationships between the features and the target (Supplementary Methods). At both $n=5$ and $n=100$, each changed $R^2$ by less than 0.002 (paired Hodges--Lehmann estimates, \sitab{22}), and the same held at every number of examples (\sifig{8} and Supplementary Tables~\sitabn{9} and \sitabn{10}).

\clearpage

\begin{figure}[H]
  \centering
  \captionsetup{font=footnotesize}
  \includegraphics[width=174.96mm]{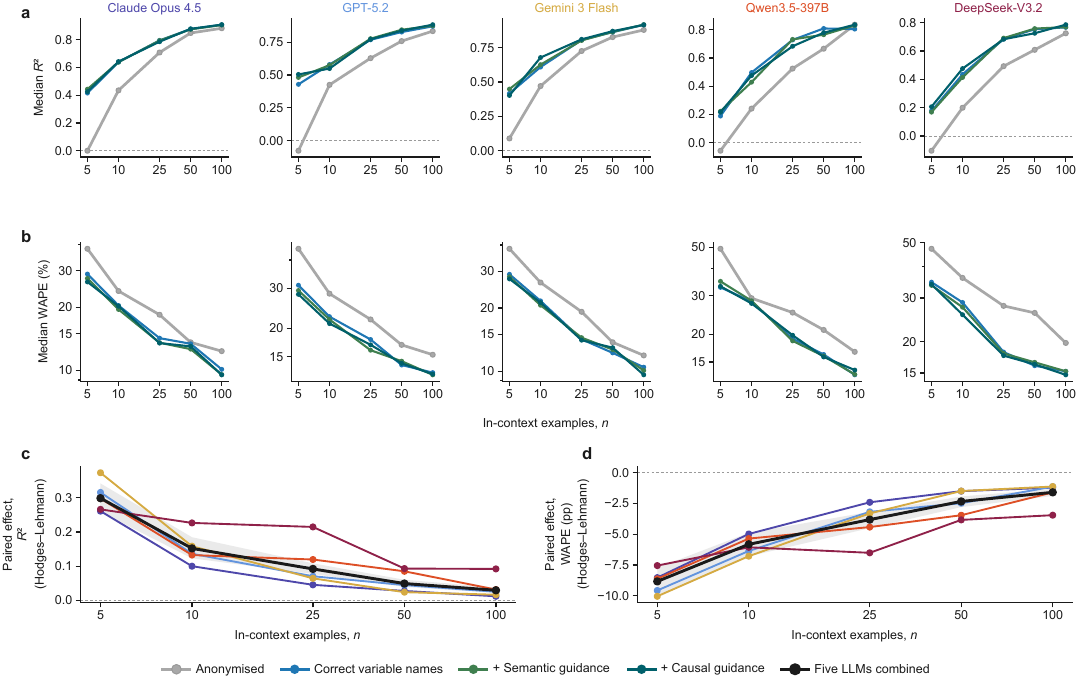}
  \caption{\textbf{\boldmath Prompt conditions and the improvement from variable names across numbers of examples.}
  \textbf{a}, \textbf{b}, Median $R^2$ and WAPE of the four prompt conditions for each LLM (\sitab{9} and \sitab{10}).
  \textbf{c}, \textbf{d}, Improvement from variable names (\textit{Correct variable names} minus \textit{Anonymised}) for each LLM (coloured lines) and for the five LLMs combined (black line), as paired Hodges--Lehmann estimates with 95\% confidence intervals (band).}
  \label{si:fig8}
  \end{figure}

\clearpage

\begin{table}[H]
\centering
\scriptsize
\setlength{\tabcolsep}{5.5pt}
\caption*{\textbf{\boldmath Supplementary Table 9 $\vert$ $R^2$ under the four prompt conditions.}}
\label{si:tab9}
\begin{tabular}{@{}lrrrrrr@{}}
\toprule
Prompt condition / model & $n=0$ & $n=5$ & $n=10$ & $n=25$ & $n=50$ & $n=100$ \\
\midrule
\multicolumn{7}{@{}l}{\textit{Anonymised}} \\
\quad Median of five LLMs & $-13.422 \pm 1.599$ & $0.003 \pm 0.104$ & $0.311 \pm 0.191$ & $0.582 \pm 0.038$ & $0.762 \pm 0.025$ & $0.825 \pm 0.023$ \\
\quad Claude Opus 4.5 & $-21.893 \pm 6.613$ & $0.000 \pm 0.120$ & $0.435 \pm 0.121$ & $0.707 \pm 0.039$ & $0.847 \pm 0.022$ & $0.881 \pm 0.024$ \\
\quad GPT-5.2 & $-2.7\times10^{3} \pm 1.9\times10^{5}$ & $-0.077 \pm 0.160$ & $0.425 \pm 0.150$ & $0.628 \pm 0.035$ & $0.759 \pm 0.023$ & $0.835 \pm 0.019$ \\
\quad Gemini 3 Flash & $-13.730 \pm 1.340$ & $0.089 \pm 0.177$ & $0.471 \pm 0.150$ & $0.726 \pm 0.052$ & $0.827 \pm 0.018$ & $0.879 \pm 0.013$ \\
\quad Qwen3.5-397B & $-12.045 \pm 1.434$ & $-0.057 \pm 0.070$ & $0.241 \pm 0.205$ & $0.525 \pm 0.098$ & $0.666 \pm 0.063$ & $0.836 \pm 0.037$ \\
\quad DeepSeek-V3.2 & $-5.419 \pm 0.728$ & $-0.103 \pm 0.139$ & $0.200 \pm 0.127$ & $0.493 \pm 0.075$ & $0.609 \pm 0.062$ & $0.725 \pm 0.098$ \\
\midrule
\multicolumn{7}{@{}l}{\textit{Correct variable names}} \\
\quad Median of five LLMs & $-0.335 \pm 0.068$ & $0.346 \pm 0.071$ & $0.547 \pm 0.053$ & $0.745 \pm 0.036$ & $0.849 \pm 0.027$ & $0.865 \pm 0.008$ \\
\quad Claude Opus 4.5 & $-0.286 \pm 0.098$ & $0.415 \pm 0.059$ & $0.638 \pm 0.044$ & $0.794 \pm 0.024$ & $0.875 \pm 0.019$ & $0.907 \pm 0.009$ \\
\quad GPT-5.2 & $-1.841 \pm 0.641$ & $0.429 \pm 0.098$ & $0.580 \pm 0.058$ & $0.769 \pm 0.024$ & $0.828 \pm 0.031$ & $0.871 \pm 0.020$ \\
\quad Gemini 3 Flash & $-0.630 \pm 0.132$ & $0.417 \pm 0.065$ & $0.609 \pm 0.068$ & $0.806 \pm 0.027$ & $0.863 \pm 0.020$ & $0.917 \pm 0.012$ \\
\quad Qwen3.5-397B & $-0.272 \pm 0.060$ & $0.189 \pm 0.072$ & $0.497 \pm 0.079$ & $0.728 \pm 0.017$ & $0.809 \pm 0.043$ & $0.805 \pm 0.017$ \\
\quad DeepSeek-V3.2 & $-1.085 \pm 0.094$ & $0.177 \pm 0.156$ & $0.436 \pm 0.065$ & $0.684 \pm 0.082$ & $0.754 \pm 0.018$ & $0.767 \pm 0.021$ \\
\midrule
\multicolumn{7}{@{}l}{\textit{$+$ Semantic guidance}} \\
\quad Median of five LLMs & $-0.480 \pm 0.127$ & $0.360 \pm 0.096$ & $0.553 \pm 0.046$ & $0.777 \pm 0.031$ & $0.846 \pm 0.038$ & $0.871 \pm 0.011$ \\
\quad Claude Opus 4.5 & $-0.195 \pm 0.119$ & $0.443 \pm 0.071$ & $0.638 \pm 0.038$ & $0.793 \pm 0.025$ & $0.874 \pm 0.022$ & $0.905 \pm 0.015$ \\
\quad GPT-5.2 & $-1.605 \pm 0.301$ & $0.481 \pm 0.075$ & $0.575 \pm 0.062$ & $0.777 \pm 0.038$ & $0.845 \pm 0.022$ & $0.876 \pm 0.023$ \\
\quad Gemini 3 Flash & $-0.614 \pm 0.258$ & $0.449 \pm 0.072$ & $0.626 \pm 0.076$ & $0.803 \pm 0.019$ & $0.865 \pm 0.022$ & $0.918 \pm 0.014$ \\
\quad Qwen3.5-397B & $-0.274 \pm 0.051$ & $0.222 \pm 0.122$ & $0.429 \pm 0.075$ & $0.732 \pm 0.031$ & $0.765 \pm 0.050$ & $0.831 \pm 0.010$ \\
\quad DeepSeek-V3.2 & $-0.982 \pm 0.128$ & $0.169 \pm 0.130$ & $0.414 \pm 0.092$ & $0.692 \pm 0.070$ & $0.758 \pm 0.026$ & $0.769 \pm 0.032$ \\
\midrule
\multicolumn{7}{@{}l}{\textit{$+$ Causal guidance}} \\
\quad Median of five LLMs & $-0.473 \pm 0.250$ & $0.346 \pm 0.054$ & $0.551 \pm 0.053$ & $0.755 \pm 0.025$ & $0.838 \pm 0.035$ & $0.866 \pm 0.006$ \\
\quad Claude Opus 4.5 & $-0.258 \pm 0.124$ & $0.425 \pm 0.088$ & $0.642 \pm 0.044$ & $0.785 \pm 0.015$ & $0.877 \pm 0.014$ & $0.907 \pm 0.011$ \\
\quad GPT-5.2 & $-1.404 \pm 0.502$ & $0.504 \pm 0.103$ & $0.549 \pm 0.078$ & $0.772 \pm 0.030$ & $0.838 \pm 0.035$ & $0.884 \pm 0.016$ \\
\quad Gemini 3 Flash & $-0.613 \pm 0.153$ & $0.402 \pm 0.076$ & $0.678 \pm 0.069$ & $0.811 \pm 0.018$ & $0.870 \pm 0.017$ & $0.916 \pm 0.016$ \\
\quad Qwen3.5-397B & $-0.289 \pm 0.133$ & $0.216 \pm 0.064$ & $0.476 \pm 0.079$ & $0.683 \pm 0.054$ & $0.779 \pm 0.033$ & $0.836 \pm 0.008$ \\
\quad DeepSeek-V3.2 & $-1.116 \pm 0.223$ & $0.207 \pm 0.097$ & $0.476 \pm 0.112$ & $0.683 \pm 0.076$ & $0.726 \pm 0.047$ & $0.786 \pm 0.039$ \\
\bottomrule
\end{tabular}

\vspace{0.4em}
\begin{minipage}{\linewidth}
{\footnotesize Each value is the median across the 20 datasets of the median across the five seeds within each dataset, and the value after $\pm$ is the s.d. across seeds. The rows for the five LLMs pool the seeds and LLMs of each dataset.}
\end{minipage}
\end{table}

\clearpage

\begin{table}[H]
\centering
\scriptsize
\setlength{\tabcolsep}{5.5pt}
\caption*{\textbf{\boldmath Supplementary Table 10 $\vert$ WAPE (\%) under the four prompt conditions.}}
\label{si:tab10}
\begin{tabular}{@{}lrrrrrr@{}}
\toprule
Prompt condition / model & $n=0$ & $n=5$ & $n=10$ & $n=25$ & $n=50$ & $n=100$ \\
\midrule
\multicolumn{7}{@{}l}{\textit{Anonymised}} \\
\quad Median of five LLMs & $126.5 \pm 5.5$ & $41.6 \pm 2.8$ & $28.0 \pm 5.4$ & $21.8 \pm 1.8$ & $17.8 \pm 1.9$ & $15.8 \pm 1.7$ \\
\quad Claude Opus 4.5 & $130.0 \pm 13.2$ & $38.2 \pm 3.4$ & $24.0 \pm 4.8$ & $18.5 \pm 2.1$ & $13.6 \pm 2.7$ & $12.4 \pm 1.5$ \\
\quad GPT-5.2 & $1{,}190.3 \pm 4{,}394.1$ & $44.8 \pm 4.0$ & $28.4 \pm 4.1$ & $21.9 \pm 1.7$ & $16.9 \pm 1.3$ & $15.3 \pm 1.4$ \\
\quad Gemini 3 Flash & $125.2 \pm 6.6$ & $38.4 \pm 3.4$ & $26.5 \pm 2.8$ & $19.2 \pm 3.1$ & $13.7 \pm 1.4$ & $11.9 \pm 1.0$ \\
\quad Qwen3.5-397B & $106.4 \pm 8.7$ & $49.1 \pm 2.9$ & $29.3 \pm 5.3$ & $25.2 \pm 2.2$ & $21.0 \pm 3.0$ & $16.7 \pm 2.8$ \\
\quad DeepSeek-V3.2 & $100.0 \pm 0.1$ & $47.2 \pm 6.2$ & $36.1 \pm 5.2$ & $27.9 \pm 1.6$ & $26.1 \pm 1.7$ & $19.8 \pm 2.4$ \\
\midrule
\multicolumn{7}{@{}l}{\textit{Correct variable names}} \\
\quad Median of five LLMs & $66.0 \pm 3.0$ & $31.7 \pm 4.1$ & $22.9 \pm 4.3$ & $17.6 \pm 1.6$ & $14.1 \pm 1.6$ & $12.6 \pm 0.8$ \\
\quad Claude Opus 4.5 & $62.1 \pm 4.5$ & $28.9 \pm 3.4$ & $20.4 \pm 1.7$ & $14.3 \pm 1.6$ & $13.4 \pm 2.6$ & $10.1 \pm 1.6$ \\
\quad GPT-5.2 & $78.7 \pm 2.4$ & $31.0 \pm 4.3$ & $22.5 \pm 4.1$ & $17.8 \pm 1.6$ & $13.8 \pm 1.2$ & $12.8 \pm 0.9$ \\
\quad Gemini 3 Flash & $59.1 \pm 3.1$ & $29.0 \pm 2.7$ & $21.6 \pm 4.3$ & $14.2 \pm 1.6$ & $12.2 \pm 1.9$ & $10.5 \pm 1.3$ \\
\quad Qwen3.5-397B & $64.2 \pm 4.1$ & $32.8 \pm 1.8$ & $28.3 \pm 3.4$ & $19.2 \pm 2.2$ & $16.2 \pm 2.0$ & $13.1 \pm 1.8$ \\
\quad DeepSeek-V3.2 & $79.2 \pm 2.8$ & $34.7 \pm 1.8$ & $28.8 \pm 3.7$ & $18.2 \pm 2.4$ & $16.0 \pm 2.2$ & $15.2 \pm 1.1$ \\
\midrule
\multicolumn{7}{@{}l}{\textit{$+$ Semantic guidance}} \\
\quad Median of five LLMs & $67.8 \pm 3.3$ & $31.6 \pm 4.2$ & $22.6 \pm 3.8$ & $17.2 \pm 1.9$ & $14.6 \pm 1.9$ & $12.3 \pm 1.3$ \\
\quad Claude Opus 4.5 & $65.4 \pm 3.2$ & $27.5 \pm 4.0$ & $19.6 \pm 1.4$ & $13.6 \pm 2.1$ & $12.6 \pm 2.7$ & $9.5 \pm 1.7$ \\
\quad GPT-5.2 & $78.5 \pm 1.5$ & $29.3 \pm 4.9$ & $21.9 \pm 4.0$ & $16.0 \pm 1.7$ & $14.3 \pm 1.7$ & $12.5 \pm 1.5$ \\
\quad Gemini 3 Flash & $59.2 \pm 5.3$ & $28.2 \pm 3.5$ & $20.6 \pm 4.2$ & $14.5 \pm 0.9$ & $12.7 \pm 2.1$ & $10.1 \pm 1.0$ \\
\quad Qwen3.5-397B & $67.5 \pm 1.3$ & $34.9 \pm 1.1$ & $28.5 \pm 3.7$ & $18.7 \pm 2.0$ & $15.9 \pm 1.3$ & $13.2 \pm 1.4$ \\
\quad DeepSeek-V3.2 & $78.0 \pm 2.2$ & $33.7 \pm 2.8$ & $27.5 \pm 4.8$ & $18.0 \pm 2.5$ & $16.5 \pm 2.4$ & $15.2 \pm 1.0$ \\
\midrule
\multicolumn{7}{@{}l}{\textit{$+$ Causal guidance}} \\
\quad Median of five LLMs & $67.8 \pm 3.2$ & $29.4 \pm 4.6$ & $22.8 \pm 4.7$ & $17.2 \pm 1.7$ & $14.6 \pm 1.6$ & $11.7 \pm 1.1$ \\
\quad Claude Opus 4.5 & $69.3 \pm 5.1$ & $26.5 \pm 3.6$ & $20.3 \pm 1.6$ & $13.5 \pm 2.3$ & $13.0 \pm 2.7$ & $9.5 \pm 1.4$ \\
\quad GPT-5.2 & $77.4 \pm 2.4$ & $28.2 \pm 3.4$ & $21.0 \pm 4.5$ & $16.9 \pm 1.9$ & $14.0 \pm 1.1$ & $12.6 \pm 1.3$ \\
\quad Gemini 3 Flash & $59.8 \pm 2.9$ & $27.5 \pm 3.0$ & $21.2 \pm 5.3$ & $14.1 \pm 0.9$ & $13.0 \pm 2.1$ & $9.6 \pm 0.8$ \\
\quad Qwen3.5-397B & $65.2 \pm 2.9$ & $33.1 \pm 2.5$ & $27.7 \pm 2.7$ & $19.8 \pm 1.4$ & $15.8 \pm 1.5$ & $13.8 \pm 1.0$ \\
\quad DeepSeek-V3.2 & $74.1 \pm 3.0$ & $34.0 \pm 3.4$ & $25.7 \pm 4.4$ & $17.6 \pm 1.5$ & $16.2 \pm 3.4$ & $14.7 \pm 0.9$ \\
\bottomrule
\end{tabular}

\vspace{0.4em}
\begin{minipage}{\linewidth}
{\footnotesize Same as Supplementary Table~9, but for WAPE.}
\end{minipage}
\end{table}

\clearpage
\section*{Supplementary Note 4 $\vert$ Variable names allow LLMs to draw on pretrained domain-specific knowledge}
\label{si:note4}

This note supports Fig.~5 of the main text. It reports how extended reasoning affected accuracy and cost, how the LLMs reasoned with and without variable names, and how their reasoning changed as more feature names were given.

\subsection*{Effect of extended reasoning on accuracy and cost}
\label{si:reasoning}

Extended reasoning did not improve predictions systematically. Across the 100 LLM--dataset pairs, it raised $R^2$ in 46 without variable names and in 49 with them, with median changes of $-0.012$ and $-0.005$ (\sitab{11}). Its effect also differed in direction between LLMs. Without variable names, for example, reasoning raised $R^2$ on 12 of the 20 datasets for GPT-5.2 but on only one for DeepSeek-V3.2.

We also compared the predictions with and without reasoning directly. Within each LLM, dataset and seed, we computed the Pearson correlation between the predictions with and without reasoning for the 50 test rows. Its median was 0.93 under \textit{Correct variable names} and 0.81 under \textit{Anonymised}, and under \textit{Correct variable names} it ranged across LLMs from 0.79 for DeepSeek-V3.2 to 0.97 for Gemini 3 Flash.

Extended reasoning was also costly. With extended reasoning, the LLMs generated a median of 463 to 2,010 output tokens per prediction, although the numerical answer itself needed only four to ten tokens. The reasoning runs cost 496 USD, about seven times the estimated cost without reasoning (\sitab{11}).

\textbf{Runs with extended reasoning.} To collect the reasoning traces for Fig.~5 of the main text, \textit{Anonymised} and \textit{Correct variable names} were rerun at $n=5$ with extended reasoning. The runs used all five seeds and all 20 datasets, with the same examples, test rows and prompt templates as the main experiment. Qwen3.5-397B and
DeepSeek-V3.2 were each routed to a single provider that
enforced the reasoning-token budget (Methods). For these two LLMs, predictions were parsed from the response text with the parser of the main experiment.
For GPT-5.2, 1,313 of the 9,918 stored responses (13.2\%) contained only encrypted reasoning and no summary, and these were excluded from the analysis of reasoning traces.
For Shear Force, Max Displacement and Torsional Strength, the runs with and without reasoning and the analyses in the next two subsections used the names without units.

\begin{table}[H]
\centering
\footnotesize
\setlength{\tabcolsep}{1.1pt}
\caption*{\textbf{\boldmath Supplementary Table 11 $\vert$ Extended reasoning at $n=5$, by LLM.}}
\label{si:tab11}
\begin{tabular}{@{}lrrrrrrrrrr@{}}
\toprule
 & \multicolumn{4}{c}{Anonymised} & \multicolumn{4}{c}{Correct variable names} & & \\
\cmidrule(lr){2-5}\cmidrule(lr){6-9}
LLM & \shortstack{Without\\reasoning} & \shortstack{With\\reasoning} & Change & \shortstack{Datasets\\improved} & \shortstack{Without\\reasoning} & \shortstack{With\\reasoning} & Change & \shortstack{Datasets\\improved} & \shortstack{Output\\tokens} & \shortstack{Cost\\ratio} \\
\midrule
Five LLMs combined & $0.003$ & $0.021$ & $-0.012$ & 46/100 & $0.346$ & $0.309$ & $-0.005$ & 49/100 & 1,374 & 7.2 \\ \midrule
Claude Opus 4.5 & $0.000$ & $0.076$ & $+0.025$ & 12/20 & $0.415$ & $0.422$ & $+0.005$ & 12/20 & 956 & 6.2 \\
GPT-5.2 & $-0.077$ & $0.101$ & $+0.075$ & 12/20 & $0.429$ & $0.306$ & $-0.096$ & 7/20 & 463 & 11.0 \\
Gemini 3 Flash & $0.089$ & $0.030$ & $+0.046$ & 14/20 & $0.417$ & $0.400$ & $-0.017$ & 8/20 & 1,167 & 11.2 \\
Qwen3.5-397B & $-0.057$ & $-0.148$ & $-0.017$ & 7/20 & $0.189$ & $0.130$ & $-0.008$ & 9/20 & 2,010 & 11.4 \\
DeepSeek-V3.2 & $-0.103$ & $-0.281$ & $-0.127$ & 1/20 & $0.177$ & $0.165$ & $+0.015$ & 13/20 & 2,003 & 2.4 \\
\bottomrule
\end{tabular}

\vspace{0.4em}
\begin{minipage}{\linewidth}
{\footnotesize The left and right halves of the table give the results under \textit{Anonymised} and \textit{Correct variable names}. Without reasoning and With reasoning give the median $R^2$ across the 20 datasets, after taking the median across seeds within each dataset. Change is the median of the per-dataset changes in $R^2$, so it can differ from the difference between the two medians. Datasets improved is the number of datasets on which reasoning raised $R^2$, out of 20 for each LLM and 100 LLM--dataset pairs in the first row, which combines the five LLMs. Output tokens is the median number of output tokens per prediction with reasoning, and Cost ratio is the application programming interface (API) cost with reasoning divided by the estimated cost without it.}
\end{minipage}
\end{table}

\subsection*{Pretrained knowledge in reasoning traces with and without variable names}
In most domains, the reasoning traces of Qwen3.5-397B and DeepSeek-V3.2 used words specific to that domain when variable names were given, whereas without them they used words describing the format of the data, such as feature, numerical and categorical. \sitab{12} lists the ten most characteristic words of each domain, and Fig.~5a of the main text shows the first seven. Words common to any reasoning, such as absolute and actually, were removed before ranking (\sitab{12}).

\begin{table}[H]
\centering
\footnotesize
\caption*{\textbf{\boldmath Supplementary Table 12 $\vert$ Words most characteristic of the reasoning traces with and without variable names.}}
\label{si:tab12}
\begin{tabular}{@{}p{0.27\linewidth}p{0.69\linewidth}@{}}
\toprule
Domain & Words ($z$) \\
\midrule
Structural engineering & $\rho$ ($47$), $\surd$ ($32$), stiffness ($28$), capacity ($26$), contribution ($25$), failure ($25$), shear ($25$), steel ($23$), load ($21$), ACI ($21$) \\
Urban mobility & weekday ($138$), peak ($17$), busy ($16$), evening ($15$), night ($14$), morning ($14$), popularity ($13$), estimate ($12$), volume ($12$), popular ($12$) \\
Additive manufacturing & thinner ($16$), bonding ($14$), stiffer ($14$), stiffness ($13$), harder ($13$), thicker ($12$), printed ($12$), stiff ($11$), potentially ($11$), adhesion ($10$) \\
Biomedicine & molecular ($17$), descriptors ($16$), temperature ($14$), compound ($11$), molecule ($9$), weight ($8$), compounds ($7$), soluble ($6$), molecules ($6$), error ($6$) \\
Chemistry & $\surd$ ($24$), surface ($18$), diffusion ($16$), molecular ($12$), descriptors ($12$), law ($12$), saturation ($11$), Sieverts ($10$), carbon ($10$), distance ($10$) \\
Dental morphology & teeth ($14$), incisors ($12$), molars ($11$), deciduous ($11$), permanent ($9$), canines ($9$), males ($9$), females ($8$), mandibular ($8$), sexes ($8$) \\
Building carbon emissions & emissions ($16$), stages ($13$), contributions ($10$), intensities ($10$), stage ($8$), transport ($8$), components ($8$), kgco ($8$), accurately ($7$), conclusion ($7$) \\
Geotechnical engineering & $\sigma$ ($28$), $N_{kt}$ ($20$), $Q_{tn}$ ($19$), soil ($16$), $N_{ke}$ ($13$), CPTu ($12$), geotechnical ($12$), $q_\mathrm{net}$ ($10$), empirical ($10$), clays ($9$) \\
Non-destructive testing & correlation ($4$), $\surd$ ($3$), distance ($3$), correlates ($3$), risk ($3$), environment ($3$), knowledge ($3$), deviation ($3$), constant ($3$), increasing ($3$) \\
Anonymised (all domains) & feature ($-139$), numerical ($-67$), location ($-61$), hour ($-54$), categorical ($-47$), gender ($-46$), number ($-42$), fourth ($-41$), numbers ($-38$), height ($-38$) \\
\bottomrule
\end{tabular}

\vspace{0.4em}
\begin{minipage}{\linewidth}
{\footnotesize Words come from the full reasoning traces of Qwen3.5-397B and DeepSeek-V3.2 at $n=5$, after every word in the prompt was removed. Within each domain, words are ranked by how much more often they appeared with variable names than without (Methods), with $z$-scores in parentheses. The last row gives the words most characteristic of the traces without variable names, which therefore have negative $z$-scores. Words occurring fewer than 30 times and 125 generic words, such as absolute, actually and approximately, were excluded. The full list is in \texttt{analyse\_reasoning\_lexical.py} of the code package.}
\end{minipage}
\end{table}

\clearpage
\subsection*{Pretrained knowledge in reasoning traces as feature names are added}
Three further measures in \sitab{13} agreed with the two in Fig.~5c of the main text. As feature names were added, the reasoning traces spread over fewer domains, more of them were closest to their own domain, and their words became closer to those of the traces with all variable names. The step-by-step experiment is described in Methods.

\begin{table}[H]
\centering
\scriptsize
\caption*{\textbf{\boldmath Supplementary Table 13 $\vert$ Reasoning traces of DeepSeek-V3.2 and Qwen3.5-397B as feature names are added at $n=5$.}}
\label{si:tab13}
\setlength{\tabcolsep}{3.2pt}
\begin{tabular}{@{}l rrrrr rrrrr@{}}
\toprule
 & \multicolumn{5}{c}{DeepSeek-V3.2} & \multicolumn{5}{c}{Qwen3.5-397B} \\
\cmidrule(lr){2-6}\cmidrule(l){7-11}
Feature names & Probability & Terms & Domains & Own domain, \% & Distance & Probability & Terms & Domains & Own domain, \% & Distance \\
\midrule
No names & 0.39 & 1.0 & 3.9 & 35 & 0.50 & 0.42 & 0.6 & 4.3 & 47 & 0.60 \\
0\% (target only) & 0.56 & 4.0 & 3.4 & 55 & 0.39 & 0.67 & 3.5 & 2.9 & 85 & 0.50 \\
25\% & 0.63 & 8.8 & 3.1 & 69 & 0.36 & 0.79 & 8.9 & 2.3 & 88 & 0.39 \\
50\% & 0.72 & 8.7 & 2.7 & 90 & 0.30 & 0.78 & 9.0 & 2.2 & 85 & 0.36 \\
75\% & 0.80 & 8.9 & 2.1 & 91 & 0.24 & 0.84 & 12.5 & 2.0 & 92 & 0.35 \\
100\% & 0.84 & 13.2 & 1.8 & 95 & 0.24 & 0.89 & 17.2 & 1.7 & 97 & 0.31 \\
\bottomrule
\end{tabular}

\vspace{0.4em}
\begin{minipage}{\linewidth}
{\footnotesize Values are medians across the 20 datasets. Probability and Terms are defined in Methods. Probability is the probability that a domain classifier assigned a trace to its own domain, and Terms is the number of domain terms per trace. Domains is the effective number of domains over which the classifier spread its probability, computed as $\exp(-\sum_d p_d\log p_d)$ over the nine domains, from $1$ for a single domain to $9$ for all equally likely. Own domain is the percentage of traces whose embedding was closer, by cosine similarity, to the mean of their own domain than to that of any other domain. Distance is the Jensen--Shannon distance between the word frequencies of the traces and those of the traces with all variable names, from $0$ for identical to $1$ for no shared words. The classifier, domain terms and reference traces came from seeds other than the evaluated one.}
\end{minipage}
\end{table}

\clearpage

\section*{Supplementary Note 5 $\vert$ Pretrained domain-specific knowledge behaves as a prior that in-context examples update}
\label{si:note5}

This note supports Fig.~6 of the main text. It reports whether the predictions of the LLMs follow the knowledge elicited by the feature names when the examples cannot distinguish between relations, how the predictions change when the examples contradict this knowledge, and whether this change is consistent with Bayesian updating.

\subsection*{Controlled relation tasks that separate knowledge from examples}
\label{si:relation-tasks}

\textbf{Relation score and separation.}
Under \textit{Original assignment} the median score was positive for all 25 tasks, and under \textit{Swapped assignment} negative for all 25, whereas under \textit{Anonymous} it lay between 0.00 and 0.02 (Fig.~6b of the main text and \sitab{14}). The separation, $T_{\mathrm{Original}}-T_{\mathrm{Swapped}}$ for the same LLM and task instance, was positive in 2,302 of 2,500 pairs (92.1\%), with a median of 1.42 (95\% confidence interval, 1.34 to 1.48). The tasks and the relation score $T$ are described in Methods, and the tasks are listed in \sitab{14}.

\clearpage
\textbf{Accuracy of the predicted values.}
The predicted values were also close to the targets implied by the names. The relation score shows only the direction of the relation that the predictions followed. Each prediction was compared with two targets computed from the relation, one with the assigned names and one with the names exchanged. Of the 29,941 valid predictions, 79.0\% were closer to the target with the assigned names. The deviation of a typical prediction from this target was 19.5\%, ranging across LLMs from 0.7\% for Gemini 3 Flash to 56.2\% for DeepSeek-V3.2 (\sitab{15}).

\clearpage

\begin{table}[H]
\centering
\footnotesize
\setlength{\tabcolsep}{3pt}
\renewcommand{\arraystretch}{1.25}
\caption*{\textbf{\boldmath Supplementary Table 14 $\vert$ Controlled relation tasks, by task.}}
\label{si:tab14}
\begin{tabular}{@{}>{\raggedright\arraybackslash}p{0.16\textwidth}>{\raggedright\arraybackslash}p{0.17\textwidth}>{\raggedright\arraybackslash}p{0.31\textwidth}c c >{\raggedright\arraybackslash}p{0.10\textwidth}@{}}
\toprule
Task & Field & Inputs ($x_1$ then $x_2$) & $(p,q)$ & $[a,b]$ & Units ($x$, $y$) \\
\midrule
Reaction rate & Chemistry and nuclear physics & concentration of NO\newline concentration of O2 & $(2,1)$ & $[-6.5,-5.5]$ & mmol L$^{-1}$\newline mmol L$^{-1}$ s$^{-1}$ \\[3pt]
Beam deflection & Structural and solid mechanics & cross-section width\newline cross-section depth in the bending direction & $(-1,-3)$ & $[7.5,8.5]$ & mm\newline mm \\[3pt]
Conductor resistance & Electrical engineering and communications & axial length of the conductor\newline radius of the circular cross-section & $(1,-2)$ & $[-1,1]$ & mm\newline $\Omega$ \\[3pt]
Steady conduction & Heat transfer, optics and atmosphere & side length of the square heat-transfer face\newline conduction thickness & $(2,-1)$ & $[-2,0]$ & mm\newline W \\[3pt]
Laminar pipe flow & Fluid mechanics & radius of the circular channel\newline length of the channel & $(4,-1)$ & $[-6.5,-5.5]$ & $\mu$m\newline nL s$^{-1}$ \\[3pt]
Bending stress & Structural and solid mechanics & cross-section width\newline cross-section depth in the bending direction & $(-1,-2)$ & $[6.5,7.5]$ & mm\newline MPa \\[3pt]
Elastic buckling & Structural and solid mechanics & cross-section width\newline cross-section depth in the buckling plane & $(1,3)$ & $[-6.5,-5.5]$ & mm\newline kN \\[3pt]
Weir discharge & Fluid mechanics & crest width of the weir\newline head of water above the crest & $(1,1.5)$ & $[-4.5,-3.5]$ & mm\newline L s$^{-1}$ \\[3pt]
Orifice discharge & Fluid mechanics & radius of the circular orifice\newline head of water above the orifice centre & $(2,0.5)$ & $[-4.5,-3.5]$ & mm\newline L s$^{-1}$ \\[3pt]
Rayleigh range & Heat transfer, optics and atmosphere & beam waist radius\newline wavelength of the light & $(2,-1)$ & $[-1.5,-0.5]$ & $\mu$m\newline $\mu$m \\[3pt]
\bottomrule
\end{tabular}

\vspace{0.4em}
\begin{minipage}{\linewidth}
{\footnotesize Power-law tasks. Relations are $y=Cx_1^p x_2^q$ under \textit{Original assignment}. The two input descriptions give \textit{Original assignment} in column order, and \textit{Swapped assignment} exchanges them. The coefficient satisfies $\log_{10}C\sim U(a,b)$ in each instance, using the units shown, and was not supplied in the prompts. Examples use $x_1=x_2\in[100,200]$. The other forms and task-level results follow.}
\end{minipage}
\end{table}
\clearpage
\begin{table}[H]
\centering
\scriptsize
\setlength{\tabcolsep}{3pt}
\renewcommand{\arraystretch}{1.2}
\caption*{\textbf{\boldmath Supplementary Table 14 $\vert$ Controlled relation tasks, by task (continued).}}
\begin{tabular}{@{}>{\raggedright\arraybackslash}p{0.14\textwidth}>{\raggedright\arraybackslash}p{0.15\textwidth}>{\raggedright\arraybackslash}p{0.25\textwidth}>{\raggedright\arraybackslash}p{0.19\textwidth}>{\raggedright\arraybackslash}p{0.14\textwidth}>{\raggedright\arraybackslash}p{0.07\textwidth}@{}}
\toprule
Task & Field & Inputs ($x_1$ then $x_2$) & Relation & Constants & Units ($x$, $y$) \\
\midrule
\multicolumn{6}{@{}l}{\textit{Linear}} \\
Soil shear strength & Geotechnics & effective cohesion of the soil\newline effective normal stress acting on the shear surface & $x_1+x_2\tan\phi'$ & $\phi'=20^{\circ}$ & kPa\newline kPa \\[2pt]
Cantilever moment & Structural and solid mechanics & load at the free end\newline load at a quarter of the span from the support & $Lx_1+\tfrac{L}{4}x_2$ & $L=4$ m & kN\newline kN m \\[2pt]
Bearing capacity & Geotechnics & effective cohesion of the soil\newline overburden pressure at the footing base & $N_c x_1+N_q x_2$ & $\phi'=20^{\circ}$; $N_c=14.83$, $N_q=6.40$ & kPa\newline kPa \\[2pt]
\midrule
\multicolumn{6}{@{}l}{\textit{Exponential}} \\
Fall with linear drag & Dynamics & time elapsed since release\newline drag time constant of the sphere & $v_t[x_1-x_2(1-e^{-x_1/x_2})]$ & $v_t=2$ m s$^{-1}$ & s\newline mm \\[2pt]
Radioactive activity & Chemistry and nuclear physics & time elapsed since preparation\newline mean lifetime of the nuclide & $(N_0/x_2)\,e^{-x_1/x_2}$ & $N_0=10^{12}$ & h\newline MBq \\[2pt]
Air column mass & Heat transfer, optics and atmosphere & altitude above the surface\newline density scale height of the atmosphere & $\rho_0 x_2\,e^{-x_1/x_2}$ & $\rho_0=1.2$ kg m$^{-3}$ & km\newline t m$^{-2}$ \\[2pt]
\midrule
\multicolumn{6}{@{}l}{\textit{Rational}} \\
Load power & Electrical engineering and communications & internal resistance of the source\newline resistance of the load & $V^2x_2/(x_1+x_2)^2$ & $V=10$ V & $\Omega$\newline mW \\[2pt]
Inelastic coupling & Dynamics & mass of the moving wagon\newline mass of the stationary wagon & $v^2x_1^2/[2(x_1+x_2)]$ & $v=2$ m s$^{-1}$ & t\newline kJ \\[2pt]
Elastic collision & Dynamics & mass of the incident ball\newline mass of the struck ball & $2v^2x_1^2x_2/(x_1+x_2)^2$ & $v=2$ m s$^{-1}$ & g\newline mJ \\[2pt]
\midrule
\multicolumn{6}{@{}l}{\textit{Square root}} \\
Equivalent bending moment & Structural and solid mechanics & bending moment at the cross-section\newline torque transmitted by the shaft & $\tfrac12\bigl(x_1+\sqrt{x_1^2+x_2^2}\bigr)$ & -- & kN m\newline kN m \\[2pt]
Von Mises stress & Structural and solid mechanics & normal stress\newline shear stress & $\sqrt{x_1^2+3x_2^2}$ & -- & MPa\newline MPa \\[2pt]
Spring--mass frequency & Dynamics & mass attached to the spring\newline mass of the spring itself & $\frac{1}{2\pi}\sqrt{k/(x_1+x_2/3)}$ & $k=4{,}000$ N m$^{-1}$ & g\newline Hz \\[2pt]
\midrule
\multicolumn{6}{@{}l}{\textit{Logarithmic}} \\
Launch height, quadratic drag & Dynamics & launch speed\newline terminal speed of the ball & $\frac{x_2^2}{2g}\ln\bigl(1+x_1^2/x_2^2\bigr)$ & $g=9.81$ m s$^{-2}$ & m s$^{-1}$\newline m \\[2pt]
Logarithmic wind profile & Heat transfer, optics and atmosphere & zero-plane displacement height\newline aerodynamic roughness length & $\frac{u_*}{\kappa}\ln\frac{z-x_1}{x_2}$ & $u_*=0.5$ m s$^{-1}$, $\kappa=0.41$, $z=60$ m & m\newline m s$^{-1}$ \\[2pt]
Staged-fill settlement & Geotechnics & height of the first fill layer\newline height of the second fill layer & $\frac{HC_c}{1+e_0}\log_{10}\frac{\sigma'_0+\gamma(x_1+x_2)}{\sigma'_0+\gamma x_1}$ & $H=4$ m, $C_c=0.3$, $e_0=1.0$, $\sigma'_0=40$ kPa, $\gamma=20$ kN m$^{-3}$ & m\newline mm \\[2pt]
\bottomrule
\end{tabular}

\vspace{0.4em}
\begin{minipage}{\linewidth}
{\footnotesize Tasks with other functional forms. The relation is given under \textit{Original assignment}, with unit-conversion factors omitted. \textit{Swapped assignment} exchanges $x_1$ and $x_2$.}
\end{minipage}
\end{table}
\clearpage
\begin{table}[H]
\centering
\footnotesize
\setlength{\tabcolsep}{4pt}
\renewcommand{\arraystretch}{1.0}
\caption*{\textbf{\boldmath Supplementary Table 14 $\vert$ Controlled relation tasks, by task (continued).}}
\begin{tabular}{@{}lrrrrr@{}}
\toprule
Task & Original & Anonymous & Swapped & Separation & Positive \\
\midrule
\multicolumn{6}{@{}l}{\textit{Power laws}} \\
Reaction rate & $+1.00$ & $+0.00$ & $-1.00$ & $1.92$ [$1.75$, $2.00$] & 99/100 \\
Beam deflection & $+0.95$ & $+0.00$ & $-0.82$ & $1.51$ [$1.36$, $1.63$] & 99/100 \\
Conductor resistance & $+0.98$ & $+0.00$ & $-1.00$ & $1.88$ [$1.80$, $1.95$] & 100/100 \\
Steady conduction & $+1.00$ & $+0.00$ & $-0.99$ & $1.71$ [$1.67$, $1.90$] & 98/100 \\
Laminar pipe flow & $+1.00$ & $+0.00$ & $-1.00$ & $1.83$ [$1.72$, $1.93$] & 100/100 \\
Bending stress & $+1.00$ & $+0.00$ & $-1.00$ & $1.64$ [$1.51$, $1.96$] & 90/100 \\
Elastic buckling & $+0.50$ & $+0.00$ & $-0.46$ & $0.87$ [$0.73$, $1.00$] & 93/100 \\
Weir discharge & $+1.00$ & $+0.00$ & $-1.05$ & $2.27$ [$2.04$, $2.62$] & 98/100 \\
Orifice discharge & $+0.87$ & $+0.02$ & $-0.83$ & $1.08$ [$0.95$, $1.40$] & 95/100 \\
Rayleigh range & $+0.97$ & $+0.00$ & $-1.00$ & $1.97$ [$1.76$, $2.00$] & 99/100 \\
\midrule
\multicolumn{6}{@{}l}{\textit{Linear}} \\
Soil shear strength & $+1.00$ & $+0.00$ & $-1.00$ & $1.30$ [$1.21$, $1.50$] & 95/100 \\
Cantilever moment & $+0.89$ & $+0.00$ & $-1.00$ & $1.78$ [$1.66$, $1.85$] & 78/100 \\
Bearing capacity & $+0.68$ & $+0.00$ & $-0.67$ & $1.34$ [$1.23$, $1.57$] & 95/100 \\
\midrule
\multicolumn{6}{@{}l}{\textit{Exponential}} \\
Fall with linear drag & $+0.66$ & $+0.00$ & $-0.54$ & $1.13$ [$0.99$, $1.23$] & 99/100 \\
Radioactive activity & $+1.12$ & $+0.00$ & $-1.08$ & $2.26$ [$2.07$, $2.46$] & 95/100 \\
Air column mass & $+0.77$ & $+0.00$ & $-0.77$ & $1.53$ [$1.36$, $1.63$] & 80/100 \\
\midrule
\multicolumn{6}{@{}l}{\textit{Rational}} \\
Load power & $+0.40$ & $+0.00$ & $-0.64$ & $1.00$ [$0.98$, $1.11$] & 86/100 \\
Inelastic coupling & $+0.56$ & $+0.00$ & $-0.66$ & $1.30$ [$1.24$, $1.37$] & 93/100 \\
Elastic collision & $+0.63$ & $+0.00$ & $-0.58$ & $1.33$ [$1.18$, $1.52$] & 82/100 \\
\midrule
\multicolumn{6}{@{}l}{\textit{Square root}} \\
Equivalent bending moment & $+0.59$ & $+0.00$ & $-0.51$ & $1.16$ [$0.99$, $1.33$] & 86/100 \\
Von Mises stress & $+1.00$ & $+0.00$ & $-1.00$ & $1.00$ [$0.89$, $1.23$] & 85/100 \\
Spring--mass frequency & $+0.39$ & $+0.00$ & $-0.36$ & $1.00$ [$0.65$, $1.24$] & 64/100 \\
\midrule
\multicolumn{6}{@{}l}{\textit{Logarithmic}} \\
Launch height, quadratic drag & $+1.65$ & $+0.00$ & $-1.77$ & $2.62$ [$2.42$, $2.91$] & 98/100 \\
Logarithmic wind profile & $+0.61$ & $+0.00$ & $-0.53$ & $0.86$ [$0.65$, $1.01$] & 95/100 \\
Staged-fill settlement & $+0.41$ & $+0.00$ & $-0.42$ & $0.86$ [$0.82$, $0.98$] & 100/100 \\
\bottomrule
\end{tabular}

\vspace{0.4em}
\begin{minipage}{\linewidth}
{\footnotesize Relation scores under the three conditions are medians across 20 instances and five LLMs. Separation is the median of $T_{\mathrm{Original}}-T_{\mathrm{Swapped}}$ within each LLM and instance, with 95\% confidence intervals, and Positive is the number of LLM and task-instance combinations with positive separation, that is, in which exchanging the names moved the predictions in the direction implied by the names.}
\end{minipage}
\end{table}

\clearpage

\begin{table}[H]
\centering
\small
\caption*{\textbf{\boldmath Supplementary Table 15 $\vert$ Controlled relation tasks, by LLM.}}
\label{si:tab15}
\begin{tabular}{@{}lrrrrr@{}}
\toprule
 & & & Deviation from & Deviation from & Closer to \\
LLM & Separation & Positive & assigned (\%) & exchanged (\%) & assigned (\%) \\
\midrule
Claude Opus 4.5 & $1.86$ [1.76, 1.91] & 495/500 & 9.2 & 100.0 & 87.4 \\
GPT-5.2 & $1.47$ [1.37, 1.57] & 476/500 & 18.7 & 100.0 & 79.5 \\
Gemini 3 Flash & $2.00$ [2.00, 2.00] & 497/500 & 0.7 & 100.0 & 91.1 \\
Qwen3.5-397B & $1.00$ [0.92, 1.06] & 459/500 & 30.9 & 93.6 & 74.1 \\
DeepSeek-V3.2 & $0.46$ [0.40, 0.55] & 375/500 & 56.2 & 93.1 & 62.9 \\
\midrule
Five LLMs combined & $1.42$ [1.34, 1.48] & 2,302/2,500 & 19.5 & 100.0 & 79.0 \\
\bottomrule
\end{tabular}

\vspace{0.4em}
\begin{minipage}{\linewidth}
{\footnotesize Separation and Positive are defined as in Supplementary Table~14. The deviation columns give how far a typical prediction was from the target implied by the assigned names and from that implied by the exchanged names, and Closer to assigned is the percentage of predictions closer to the former. These columns use the 29,941 valid predictions under \textit{Original assignment} and \textit{Swapped assignment}.}
\end{minipage}
\end{table}

\clearpage

\subsection*{Predictions when in-context examples contradict domain-specific knowledge}
\label{si:conflict}

When the examples contradicted the knowledge elicited by the feature names, the predictions followed the examples more slowly than without names. The design and the prior weight are described in Methods, and the 12 tasks are listed in \sitab{16}. Because each relation score is the logarithm of the ratio between two predictions, one for an input and one with its two values swapped (Methods), pairs were excluded when either prediction could not be read as a number or was not positive. This removed 118 of 18,000 pairs with contradicting examples and 46 of 9,000 with supporting examples. When a response contained more than one number, the first was used.

\textbf{Supporting examples.}
Without names, the predictions moved towards supporting examples (Methods) almost as far as towards contradicting ones, reaching $+0.861$ with five supporting examples and $-0.858$ with five contradicting ones (Fig.~6c of the main text). The same held for every LLM, with DeepSeek-V3.2 responding most weakly in both directions. Under \textit{Anonymous}, the predictions therefore followed the examples rather than a default preference.

\textbf{Contradicting examples.}
With feature names, five contradicting examples moved the median relation score only to $-0.24$, a shift that took about 2.4 contradicting examples without names (\sitab{17}). The prior weight, the share of the prediction attributable to the prior rather than to the examples, was 0.43 (95\% confidence interval, 0.38 to 0.51). It ranged from 0.81 for Gemini 3 Flash to 0.04 for DeepSeek-V3.2, in the same order as the separation of the LLMs in the controlled relation tasks (\sitab{15}). Across tasks, the prior weight varied from 0.17 for Equivalent bending moment to 1.00 for Soil shear strength, whose predictions followed the names even when all five examples contradicted them (\sitab{16}).

\begin{table}[H]
\centering
\footnotesize
\setlength{\tabcolsep}{4pt}
\renewcommand{\arraystretch}{1.05}
\caption*{\textbf{\boldmath Supplementary Table 16 $\vert$ Contradicting examples, by task.}}
\label{si:tab16}
\begin{tabular}{@{}llrrrrrrr@{}}
\toprule
 & & \multicolumn{6}{c}{Contradicting examples, $k$} & \\
\cmidrule(lr){3-8}
Task & Condition & 0 & 1 & 2 & 3 & 4 & 5 & Prior weight \\
\midrule
\multicolumn{9}{@{}l}{\textit{Power law}} \\
Reaction rate & Original & $+1.00$ & $+0.50$ & $+0.65$ & $+0.26$ & $+0.02$ & $-0.43$ & $0.46$ [$0.34$, $0.60$] \\
 & Anonymous & $+0.00$ & $-0.08$ & $-0.19$ & $-0.45$ & $-0.60$ & $-0.70$ & \\[2pt]
Beam deflection & Original & $+0.91$ & $+0.47$ & $+0.18$ & $-0.22$ & $-0.60$ & $-0.70$ & $0.29$ [$0.24$, $0.36$] \\
 & Anonymous & $+0.00$ & $-0.20$ & $-0.36$ & $-0.63$ & $-0.80$ & $-0.93$ & \\[2pt]
\midrule
\multicolumn{9}{@{}l}{\textit{Linear}} \\
Soil shear strength & Original & $+1.00$ & $+1.00$ & $+1.00$ & $+1.00$ & $+1.00$ & $+1.00$ & $1.00$ [$1.00$, $1.00$] \\
 & Anonymous & $+0.00$ & $-0.12$ & $-0.03$ & $-0.47$ & $-0.84$ & $-0.97$ & \\[2pt]
Cantilever moment & Original & $+0.78$ & $+0.98$ & $+0.57$ & $+0.97$ & $+0.56$ & $+0.30$ & $0.71$ [$0.44$, $0.86$] \\
 & Anonymous & $+0.00$ & $+0.00$ & $+0.00$ & $+0.00$ & $-0.03$ & $-0.22$ & \\[2pt]
\midrule
\multicolumn{9}{@{}l}{\textit{Exponential}} \\
Fall with linear drag & Original & $+0.66$ & $+0.36$ & $+0.24$ & $+0.00$ & $-0.49$ & $-0.84$ & $0.25$ [$0.21$, $0.33$] \\
 & Anonymous & $+0.00$ & $-0.02$ & $-0.18$ & $-0.47$ & $-0.81$ & $-0.93$ & \\[2pt]
Radioactive activity & Original & $+1.05$ & $+1.06$ & $+1.00$ & $+0.93$ & $+0.66$ & $+0.20$ & $0.87$ [$0.74$, $0.96$] \\
 & Anonymous & $+0.00$ & $-0.14$ & $-0.22$ & $-0.36$ & $-0.69$ & $-0.80$ & \\[2pt]
\midrule
\multicolumn{9}{@{}l}{\textit{Rational}} \\
Load power & Original & $+0.21$ & $+0.20$ & $+0.11$ & $+0.00$ & $-0.12$ & $-0.42$ & $0.20$ [$0.11$, $0.32$] \\
 & Anonymous & $+0.00$ & $+0.00$ & $+0.00$ & $-0.13$ & $-0.66$ & $-0.89$ & \\[2pt]
Inelastic coupling & Original & $+0.58$ & $+0.56$ & $+0.51$ & $+0.50$ & $+0.33$ & $+0.06$ & $0.48$ [$0.40$, $0.61$] \\
 & Anonymous & $+0.00$ & $+0.00$ & $-0.05$ & $-0.12$ & $-0.04$ & $-0.51$ & \\[2pt]
\midrule
\multicolumn{9}{@{}l}{\textit{Square root}} \\
Equivalent bending moment & Original & $+0.84$ & $+0.08$ & $+0.00$ & $+0.00$ & $-0.21$ & $-0.51$ & $0.17$ [$0.06$, $0.25$] \\
 & Anonymous & $+0.00$ & $+0.00$ & $+0.00$ & $-0.26$ & $-0.86$ & $-0.92$ & \\[2pt]
Von Mises stress & Original & $+1.00$ & $+0.22$ & $+0.12$ & $+0.05$ & $+0.04$ & $+0.00$ & $0.23$ [$0.18$, $0.55$] \\
 & Anonymous & $+0.00$ & $+0.00$ & $+0.00$ & $-0.02$ & $-0.28$ & $-0.88$ & \\[2pt]
\midrule
\multicolumn{9}{@{}l}{\textit{Logarithmic}} \\
Launch height, quadratic drag & Original & $+1.68$ & $+0.81$ & $+0.59$ & $+0.10$ & $+0.00$ & $-0.12$ & $0.56$ [$0.16$, $0.82$] \\
 & Anonymous & $-0.01$ & $+0.00$ & $-0.30$ & $-1.01$ & $-0.77$ & $-0.66$ & \\[2pt]
Logarithmic wind profile & Original & $+0.61$ & $+0.00$ & $+0.00$ & $-0.12$ & $-0.64$ & $-0.77$ & $0.20$ [$0.14$, $0.30$] \\
 & Anonymous & $+0.00$ & $-0.19$ & $-0.38$ & $-0.72$ & $-0.85$ & $-0.95$ & \\[2pt]
\bottomrule
\end{tabular}

\vspace{0.4em}
\begin{minipage}{\linewidth}
{\footnotesize Values are median relation scores over five LLMs and ten instances, and $k=0$ repeats the corresponding instances of \sitab{14}. The design and the prior weight, with its 95\% confidence interval, are described in Methods.}
\end{minipage}
\end{table}

\clearpage

\begin{table}[H]
\centering
\footnotesize
\setlength{\tabcolsep}{4pt}
\renewcommand{\arraystretch}{1.05}
\caption*{\textbf{\boldmath Supplementary Table 17 $\vert$ Contradicting examples, by LLM.}}
\label{si:tab17}
\begin{tabular}{@{}llrrrrrrr@{}}
\toprule
 & & \multicolumn{6}{c}{Contradicting examples, $k$} & \\
\cmidrule(lr){3-8}
LLM & Condition & 0 & 1 & 2 & 3 & 4 & 5 & Prior weight \\
\midrule
Five LLMs combined & Original & $+0.89$ & $+0.56$ & $+0.37$ & $+0.17$ & $+0.00$ & $-0.24$ & $0.43$ [$0.38$, $0.51$] \\
 & Anonymous & $+0.00$ & $+0.00$ & $-0.11$ & $-0.41$ & $-0.68$ & $-0.86$ & \\
\midrule
Claude Opus 4.5 & Original & $+1.00$ & $+0.88$ & $+0.93$ & $+0.62$ & $+0.42$ & $+0.00$ & $0.75$ [$0.64$, $0.85$] \\
 & Anonymous & $+0.00$ & $-0.15$ & $-0.31$ & $-0.55$ & $-0.81$ & $-0.94$ & \\[2pt]
GPT-5.2 & Original & $+0.84$ & $+0.52$ & $+0.28$ & $+0.13$ & $-0.08$ & $-0.65$ & $0.41$ [$0.32$, $0.50$] \\
 & Anonymous & $+0.00$ & $-0.10$ & $-0.35$ & $-0.62$ & $-0.84$ & $-0.95$ & \\[2pt]
Gemini 3 Flash & Original & $+1.00$ & $+1.00$ & $+1.00$ & $+0.76$ & $+0.34$ & $+0.13$ & $0.81$ [$0.73$, $0.91$] \\
 & Anonymous & $+0.00$ & $-0.05$ & $-0.25$ & $-0.67$ & $-0.86$ & $-0.99$ & \\[2pt]
Qwen3.5-397B & Original & $+0.71$ & $+0.35$ & $+0.16$ & $+0.05$ & $+0.00$ & $-0.50$ & $0.29$ [$0.22$, $0.38$] \\
 & Anonymous & $+0.00$ & $+0.00$ & $-0.09$ & $-0.34$ & $-0.62$ & $-0.83$ & \\[2pt]
DeepSeek-V3.2 & Original & $+0.20$ & $+0.04$ & $+0.00$ & $+0.00$ & $-0.10$ & $-0.41$ & $0.04$ [$0.01$, $0.10$] \\
 & Anonymous & $+0.00$ & $+0.00$ & $+0.00$ & $-0.04$ & $-0.30$ & $-0.34$ & \\[2pt]
\bottomrule
\end{tabular}

\vspace{0.4em}
\begin{minipage}{\linewidth}
{\footnotesize Values are median relation scores over 12 tasks and ten instances per task under \textit{Original assignment} and \textit{Anonymous}. The first block combines the five LLMs. Prior weight is defined in Methods. Separation in the controlled relation tasks is given in \sitab{15} for comparison.}
\end{minipage}
\end{table}

\clearpage

\subsection*{Tests of Bayesian updating with 20 in-context examples}
\label{si:bayes}

A Bayesian observer in which the knowledge elicited by the names forms the prior described how the predictions of four LLMs changed with up to 20 contradicting examples, but not those of DeepSeek-V3.2 (Methods). The condition without contradicting examples ($k=0$) was run anew with 20 examples rather than taken from the five-example experiment. Pairs of predictions were excluded when either could not be parsed or was not positive (69 of 36,000), and confidence intervals used 2,000 resamples of instances within tasks.

\textbf{Bayesian observer.}
Fitted to all $k$, the observer followed the observed scores of both conditions closely for four LLMs and less closely for DeepSeek-V3.2 (\sifig{9}).

\textbf{From posterior probability to relation score.}
The choice of mixing rule had little effect on the fitted observer. The observer converts the posterior probability $\pi$ of the Original relation into a relation score by assuming that each prediction mixes the two relations in log space, which gives $T=2\pi-1$ (Methods). To test this assumption, we estimated $\pi$ for each instance from its predictions, assuming either log-space or linear-space mixing. The two rules fitted the predictions about equally well, with log-space mixing giving the smaller error in 48\% of instances, so the data did not distinguish between them. The values of $\pi$ inferred under the two rules were nevertheless nearly identical, correlating at 0.97 with a median absolute difference of 0.02 across 11{,}992 instances.

\textbf{Likelihood with and without names.}
For four LLMs, the likelihood did not differ detectably with and without names, whereas for DeepSeek-V3.2 it did. To test whether the names changed the likelihood, we fitted a second version of the observer in which the log likelihood ratio per example was multiplied by a ratio $\rho$ when names were given, with one $\rho$ shared by the 12 tasks of each LLM. Under the observer model, if variable names change only the prior and do not alter the likelihood, the ratio $\rho$ equals 1. For GPT-5.2, Gemini 3 Flash, Claude Opus 4.5 and Qwen3.5-397B, $\rho$ ranged from 1.01 to 1.35, and every 95\% confidence interval included 1 (\sifig{10}a). For GPT-5.2 and Gemini 3 Flash, the 90\% intervals also lay within the equivalence bounds of 1/1.5 to 1.5, establishing equivalence\cite{lakens2017equivalence}. DeepSeek-V3.2 was the exception, with $\rho=2.20$ (1.69 to 2.59), meaning that with the names its predictions moved towards the examples faster than without them.

The three LLMs with the strongest priors also responded most strongly to each contradicting example (\sifig{10}b). Gemini 3 Flash, Claude Opus 4.5 and GPT-5.2 had median prior probabilities of the Original relation of 0.92, 0.90 and 0.81, and each contradicting example under \textit{Anonymous} multiplied the odds of the Swapped relation by 1.52, 1.35 and 1.58, respectively. By contrast, Qwen3.5-397B and DeepSeek-V3.2 had lower priors of 0.72 and 0.60 and smaller likelihood ratios of 1.22 and 1.07.

\textbf{Goodness of fit.}
The goodness-of-fit test did not reject the Bayesian observer for any of the five LLMs. For each LLM, we simulated 1,000 datasets from the fitted observer, adding noise of the size observed among task instances, and computed the sum of squared deviations between task medians and the observer in each dataset (Methods)\cite{wichmann2001psychometric}. $P$ is the share of simulations with a sum at least as large as the observed one. For Claude Opus 4.5, GPT-5.2, Gemini 3 Flash and Qwen3.5-397B, the observed sums (12.6, 6.3, 6.7 and 9.0) lay within the simulated distributions (medians 10.3, 6.2, 7.2 and 9.7), giving $P$ values of 0.22, 0.46, 0.63 and 0.61 (Fig.~6f of the main text). Task by task, none of the 60 tests across the five LLMs rejected the observer after Holm correction. The observer gave the poorest fit for DeepSeek-V3.2. Its observed sum of 10.0 exceeded the simulated median of 7.0, although the test did not reject the observer ($P=0.15$). Allowing the names to change the likelihood ratio reduced the sum for DeepSeek-V3.2 to 7.3 ($P = 0.37$), consistent with its $\rho$ of 2.20.

\textbf{Out-of-sample prediction.}
We fitted the observer only to prompts with at most five contradicting examples and used it to predict the relation scores with 8 to 20 contradicting examples, in both conditions (Methods). At the held-out values $k=8,12,16,20$, the predicted and observed prior weights were correlated at $r=0.970$ for Claude Opus 4.5, $0.914$ for GPT-5.2, $0.870$ for Gemini 3 Flash, $0.997$ for Qwen3.5-397B and $-0.799$ for DeepSeek-V3.2 (Fig.~6e of the main text). With names, the observed relation scores deviated from the predicted ones by $-0.09$ ($-0.23$ to $+0.29$) for Claude Opus 4.5, $+0.02$ ($-0.07$ to $+0.16$) for GPT-5.2, $-0.06$ ($-0.15$ to $+0.08$) for Gemini 3 Flash and $-0.10$ ($-0.25$ to $+0.04$) for Qwen3.5-397B, with every interval including zero (\sifig{11}). Under \textit{Anonymous}, their deviations ranged from $-0.07$ to $-0.02$, and the interval excluded zero only for Claude Opus 4.5 ($-0.04$, $-0.24$ to $-0.01$), whose predictions followed the examples slightly faster than the observer expected. DeepSeek-V3.2 deviated by $-0.40$ ($-0.51$ to $-0.24$) with names and by $-0.15$ ($-0.24$ to $-0.04$) under \textit{Anonymous}. In both conditions, its predictions moved towards the contradicting examples faster than the observer predicted. Under \textit{Original assignment}, the root mean squared error (RMSE) of the predicted relation scores, calculated across the 12 tasks at $k=8,12,16,20$, was 0.583 for Claude Opus 4.5, 0.260 for GPT-5.2, 0.401 for Gemini 3 Flash, 0.511 for Qwen3.5-397B and 0.516 for DeepSeek-V3.2. Each RMSE summarises 48 errors, comparing the observer's prediction with the median observed relation score for each task and held-out value of $k$. Across individual tasks, the mean absolute deviation at $k=8$--20 exceeded 0.5 in 5 of the 48 task--LLM pairs among the four LLMs, and in 4 of the 12 tasks for DeepSeek-V3.2 (\sifig{11}b). For three of these five pairs, relation scores with names remained near $+1$ up to $k=5$, so the fitted range could not constrain the prior. The remaining two pairs were both for Soil shear strength, where Claude Opus 4.5 adhered strictly to the names across all $k$ and Qwen3.5-397B showed relation scores near zero.

\clearpage
\begin{figure}[H]
  \centering
  \captionsetup{font=footnotesize}
  \includegraphics[width=142.13mm]{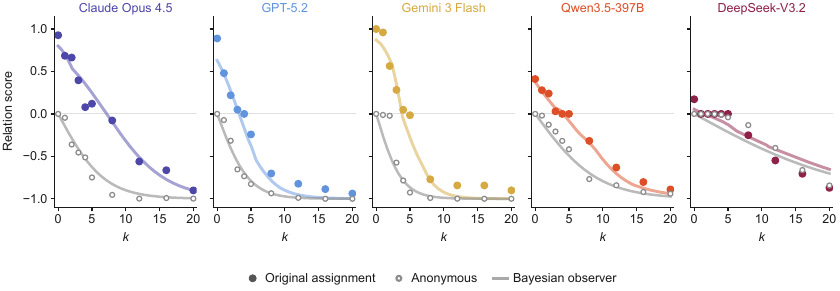}
  \caption{\textbf{\boldmath Relation scores with 20 in-context examples and the Bayesian observer.}
  Points show the median relation score over 12 tasks and 10 instances when $k$ of 20 examples contradict the knowledge elicited by the names, under \textit{Original assignment} (filled) and \textit{Anonymous} (open). Translucent curves show the Bayesian observer, in which the knowledge elicited by the names forms the prior and both conditions share the likelihood ratio per example. The observer was fitted to each task, and the curves show the median across tasks.}
  \label{si:fig9}
\end{figure}

\begin{figure}[H]
  \centering
  \captionsetup{font=footnotesize}
  \includegraphics[width=148.27mm]{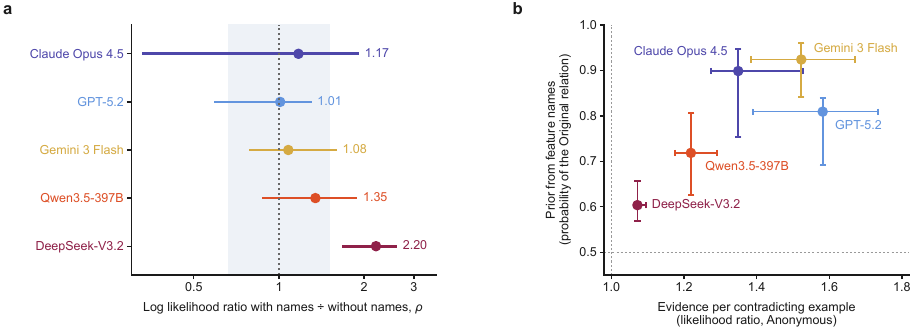}
  \caption{\textbf{\boldmath Prior and likelihood ratio of each LLM.}
  \textbf{a}, Ratio $\rho$ of the log likelihood ratio per example with names to that without names. $\rho=1$ means that, in the observer, the knowledge elicited by the names changes only the prior, and the shaded band is the equivalence margin of 1.5-fold.
  \textbf{b}, Median prior probability of the Original relation, the relation implied by the names, against the median likelihood ratio per contradicting example under \textit{Anonymous}, across 12 tasks. Both come from the observer with a shared $\rho$. Dotted lines mark an uninformative prior (0.5) and uninformative examples (1).
  Error bars show 95\% confidence intervals.}
  \label{si:fig10}
\end{figure}

\clearpage
\begin{figure}[H]
  \centering
  \captionsetup{font=footnotesize}
  \includegraphics[width=144.46mm]{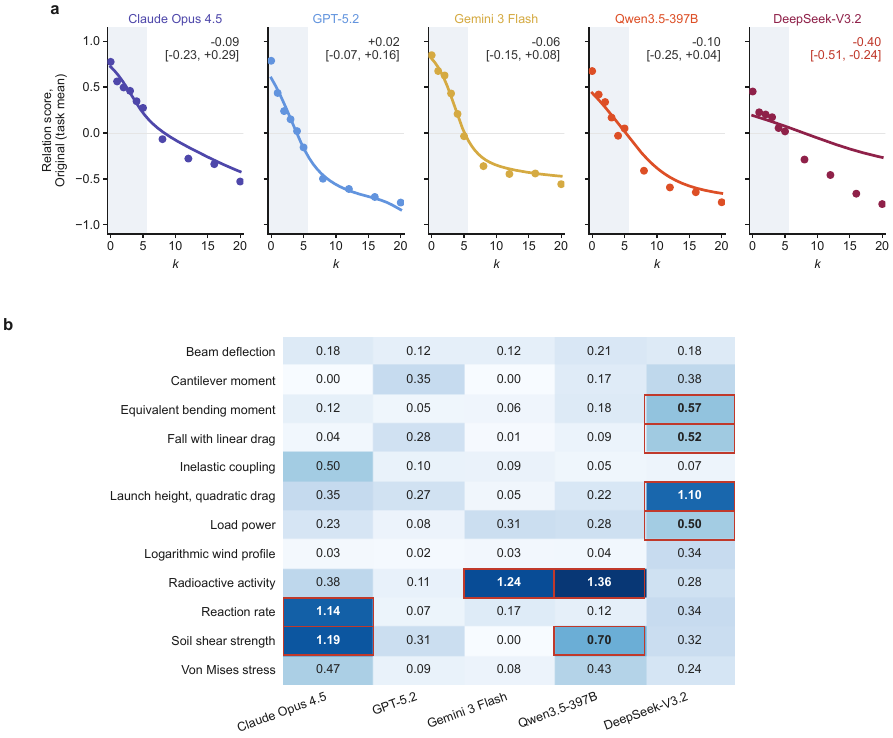}
  \caption{\textbf{\boldmath Out-of-sample prediction of the Bayesian observer.}
  \textbf{a}, Observed relation scores under \textit{Original assignment} (points, mean of 12 task medians) and the observer's prediction (line) when the prior and the likelihood ratio were fitted to both conditions at $k\leq5$ only (shaded). Numbers give the mean deviation of the observed from the predicted scores at $k=8$--20 with its 95\% confidence interval, in red where it excludes zero. \textbf{b}, Mean absolute deviation of the observed from the predicted relation scores at $k=8$--20 under \textit{Original assignment}, for each task and LLM. Red outlines mark deviations larger than 0.5.}
  \label{si:fig11}
\end{figure}

\clearpage

\clearpage

\section*{Supplementary Methods}
\label{si:methods}

\subsection*{Datasets and their sources}
\label{si:datasets}

\enlargethispage{4\baselineskip}

\begin{table}[H]
\centering
\caption*{\textbf{\boldmath Supplementary Table 18 $\vert$ The 20 real-world tabular datasets.}}
\label{tab:si_datasets}\label{si:tab18}
\footnotesize
\setlength{\tabcolsep}{4pt}
\begin{tabular}{@{}L{3.9cm} L{4.1cm} r r L{3.2cm} L{2.2cm}@{}}
\toprule
Dataset & Domain & Samples & Features & Target variable & Split \\
\midrule
Splitting Strength & Structural engineering & 550 & 9 & Splitting tensile strength (MPa) & Random \\
Shear Force & Structural engineering & 156 & 9 & Shear force (kN) & Random \\
Max Displacement & Structural engineering & 230 & 12 & Maximum impact displacement (mm) & Random \\
MIDR (MDOF) & Structural engineering & 320 & 15 & Maximum interstorey drift ratio & Random \\
Torsional Strength & Structural engineering & 150 & 7 & Torsional strength (kN\,m) & Random \\
FRP--Steel Bond & Structural engineering & 113 & 12 & Ultimate bearing capacity (kN) & Random \\
Corrosion Degree & Non-destructive testing & 640 & 5 & Corrosion degree & Random \\
Undrained Shear Strength & Geotechnical engineering & 523 & 7 & Undrained shear strength & Random \\
Formation Enthalpy & Chemistry & 296 & 30 & Enthalpy of formation (kcal\,mol$^{-1}$) & Random \\
Hydrogen Uptake$^{\dagger}$ & Chemistry & 150 & 5 & H uptake (appm H/C) & Isotherm (19) \\
Carbon Intensity & Building carbon emissions & 821 & 14 & Carbon intensity (ln) & Random \\
Drug--Polymer Solubility & Biomedicine & 199 & 30 & Drug mole fraction & Random \\
Drug--Water Solubility & Biomedicine & 199 & 30 & Solubility (log) & Random \\
Daily Trips (Citi Bike)$^{\dagger}$ & Urban mobility & 150 & 6 & Daily station trips & Random \\
Daily Trips (Divvy)$^{\dagger}$ & Urban mobility & 150 & 6 & Daily station trips & Random \\
Hourly Ridership$^{\dagger}$ & Urban mobility & 150 & 6 & Hourly route ridership & Random \\
Mesiodistal Diameter & Dental morphology & 150 & 4 & Mesiodistal diameter (mm) & Individual (49) \\
Buccolingual Diameter & Dental morphology & 150 & 4 & Buccolingual diameter (mm) & Individual (49) \\
Elastic Modulus (FDM)$^{\dagger}$ & Additive manufacturing & 150 & 8 & Elastic modulus (MPa) & Process (150) \\
Shore Hardness (FDM)$^{\dagger}$ & Additive manufacturing & 150 & 8 & Shore hardness & Process (150) \\
\bottomrule
\end{tabular}

\vspace{0.4em}
\begin{minipage}{\linewidth}
{\footnotesize Numbers in parentheses in the Split column give the number of groups used for grouped splits. For the two drug-solubility datasets, compounds were shared between the example and test sets. Daggers mark datasets first released after the training cutoffs of all five LLMs (\sitab{19}). The abbreviations are MIDR for maximum interstorey drift ratio, MDOF for multiple-degree-of-freedom, FDM for fused deposition modelling and FRP for fibre-reinforced polymer.}
\end{minipage}

\end{table}

\clearpage

\clearpage

\begin{table}[H]
\centering
\caption*{\textbf{\boldmath Supplementary Table 19 $\vert$ Provenance of the 20
datasets.}}
\label{tab:si_provenance}\label{si:tab19}
\scriptsize
\setlength{\tabcolsep}{4pt}
\begin{tabular}{@{}L{3.3cm} L{3.3cm} L{4.4cm} L{2.3cm} L{2.2cm}@{}}
\toprule
Source & Datasets & DOI or access URL & Licence & First public release \\
\midrule
Citi Bike system data (Jersey City) & Daily Trips (Citi Bike) &
\url{https://citibikenyc.com/system-data} & Operator system-data licence &
Monthly; 2026-06 slice used \\
Divvy system data (Chicago) & Daily Trips (Divvy) &
\url{https://divvybikes.com/system-data} & Operator system-data licence &
Monthly; 2026-06 slice used \\
MTA Bus Hourly Ridership (Beginning 2025) & Hourly Ridership &
\url{https://data.ny.gov/Transportation/MTA-Bus-Hourly-Ridership-Beginning-2025/gxb3-akrn} &
NY Open Data terms of use & Monthly; 2026-06 slice used \\
\midrule
FDM printed specimens, mechanical properties &
Elastic Modulus (FDM); Shore Hardness (FDM) &
\href{https://doi.org/10.17632/zd6td6svd6}{10.17632/zd6td6svd6} (Mendeley Data) &
CC BY 4.0 & 2026-07-17 \\
Nuclear graphite hydrogen uptake (Univ.\ Illinois Urbana--Champaign) &
Hydrogen Uptake &
\href{https://doi.org/10.17632/wn33rrkvj4}{10.17632/wn33rrkvj4} (Mendeley Data) &
CC BY 4.0 & 2026-07-23 \\
\midrule
Rat\'{o}n P\'{e}rez deciduous tooth collection (CENIEH, Spain) &
Mesiodistal Diameter; Buccolingual Diameter &
\href{https://doi.org/10.5281/zenodo.17255973}{10.5281/zenodo.17255973} (Zenodo) &
CC BY 4.0 & 2025-10-03 (v1) \\
Yalamanchi \emph{et al.} (2019) & Formation Enthalpy &
\href{https://doi.org/10.1021/acs.jpca.9b04771}{10.1021/acs.jpca.9b04771} &
See publisher terms & 2019-09-16 \\
This work (authors' research group) & Splitting Strength; Shear Force; Max
Displacement; Corrosion Degree; Undrained Shear Strength; MIDR (MDOF);
Torsional Strength; FRP--Steel Bond & Deposited with this paper &
CC BY 4.0 & Not previously released \\
Provided by collaborators & Carbon Intensity; Drug--Polymer Solubility; Drug--Water Solubility &
Deposited with this paper & CC BY 4.0 &
Not previously released \\
\bottomrule
\end{tabular}

\vspace{0.4em}
\begin{minipage}{\linewidth}
{\footnotesize ``First public release'' is the date on which the data became public, not that of an accompanying article. A dataset is post-cutoff if this date follows 16 February 2026, the latest training cutoff or release date of the five LLMs (Methods). The cutoffs were August 2025 for Claude Opus 4.5\cite{anthropic2025opus45}, 31 August 2025 for GPT-5.2\cite{openai2026gpt52}, January 2025 for Gemini 3 Flash\cite{google2026gemini3flash} and March 2025 for DeepSeek-V3.2\cite{aws2025deepseekv32}. Month-level cutoffs were taken as the last day of the month. Qwen3.5-397B was released on 16 February 2026\cite{qwen2026qwen35}. The dental datasets are not post-cutoff, because version 3 used here shares all rows with version 1, released on 3 October 2025. The urban-mobility datasets use June 2026 records of services that have published data for years, so Supplementary Table~\sitabn{1} also reports the three experimental datasets separately.}
\end{minipage}
\end{table}

\clearpage

\subsection*{Prompt templates of the main experiment}
\label{si:prompts}

The main experiment used two templates, one without variable names (\textit{Anonymised}) and one with them (\textit{Correct variable names}). All other conditions modified one of these two templates, as described below. The templates contain four placeholders. \verb|{target}| is the target name, \verb|{features}| the comma-separated feature names, \verb|{examples}| the in-context examples and \verb|{test_sample}| the row to be predicted. In each in-context example, the features were written on one line as \verb|name: value| pairs and the target on the next line. In the row to be predicted, each feature was written on its own line. \textit{Anonymised} used the same layout without the names. When $n=0$, the \verb|TRAINING EXAMPLES| section was omitted.

\vspace{0.4em}
\textbf{Anonymised.}

{\footnotesize\begin{verbatim}
You are an expert in predicting target values.

TASK:
Predict the target value based on the provided numerical features.

TRAINING EXAMPLES:
The following examples show the relationship between features and target values.

{examples}

INFERENCE QUERY:
Predict the target value for the following sample.

{test_sample}

OUTPUT FORMAT:
Return only a single numeric value.
Do not provide any explanation, reasoning steps, or additional text.
\end{verbatim}}

\textbf{Correct variable names.}

{\footnotesize\begin{verbatim}
You are an expert in predicting {target}.

TASK:
Predict the target value ({target}) based on the provided features: {features}.

TRAINING EXAMPLES:
The following examples are provided for learning the underlying relationships.

{examples}

INFERENCE QUERY:
Now predict the target value for the following sample.

{test_sample}

OUTPUT FORMAT:
Return only a single numeric value representing the predicted target value
({target}).
Do not provide any explanation, reasoning steps, or additional text.
\end{verbatim}}

\textbf{$+$ Semantic guidance.} This condition inserts the following text into the \textit{Correct variable names} template, between TASK and TRAINING EXAMPLES.

{\footnotesize\begin{verbatim}
FEATURE SEMANTICS:
Each feature name (column header) encodes semantic, physical, or domain-specific
meaning.
Treat feature names as informative variables rather than anonymous numerical
dimensions.
\end{verbatim}}

\textbf{$+$ Causal guidance.} This condition replaces that text with the following one, and the opening line is extended to ``You are an expert in predicting \verb|{target}|, specializing in understanding semantic relationships and causal patterns.''

{\footnotesize\begin{verbatim}
FEATURE SEMANTICS AND PRINCIPLES:
Each feature name (column header) encodes semantic, physical, or domain-specific
meaning.
Treat feature names as informative variables rather than anonymous numerical
dimensions.

When learning from the training examples, implicitly reason about:

* the semantic meaning of each feature,
* the underlying functional or causal relationships between features and the
    target.

Prioritize semantically plausible predictions
(e.g., respecting monotonicity, scaling behavior, physical or logical
constraints).
\end{verbatim}}

\textbf{Conditions of Supplementary Note~3.} \textit{Target name only} used the \textit{Anonymised} template with the target name in place of the generic label. \textit{Feature names only} used the \textit{Correct variable names} template with the target name replaced by ``target values''. \textit{Shuffled names} and \textit{Cross-domain names} used the \textit{Correct variable names} template with the same features, feature order and values but other names. Under \textit{Cross-domain names}, each dataset received the names of a dataset from another group of related datasets. The donors were the 12 datasets in four groups, namely structural and geotechnical engineering, non-destructive testing, building carbon emissions and urban mobility. When the donor had fewer features than the recipient, the remaining names came from other donors, and the target name came from the first donor. \textit{Rescaled values} used both templates with the rescaled values, rounded to a fixed number of significant figures per column. In all other conditions, values kept their recorded form.

\textbf{Parsing of responses.} Responses that were a single number were used directly. From other responses, dollar signs and thousands separators were removed. The first number matching \verb|-?\d+\.?\d*| was then used for responses of at most 50 characters, and the last for longer ones. Responses without a number were recorded as unparseable.

\clearpage

\subsubsection*{Prompt template of the standardised format}
\label{si:alternative-prompts}

{\sloppy \textit{Anonymised} and \textit{Feature names} share the following template and differ only in \verb|{field_definitions}|.\par}

{\footnotesize\begin{verbatim}
TASK
Predict y for the query using the examples below.

TARGET
y: {target_name}{target_unit_line}{feature_units_block}

FIELD DEFINITIONS
{field_definitions}

EXAMPLES
{examples}

QUERY
{query}

OUTPUT
Return only one numeric prediction for y.
Do not include an explanation.
\end{verbatim}}

\verb|{target_name}| is the recorded target name. The target unit and the feature units, where recorded, are added below it under \verb|unit:| and \verb|FEATURE UNITS|.

Features are coded \verb|x1| to \verb|xk| in the original column order. \textit{Anonymised} defines each code by its position, such as \verb|x1: input variable 1|, and \textit{Feature names} by its original name, such as \verb|x1: {original_feature_name_1}|.

Each example occupies one line, {\footnotesize\verb|x1={value_1}; ...; xk={value_k}; y={target_value}|}, and the row to be predicted omits \verb|y|. Text values are replaced by codes such as \verb|v1_1| and \verb|v1_2|, numbered within each column by first appearance in the dataset.

To separate the effect of the names from that of the values, \textit{Original values} restores the recorded values of the nine datasets with text values, keeping the \textit{Feature names} definitions. Results are reported in \sitab{8}.

\clearpage

\subsubsection*{Prompt template of the controlled relation tasks}
\label{si:relation-prompts}

\textit{Original assignment}, \textit{Swapped assignment} and \textit{Anonymous} used the following template and differed only in the descriptions under \verb|FIELD DEFINITIONS|.

{\footnotesize\begin{verbatim}
TASK
{task_setting}

TARGET
y: {target_name}
unit: {target_unit}

FEATURE UNITS
x1: {input_unit}
x2: {input_unit}

FIELD DEFINITIONS
x1: {description_1}
x2: {description_2}

EXAMPLES
{examples}

QUERY
x1={query_value_1}; x2={query_value_2}

OUTPUT
Return only one numeric prediction for y.
Do not include an explanation.
\end{verbatim}}

Each example was written as \verb|x1={u}; x2={u}; y={value}|. For the power laws, the six prediction inputs were $(100,200)$, $(200,100)$, $(110,190)$, $(190,110)$, $(125,180)$ and $(180,125)$. \textit{Original assignment} used the two input descriptions of \sitab{14} in order, and \textit{Swapped assignment} exchanged them. \textit{Anonymous} gave the two inputs neutral descriptions, such as ``first geometric dimension'' and ``second geometric dimension'' for Beam deflection or ``one of the two speeds'' for Launch height, quadratic drag. Predictions were parsed by removing commas and taking the first number matching the following regular expression.

{\footnotesize\begin{verbatim}
[-+]?\d*\.?\d+(?:[eE][-+]?\d+)?
\end{verbatim}}

Prediction pairs containing an unparseable, zero or negative prediction were excluded (107 of 22,500).

As an example, the task setting of Beam deflection was ``Slender simply supported beam of
rectangular cross-section under a concentrated transverse load at mid-span,
linear elastic small-deflection setting, shear deformation neglected. Span,
load and elastic modulus are fixed within the table. Predict the mid-span
deflection magnitude.'' The target was ``mid-span deflection magnitude'', and
both input and target units were \verb|mm|. \textit{Original assignment} defined
\verb|x1| as ``cross-section width'' and \verb|x2| as ``cross-section depth in
the bending direction''.

\clearpage

\subsection*{Inference settings of the LLMs}
\label{si:inference}

\begin{table}[H]
\centering
\footnotesize
\caption*{\textbf{\boldmath Supplementary Table 20 $\vert$ Inference settings
for the five evaluated LLMs.}}
\label{si:tab20}
\scriptsize
\begin{tabular}{@{}lL{5.0cm}rrl@{}}
\toprule
Model & Resolved identifier & Temp. & Max tokens & Reasoning \\
\midrule
Claude Opus 4.5\cite{anthropic2025opus45} & \texttt{anthropic/claude-opus-4.5} & 0 & 200 & Off by default \\
GPT-5.2\cite{openai2026gpt52} & \texttt{openai/gpt-5.2} & 0 & 200 & \texttt{enabled:\,false} \\
Gemini 3 Flash\cite{google2026gemini3flash} & \texttt{google/gemini-3-flash-preview} & 0 & 200 & Off by default \\
Qwen3.5-397B\cite{qwen2026qwen35} & \texttt{qwen/qwen3.5-397b-a17b} & 0 & 200 & \texttt{effort:\,"none"} \\
DeepSeek-V3.2\cite{deepseek2025v32} & \texttt{deepseek/deepseek-v3.2} & 0 & 200 & \texttt{effort:\,"none"} \\
\bottomrule
\end{tabular}

\vspace{0.4em}
\begin{minipage}{\linewidth}
{\footnotesize All experiments in this study were run between January and October 2026.}
\end{minipage}
\end{table}

\clearpage

\subsection*{Hyperparameters of the supervised ML models}
\label{si:supervised}

\begin{table}[H]
\centering
\footnotesize
\captionsetup{justification=raggedright,singlelinecheck=false}
\caption*{\textbf{\boldmath Supplementary Table 21 $\vert$ Hyperparameter
search spaces and tuning procedures for the supervised ML models.}}
\label{si:tab21}
\begin{tabular}{@{}llL{6.0cm}@{}}
\toprule
Supervised ML model & Hyperparameter & Search space or procedure \\
\midrule
\multirow{8}{*}{Tuned XGBoost}
 & \texttt{n\_estimators} & 50 to 500 (integer) \\
 & \texttt{max\_depth} & 3 to 10 (integer) \\
 & \texttt{learning\_rate} & $0.01$ to $0.3$ (log-uniform) \\
 & \texttt{subsample} & 0.6 to 1.0 \\
 & \texttt{colsample\_bytree} & 0.6 to 1.0 \\
 & \texttt{min\_child\_weight} & 1 to 10 (integer) \\
 & \texttt{reg\_alpha} & $10^{-8}$ to 10 (log-uniform) \\
 & \texttt{reg\_lambda} & $10^{-8}$ to 10 (log-uniform) \\
\midrule
\multirow{5}{*}{Tuned RF}
 & \texttt{n\_estimators} & 50 to 500 (integer) \\
 & \texttt{max\_depth} & 3 to 30 (integer) \\
 & \texttt{min\_samples\_split} & 2 to 20 (integer) \\
 & \texttt{min\_samples\_leaf} & 1 to 10 (integer) \\
 & \texttt{max\_features} & 0.3 to 1.0 \\
\midrule
Ridge & \texttt{alpha} & $10^{-3}$ to $10^{3}$, 25 points on a log grid; selected
by efficient leave-one-out cross-validation \\
\midrule
\multirow{2}{*}{Gaussian process}
 & kernel & constant $\times$ radial-basis-function kernel $+$ white noise \\
 & length scale, noise & fitted by marginal likelihood, 3 restarts; no
validation split \\
\midrule
Ordinary least squares & --- & no hyperparameter; the minimum-norm solution is
taken when the number of features exceeds $n$ \\
\bottomrule
\end{tabular}

\vspace{0.4em}
\begin{minipage}{\linewidth}
{\footnotesize\raggedright Optuna used a tree-structured Parzen estimator sampler seeded with the random seed of each run. TabPFN used the TabPFN-3 checkpoint \texttt{tabpfn-v3-regressor-v3\_default.ckpt} (SHA-256 \texttt{311ce18d97e9533d8585eaadafe040fbdd8070533209ed8696641dadc97a7301}). CARTE used version 0.0.26 with scikit-learn 1.5.2, \texttt{category\_encoders} 2.6.3 and the released \texttt{cc.en.300.bin} FastText vectors. CARTE was run on an NVIDIA RTX 4090. The default \texttt{freeze\_pretrain=True} setting froze the graph-attention blocks, while the input transformations and prediction head were trained with early stopping, using a single model per fit and a validation fraction of 0.2. Early stopping used squared error, because $R^2$ is undefined for the single validation row at $n=5$.\par}
\end{minipage}
\end{table}

\subsection*{Paired comparisons and confidence intervals}

Unless stated otherwise, effects in the Supplementary Notes are paired Hodges--Lehmann estimates, the median of all pairwise averages of the differences. \sitab{22} compares two conditions on the same datasets, seeds and LLMs, giving 500 paired differences per comparison (20 datasets $\times$ 5 seeds $\times$ 5 LLMs). Comparisons between the LLMs and supervised ML models were paired by dataset and seed only, giving 100 differences per LLM (20 datasets $\times$ 5 seeds). The $P$ value of each comparison comes from a two-sided Wilcoxon signed-rank test. Each effect has two 95\% confidence intervals, one from the signed-rank test and one from resampling the 20 datasets 2,000 times (Methods). Standard deviations in the tables describe the variation across the five seeds.

\clearpage

\section*{Supplementary Statistical Results}
\label{si:statistics}

{\scriptsize
\begin{longtable}{{@{}J{5.1cm}lrrrr@{}}}
\caption*{\textbf{\boldmath Supplementary Table 22 $\vert$ Paired comparisons.}}\label{si:tab22} \\
\toprule
Comparison & Metric & HL estimate & \shortstack{95\% CI\\(signed rank)} & \shortstack{95\% CI\\(dataset bootstrap)} & $P$ \\
\midrule
\endfirsthead
\toprule
Comparison & Metric & HL estimate & \shortstack{95\% CI\\(signed rank)} & \shortstack{95\% CI\\(dataset bootstrap)} & $P$ \\
\midrule
\endhead
\multicolumn{6}{@{}l}{\textit{LLMs versus supervised ML models across numbers of examples}} \\
\quad Five LLMs versus three supervised ML models, $n=5$ & $R^2$ & $0.3526$ & $0.2700$ to $0.4360$ & $0.2354$ to $0.4848$ & $2.0\times10^{-11}$ \\
\quad Five LLMs versus three supervised ML models, $n=5$ & WAPE & $12.22$ & $9.74$ to $15.27$ & $8.39$ to $17.83$ & $2.0\times10^{-14}$ \\
\quad Five LLMs versus three supervised ML models, $n=10$ & $R^2$ & $0.1976$ & $0.1490$ to $0.2513$ & $0.1317$ to $0.2659$ & $2.9\times10^{-11}$ \\
\quad Five LLMs versus three supervised ML models, $n=10$ & WAPE & $8.90$ & $6.96$ to $11.06$ & $5.79$ to $12.79$ & $4.4\times10^{-15}$ \\
\quad Five LLMs versus three supervised ML models, $n=25$ & $R^2$ & $0.0810$ & $0.0493$ to $0.1202$ & $0.0425$ to $0.1369$ & $3.1\times10^{-7}$ \\
\quad Five LLMs versus three supervised ML models, $n=25$ & WAPE & $5.19$ & $3.10$ to $6.77$ & $2.42$ to $7.47$ & $2.3\times10^{-11}$ \\
\quad Five LLMs versus three supervised ML models, $n=50$ & $R^2$ & $0.0488$ & $0.0285$ to $0.0730$ & $0.0196$ to $0.0834$ & $1.1\times10^{-6}$ \\
\quad Five LLMs versus three supervised ML models, $n=50$ & WAPE & $3.13$ & $2.29$ to $4.24$ & $1.77$ to $5.12$ & $1.6\times10^{-10}$ \\
\quad Five LLMs versus three supervised ML models, $n=100$ & $R^2$ & $0.0004$ & $-0.0117$ to $0.0146$ & $-0.0177$ to $0.0222$ & $0.937$ \\
\quad Five LLMs versus three supervised ML models, $n=100$ & WAPE & $0.75$ & $0.10$ to $1.41$ & $-0.21$ to $1.86$ & $0.019$ \\
\midrule
\multicolumn{6}{@{}l}{\textit{Prompt condition, $n=5$}} \\
\quad Correct variable names $-$ Anonymised & $R^2$ & $0.2988$ & $0.2587$ to $0.3418$ & $0.2112$ to $0.4067$ & $5.9\times10^{-54}$ \\
\quad $+$ Semantic guidance $-$ Correct variable names & $R^2$ & $0.0008$ & $-0.0049$ to $0.0062$ & $-0.0051$ to $0.0061$ & $0.773$ \\
\quad $+$ Causal guidance $-$ Correct variable names & $R^2$ & $0.0000$ & $-0.0071$ to $0.0072$ & $-0.0119$ to $0.0113$ & $1.000$ \\
\quad Correct variable names $-$ Anonymised & WAPE & $-8.84$ & $-10.00$ to $-7.73$ & $-12.06$ to $-5.98$ & $1.5\times10^{-57}$ \\
\quad $+$ Semantic guidance $-$ Correct variable names & WAPE & $-0.03$ & $-0.19$ to $0.14$ & $-0.17$ to $0.12$ & $0.739$ \\
\quad $+$ Causal guidance $-$ Correct variable names & WAPE & $0.04$ & $-0.16$ to $0.23$ & $-0.25$ to $0.32$ & $0.722$ \\
\midrule
\multicolumn{6}{@{}l}{\textit{Prompt condition, $n=10$}} \\
\quad Correct variable names $-$ Anonymised & $R^2$ & $0.1522$ & $0.1232$ to $0.1848$ & $0.0883$ to $0.2505$ & $4.5\times10^{-41}$ \\
\quad Correct variable names $-$ Anonymised & WAPE & $-5.82$ & $-6.80$ to $-4.89$ & $-9.01$ to $-2.94$ & $6.9\times10^{-50}$ \\
\midrule
\multicolumn{6}{@{}l}{\textit{Prompt condition, $n=25$}} \\
\quad Correct variable names $-$ Anonymised & $R^2$ & $0.0921$ & $0.0775$ to $0.1084$ & $0.0581$ to $0.1413$ & $1.2\times10^{-43}$ \\
\quad Correct variable names $-$ Anonymised & WAPE & $-3.81$ & $-4.50$ to $-3.18$ & $-5.80$ to $-2.12$ & $1.4\times10^{-51}$ \\
\midrule
\multicolumn{6}{@{}l}{\textit{Prompt condition, $n=50$}} \\
\quad Correct variable names $-$ Anonymised & $R^2$ & $0.0495$ & $0.0395$ to $0.0605$ & $0.0281$ to $0.0811$ & $7.6\times10^{-36}$ \\
\quad Correct variable names $-$ Anonymised & WAPE & $-2.33$ & $-2.79$ to $-1.93$ & $-3.88$ to $-1.15$ & $5.8\times10^{-41}$ \\
\midrule
\multicolumn{6}{@{}l}{\textit{Prompt condition, $n=100$}} \\
\quad Correct variable names $-$ Anonymised & $R^2$ & $0.0298$ & $0.0225$ to $0.0379$ & $0.0153$ to $0.0514$ & $1.1\times10^{-23}$ \\
\quad $+$ Semantic guidance $-$ Correct variable names & $R^2$ & $0.0012$ & $-0.0001$ to $0.0028$ & $-0.0001$ to $0.0035$ & $0.075$ \\
\quad $+$ Causal guidance $-$ Correct variable names & $R^2$ & $0.0017$ & $0.0003$ to $0.0036$ & $0.0002$ to $0.0043$ & $0.017$ \\
\quad Correct variable names $-$ Anonymised & WAPE & $-1.59$ & $-1.89$ to $-1.31$ & $-2.52$ to $-0.80$ & $4.3\times10^{-31}$ \\
\quad $+$ Semantic guidance $-$ Correct variable names & WAPE & $-0.05$ & $-0.14$ to $0.02$ & $-0.16$ to $0.02$ & $0.185$ \\
\quad $+$ Causal guidance $-$ Correct variable names & WAPE & $-0.121$ & $-0.215$ to $-0.036$ & $-0.305$ to $-0.003$ & $0.0042$ \\
\midrule
\multicolumn{6}{@{}l}{\textit{Adding examples versus adding names, $n=5$}} \\
\quad Correct variable names, $n=5$ $-$ Correct variable names, $n=0$ (adding examples) & $R^2$ & $1.9845$ & $1.4819$ to $2.6028$ & $0.7402$ to $4.6837$ & $1.7\times10^{-69}$ \\
\quad Correct variable names, $n=5$ $-$ Anonymised, $n=5$ (adding names) & $R^2$ & $0.2988$ & $0.2587$ to $0.3418$ & $0.2112$ to $0.4067$ & $5.9\times10^{-54}$ \\
\quad Correct variable names, $n=5$ $-$ Correct variable names, $n=0$ (adding examples) & WAPE & $-37.97$ & $-44.62$ to $-31.96$ & $-68.05$ to $-19.75$ & $4.8\times10^{-74}$ \\
\quad Correct variable names, $n=5$ $-$ Anonymised, $n=5$ (adding names) & WAPE & $-8.84$ & $-10.00$ to $-7.73$ & $-12.06$ to $-5.98$ & $1.5\times10^{-57}$ \\
\midrule
\multicolumn{6}{@{}l}{\textit{Semantic perturbation, $n=5$}} \\
\quad Correct variable names $-$ Shuffled names & $R^2$ & $0.1643$ & $0.1375$ to $0.1925$ & $0.0952$ to $0.2445$ & $1.4\times10^{-35}$ \\
\quad Correct variable names $-$ Shuffled names & WAPE & $-5.51$ & $-6.51$ to $-4.66$ & $-8.43$ to $-3.03$ & $2.4\times10^{-46}$ \\
\quad Shuffled names $-$ Cross-domain names & $R^2$ & $0.1184$ & $0.0839$ to $0.1555$ & $0.0387$ to $0.2138$ & $9.7\times10^{-14}$ \\
\quad Shuffled names $-$ Cross-domain names & WAPE & $-2.90$ & $-3.68$ to $-2.13$ & $-5.04$ to $-1.13$ & $5.7\times10^{-14}$ \\
\quad Cross-domain names $-$ Anonymised & $R^2$ & $-0.0104$ & $-0.0375$ to $0.0151$ & $-0.0798$ to $0.0483$ & $0.423$ \\
\quad Cross-domain names $-$ Anonymised & WAPE & $0.51$ & $-0.09$ to $1.21$ & $-0.90$ to $2.35$ & $0.104$ \\
\quad Correct variable names $-$ Cross-domain names & $R^2$ & $0.3212$ & $0.2836$ to $0.3618$ & $0.2229$ to $0.4165$ & $1.6\times10^{-53}$ \\
\quad Correct variable names $-$ Cross-domain names & WAPE & $-9.02$ & $-10.14$ to $-7.98$ & $-12.75$ to $-5.97$ & $2.3\times10^{-58}$ \\
\quad Correct variable names $-$ Anonymised & $R^2$ & $0.2988$ & $0.2587$ to $0.3418$ & $0.2112$ to $0.4067$ & $5.9\times10^{-54}$ \\
\quad Correct variable names $-$ Anonymised & WAPE & $-8.84$ & $-10.00$ to $-7.73$ & $-12.06$ to $-5.98$ & $1.5\times10^{-57}$ \\
\quad Shuffled names $-$ Anonymised & $R^2$ & $0.0966$ & $0.0675$ to $0.1293$ & $0.0406$ to $0.1737$ & $5.0\times10^{-12}$ \\
\quad Shuffled names $-$ Anonymised & WAPE & $-2.37$ & $-3.07$ to $-1.71$ & $-3.81$ to $-1.09$ & $8.2\times10^{-12}$ \\
\midrule
\multicolumn{6}{@{}l}{\textit{Against the strongest supervised ML model, $n=100$}} \\
\quad Claude Opus 4.5 $-$ TabPFN & $R^2$ & $0.0006$ & $-0.0096$ to $0.0123$ & $-0.0137$ to $0.0182$ & $0.912$ \\
\quad Claude Opus 4.5 $-$ TabPFN & WAPE & $-0.27$ & $-0.87$ to $0.20$ & $-1.19$ to $0.40$ & $0.274$ \\
\midrule
\multicolumn{6}{@{}l}{\textit{Comparisons with supervised ML models, $n=5$}} \\
\quad Correct variable names, five LLMs combined $-$ TabPFN & $R^2$ & $0.2999$ & $0.2058$ to $0.3997$ & $0.1496$ to $0.4538$ & $8.8\times10^{-9}$ \\
\quad Correct variable names, five LLMs combined $-$ TabPFN & WAPE & $-10.05$ & $-13.38$ to $-7.36$ & $-16.46$ to $-5.61$ & $1.8\times10^{-10}$ \\
\quad Cross-domain names, five LLMs combined $-$ TabPFN & $R^2$ & $-0.0110$ & $-0.0761$ to $0.0521$ & $-0.0918$ to $0.0523$ & $0.705$ \\
\quad Cross-domain names, five LLMs combined $-$ TabPFN & WAPE & $-1.60$ & $-3.65$ to $0.15$ & $-4.68$ to $0.89$ & $0.071$ \\
\quad Anonymised, five LLMs combined $-$ TabPFN & $R^2$ & $0.0145$ & $-0.0708$ to $0.0997$ & $-0.1177$ to $0.1315$ & $0.713$ \\
\quad Anonymised, five LLMs combined $-$ TabPFN & WAPE & $-2.86$ & $-5.53$ to $-0.20$ & $-7.12$ to $1.45$ & $0.032$ \\
\pagebreak
\multicolumn{6}{@{}l}{\textit{Target name against feature names, $n=5$}} \\
\quad Feature names, target present & $R^2$ & $0.1960$ & $0.1665$ to $0.2259$ & $0.1298$ to $0.2729$ & $1.5\times10^{-44}$ \\
\quad Feature names, target absent & $R^2$ & $0.2373$ & $0.2025$ to $0.2756$ & $0.1587$ to $0.3250$ & $3.3\times10^{-46}$ \\
\quad Target name, features present & $R^2$ & $0.0450$ & $0.0318$ to $0.0594$ & $0.0237$ to $0.0780$ & $6.1\times10^{-12}$ \\
\quad Target name, features absent & $R^2$ & $0.0839$ & $0.0595$ to $0.1108$ & $0.0425$ to $0.1420$ & $5.5\times10^{-14}$ \\
\quad Shuffled names $-$ Target name only & $R^2$ & $0.0102$ & $-0.0085$ to $0.0291$ & $-0.0171$ to $0.0387$ & $0.287$ \\
\quad Interaction (feature names $\times$ target name) & $R^2$ & $-0.0347$ & $-0.0636$ to $-0.0080$ & $-0.0959$ to $0.0148$ & $0.0095$ \\
\quad Feature names, target present & WAPE & $-5.60$ & $-6.52$ to $-4.76$ & $-8.30$ to $-3.46$ & $7.3\times10^{-46}$ \\
\quad Feature names, target absent & WAPE & $-6.46$ & $-7.51$ to $-5.48$ & $-9.03$ to $-4.06$ & $7.3\times10^{-46}$ \\
\quad Target name, features present & WAPE & $-1.67$ & $-2.17$ to $-1.22$ & $-3.10$ to $-0.76$ & $4.5\times10^{-15}$ \\
\quad Target name, features absent & WAPE & $-2.23$ & $-2.86$ to $-1.63$ & $-3.52$ to $-1.20$ & $5.3\times10^{-15}$ \\
\quad Shuffled names $-$ Target name only & WAPE & $-0.05$ & $-0.55$ to $0.48$ & $-0.75$ to $0.74$ & $0.865$ \\
\quad Interaction (feature names $\times$ target name) & WAPE & $0.45$ & $-0.18$ to $1.14$ & $-0.83$ to $1.96$ & $0.162$ \\
\midrule
\multicolumn{6}{@{}l}{\textit{Numerical rescaling, $n=5$}} \\
\quad Correct variable names $-$ Anonymised & $R^2$ & $0.2988$ & $0.2587$ to $0.3418$ & $0.2112$ to $0.4067$ & $5.9\times10^{-54}$ \\
\quad Rescaled values $-$ Anonymised, rescaled values & $R^2$ & $0.1893$ & $0.1561$ to $0.2247$ & $0.1376$ to $0.2435$ & $1.5\times10^{-27}$ \\
\quad Correct variable names $-$ Rescaled values & $R^2$ & $0.1136$ & $0.0925$ to $0.1367$ & $0.0716$ to $0.1764$ & $1.6\times10^{-32}$ \\
\quad Anonymised $-$ Anonymised, rescaled values & $R^2$ & $0.0384$ & $0.0127$ to $0.0629$ & $-0.0083$ to $0.0819$ & $0.0039$ \\
\quad Interaction (rescaling $\times$ variable names) & $R^2$ & $0.0937$ & $0.0571$ to $0.1356$ & $0.0364$ to $0.1732$ & $6.1\times10^{-8}$ \\
\quad Correct variable names $-$ Anonymised & WAPE & $-8.84$ & $-10.00$ to $-7.73$ & $-12.06$ to $-5.98$ & $1.5\times10^{-57}$ \\
\quad Rescaled values $-$ Anonymised, rescaled values & WAPE & $-5.36$ & $-6.29$ to $-4.53$ & $-7.52$ to $-3.61$ & $7.6\times10^{-33}$ \\
\quad Correct variable names $-$ Rescaled values & WAPE & $-3.36$ & $-4.10$ to $-2.73$ & $-5.48$ to $-2.00$ & $6.5\times10^{-33}$ \\
\quad Anonymised $-$ Anonymised, rescaled values & WAPE & $-1.13$ & $-1.76$ to $-0.50$ & $-2.27$ to $0.04$ & $0.0005$ \\
\quad Interaction (rescaling $\times$ variable names) & WAPE & $-2.73$ & $-3.76$ to $-1.82$ & $-4.94$ to $-1.15$ & $8.0\times10^{-10}$ \\
\bottomrule
\end{longtable}
\vspace{0.3em}
\noindent{\footnotesize Rows labelled versus compare the LLMs with a supervised ML model, and positive values indicate an LLM advantage. All other rows give the first condition minus the second. For $R^2$, positive values indicate an advantage for the first condition, and for WAPE, in percentage points (pp), negative values do. The interaction of feature names and target name is the improvement from feature names with the target name minus that without it. The interaction of rescaling and variable names is the improvement from variable names with the original values minus that with rescaled values.\par}

}

\clearpage
\phantomsection
\label{si:references}
\bibliographystyle{naturemag}
\bibliography{references}